\documentclass{article}
\usepackage{ijcai26}

\usepackage{amsmath,bm}
\usepackage[dvipsnames]{xcolor}
\usepackage{amsfonts}

\definecolor{darkgreen}{rgb}{0.0, 0.7, 0.0}

\def\eqref#1{equation~\ref{#1}}

\def\1{\bm{1}}

\def\vmu{{\bm{\mu}}}

\def\vc{{\bm{c}}}
\def\vd{{\bm{d}}}

\def\vo{{\bm{o}}}

\def\vq{{\bm{q}}}
\def\vr{{\bm{r}}}
\def\vs{{\bm{s}}}
\def\vt{{\bm{t}}}

\def\vv{{\bm{v}}}
\def\vw{{\bm{w}}}
\def\vx{{\bm{x}}}

\def\mR{{\bm{R}}}
\def\mS{{\bm{S}}}

\def\mSigma{{\bm{\Sigma}}}

\DeclareMathAlphabet{\mathsfit}{\encodingdefault}{\sfdefault}{m}{sl}
\SetMathAlphabet{\mathsfit}{bold}{\encodingdefault}{\sfdefault}{bx}{n}

\def\sR{{\mathbb{R}}}

\def\sZ{{\mathbb{Z}}}

\DeclareMathOperator*{\atantwo}{atan2}

\usepackage{tabularx}
\usepackage{times}
\usepackage{soul}
\usepackage{url}
\usepackage[hidelinks]{hyperref}
\usepackage{cleveref}
\usepackage[utf8]{inputenc}
\usepackage[small]{caption}
\usepackage{graphicx}
\usepackage{amsmath}
\usepackage{amsthm}
\usepackage{booktabs}
\usepackage{algorithm}
\usepackage{algorithmic}
\usepackage[switch]{lineno}
\usepackage{multirow}
\usepackage[normalem]{ulem}
\useunder{\uline}{\ul}{}
\usepackage{svg}
\usepackage{xcolor}
\usepackage[toc,page]{appendix}
\usepackage{mathtools}

\title{Unified Sensor Simulation for Autonomous Driving}

\author{
Nikolay Patakin$^1$\and
Arsenii Shirokov$^1$\and
Anton Konushin$^1$\and
Dmitry Senushkin$^1$\\
\affiliations
$^1$Lomonosov Moscow State University\\
\emails
}

\begin{document}

\maketitle

\begin{abstract}
In this work, we introduce \textbf{XSIM}, a sensor simulation framework for autonomous driving. XSIM extends 3DGUT splatting with a generalized rolling-shutter modeling tailored for autonomous driving applications. Our framework provides a unified and flexible formulation for appearance and geometric sensor modeling, enabling rendering of complex sensor distortions in dynamic environments.
We identify spherical cameras, such as LiDARs, as a critical edge case for existing 3DGUT splatting due to cyclic projection and time discontinuities at azimuth boundaries leading to incorrect particle projection.
To address this issue, we propose a phase modeling mechanism that explicitly accounts temporal and shape discontinuities of Gaussians projected by the Unscented Transform at azimuth borders.
In addition, we introduce an extended 3D Gaussian representation that incorporates two distinct opacity parameters to resolve mismatches between geometry and color distributions.
As a result, our framework provides enhanced scene representations with improved geometric consistency and photorealistic appearance. We evaluate our framework extensively on multiple autonomous driving datasets, including Waymo Open Dataset, Argoverse 2, and PandaSet. Our framework consistently outperforms strong recent baselines and achieves state-of-the-art performance across all datasets.
The source code is publicly available at \href{https://github.com/whesense/XSIM}{https://github.com/whesense/XSIM}.
\end{abstract}

\begin{figure}[!ht]
    \centering
    \includegraphics[width=\linewidth]{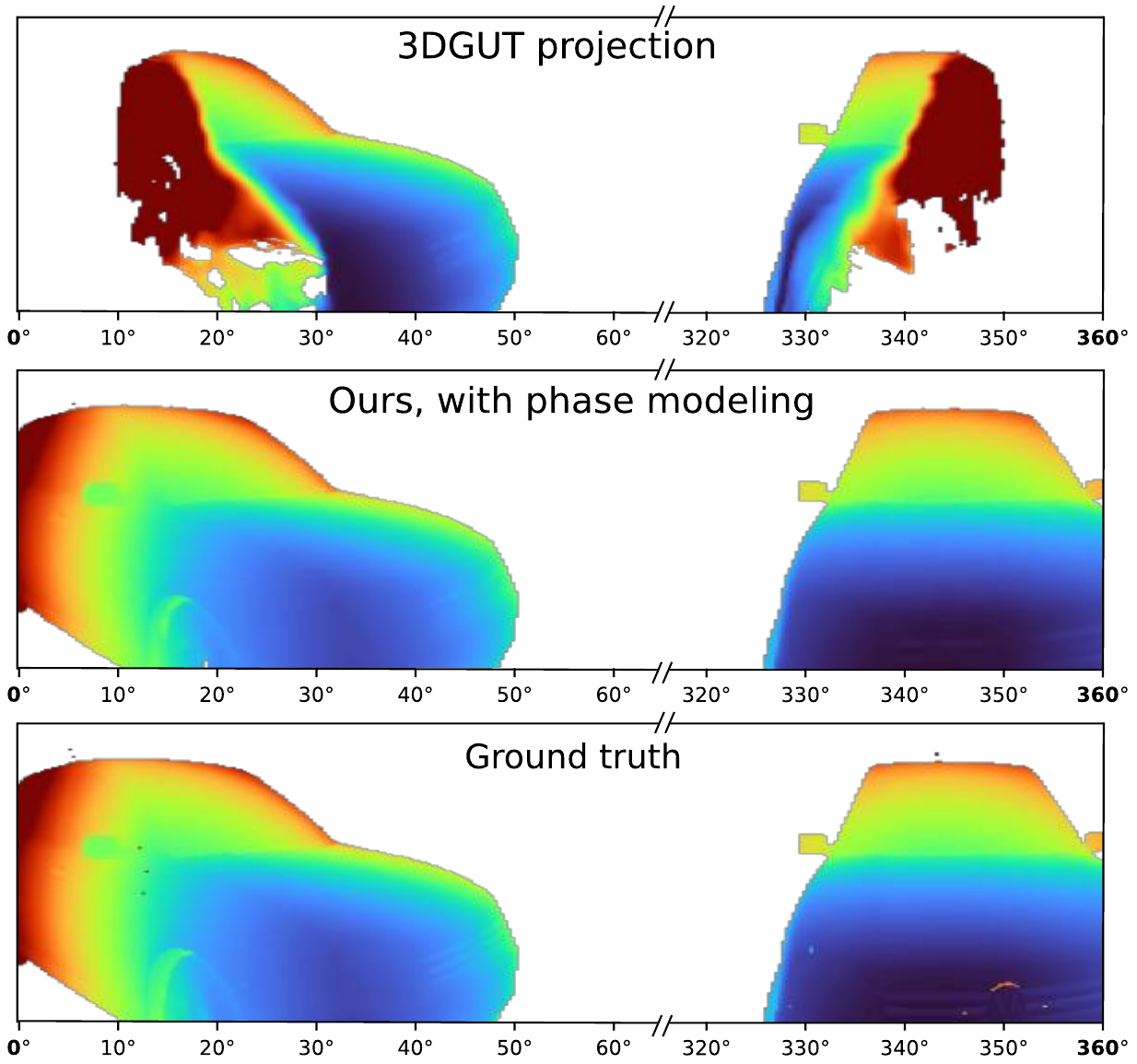}
    \caption{Synthetic example of LiDAR rendering. In regions near the azimuth discontinuity border, the standard 3DGUT projection results in partially missing and distorted range image renders. Our phase modeling approach alleviates this issue and also effectively handles surfaces observed twice due to the rolling shutter.}
    \label{fig:projection-ablation}
\end{figure}

\section{Introduction}
\label{sec:intro}
Development of safe and reliable autonomous vehicles necessitates access to large-scale and diverse datasets for both training and evaluation. However, acquiring diverse real-world data remains prohibitively expensive and labor-intensive. Simulation based on real data has emerged as a promising alternative, enabling the creation of augmented datasets in a cost and time efficient manner for various downstream applications~\cite{tianyuan2024presight,adamkiewicz2022vsnerfnav,ljungbergh2024neuroncap}. Recently, 3D Gaussian Splatting (3DGS)~\cite{kerbl20233dgs,wu20253dgut,hess2024splatad} has introduced a photorealistic rendering engine lowering sim-to-real gap and offering an effective balance among visual fidelity and computational efficiency.

\begin{figure*}[!ht]
    \centering
        \setlength{\tabcolsep}{1pt}
        \Large
        \begin{tabular}{ccc}
        
            \textbf{XSIM, Ours} & SplatAD & OmniRE \\

             \includegraphics[width=0.33\linewidth,trim={0 0 0 3cm},clip]{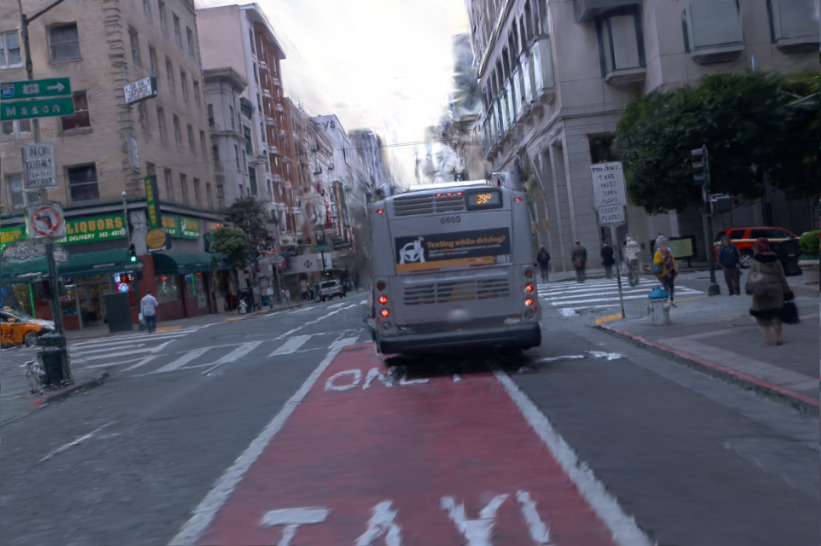} & 
             \includegraphics[width=0.33\linewidth,trim={0 0 0 3cm},clip]{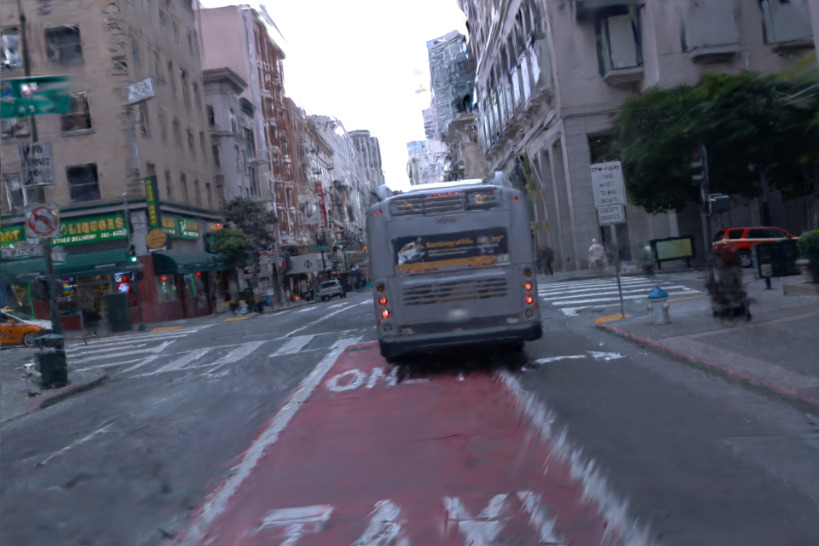} & 
             \includegraphics[width=0.33\linewidth,trim={0 0 0 3cm},clip]{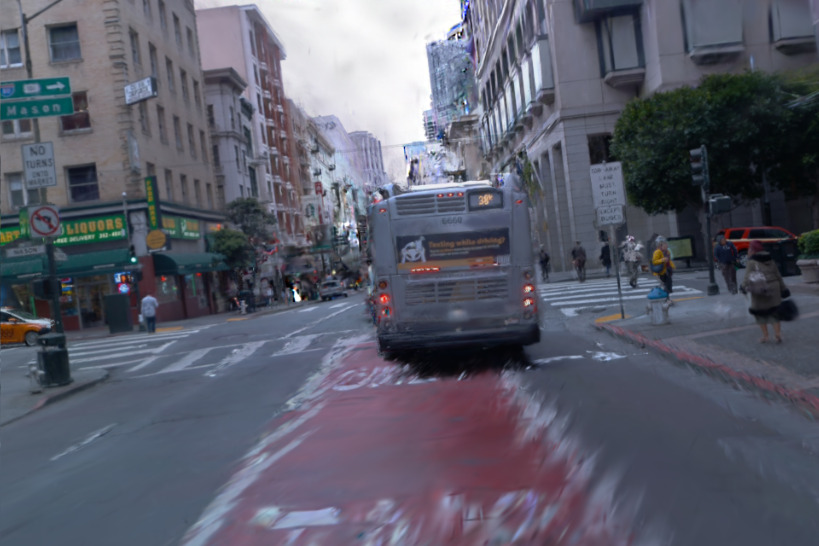} \\

             \includegraphics[width=0.33\linewidth]{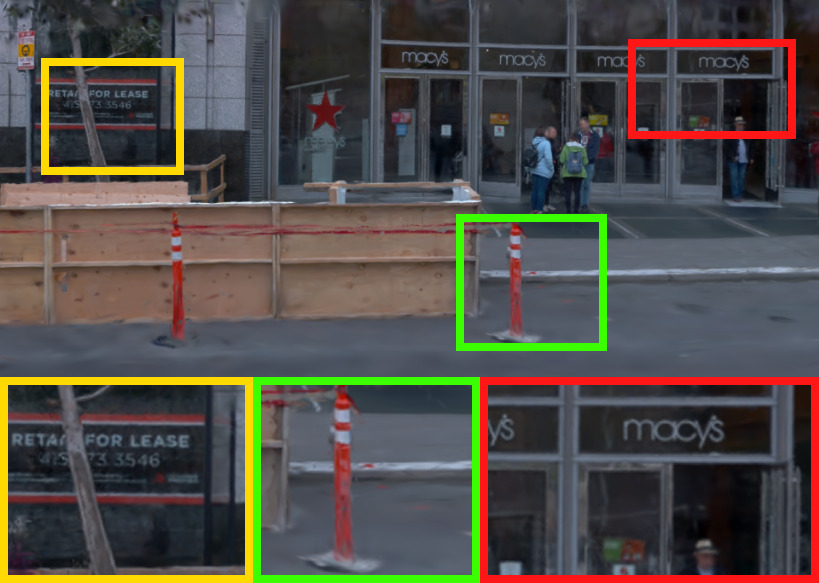} & 
             \includegraphics[width=0.33\linewidth]{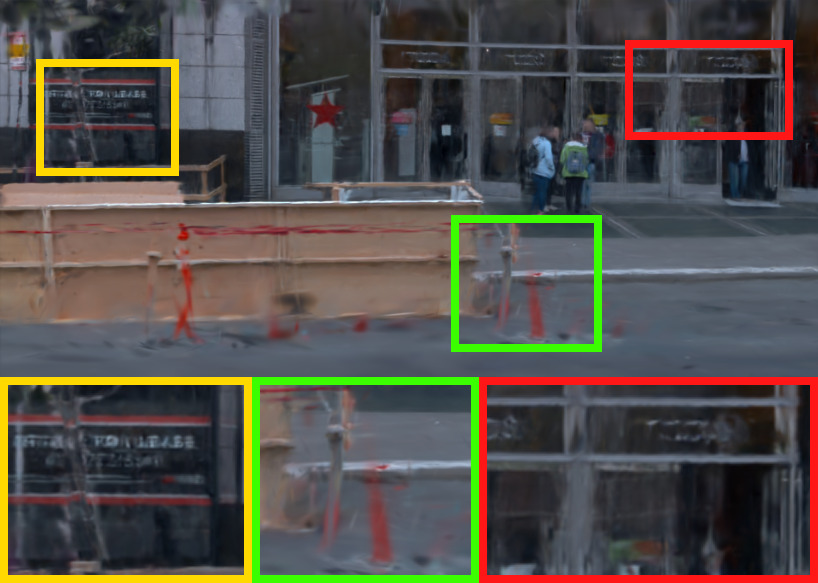} & 
             \includegraphics[width=0.33\linewidth]{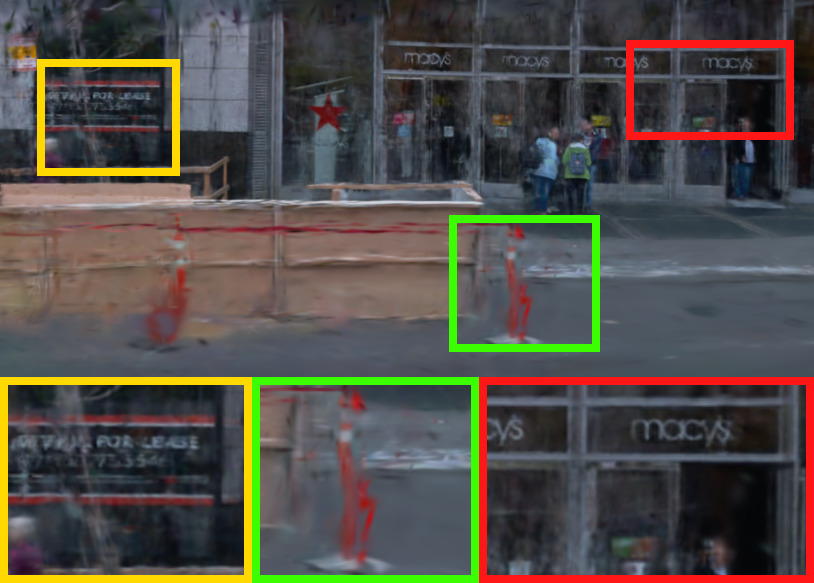} \\
    
        \end{tabular}
    \normalsize \caption{\normalsize \textbf{Novel view synthesis (Lane shift 3m)} Qualitative comparison on Waymo Open Dataset demonstrates that XSIM provides scene representation which can be consistently rendered from novel ego-vehicle trajectories. }
    \label{fig:nvs-lane-shift}
\end{figure*}

While existing simulators predominantly rely on EWA splatting \cite{zwicker2001ewags}, the recently proposed 3DGUT framework \cite{wu20253dgut} offers benefits for sensor simulation in autonomous driving scenarios. Real-world driving data is typically captured using cameras with highly nonlinear and dynamic effects, such as rolling shutter and optical distortions. These effects are difficult to model accurately with EWA splatting due to its reliance on linearized particle projection during rasterization, resulting in specialized rendering procedures for each camera type.
In contrast, 3DGUT projects Gaussians through arbitrary camera models using the Unscented Transform (UT), effectively emulating ray-based rendering within a rasterization framework and enabling accurate modeling of complex distortions. Building on this capability, we introduce a sensor simulation framework that adapts 3DGUT splatting to autonomous driving data. We further extend the framework with a generalized rolling-shutter modeling, enabling accurate simulation of LiDAR and arbitrary camera sensors in unified manner. 
However, we observe that UT-based projection for spherical cameras introduces challenges for Gaussian particles spanning the azimuth boundary. 
Naive processing of such particles leads to missing projections or distorted positions and shapes due to cyclic azimuth parameterization and temporal discontinuities at azimuth edges.
To address this issue, we propose a phase modeling mechanism that explicitly accounts for these effects, enabling accurate rendering of spherical cameras with rolling shutter.

Autonomous driving scenarios require jointly modeling geometric and appearance sensors. Accurate appearance modeling typically necessitates multiple transparent Gaussian particles to capture complex lighting effects, whereas geometric sensors provide precise surface measurements that can be faithfully represented using a single opaque Gaussian. To address this distribution mismatch, we extend scene parameterization with two distinct opacity parameters per Gaussian, jointly optimized for color and geometry distributions. Such parameterization alleviates potential mismatches within a unified representation and results in enhanced quality.

Eventually, our main contributions are three-fold:
\begin{itemize}
    \item We introduce XSIM -- the sensor simulation framework for autonomous driving extending 3DGUT splatting and enabling rendering LiDAR and camera sensors in a unified manner with generalized rolling shutter modeling.
    
    \item We propose a phase modeling mechanism for spherical rasterization that explicitly accounts temporal and shape discontinuities of Gaussians projected by the Unscented Transform at azimuth borders.
    
    \item We introduce an extended 3D Gaussian representation that incorporates two distinct opacity parameters per Gaussian, jointly optimized for color and geometry distributions, mitigating distribution mismatches within a unified representation.
\end{itemize}

\section{Related Work}
\label{sec:related}

\begin{figure*}[!ht]
    \centering
        \setlength{\tabcolsep}{1pt}
        \Large
        \begin{tabular}{cccc}
            Ground-truth image & \textbf{XSIM, Ours} & SplatAD & OmniRE \\
            
            \includegraphics[width=0.245\linewidth]{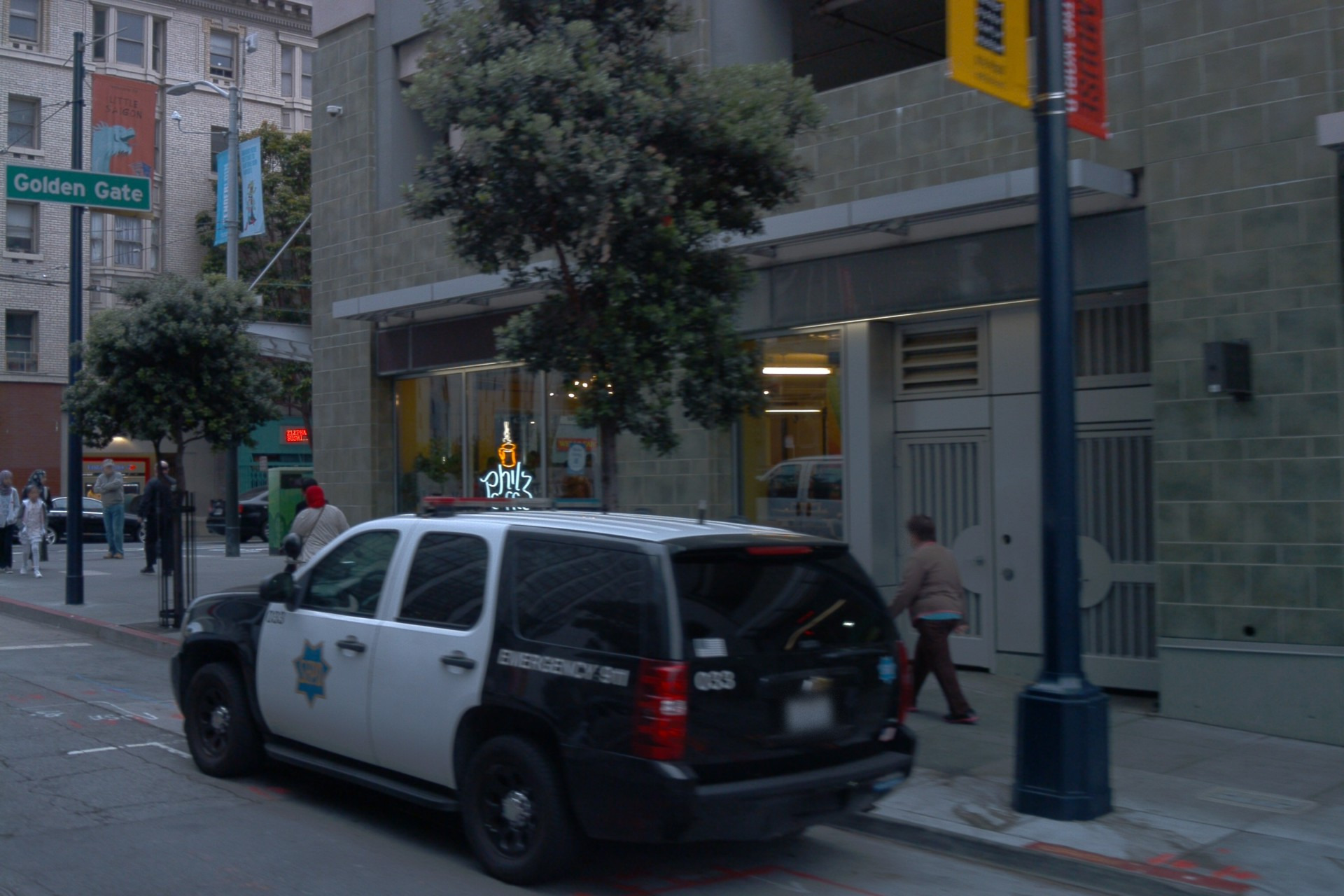} &
             \includegraphics[width=0.245\linewidth]{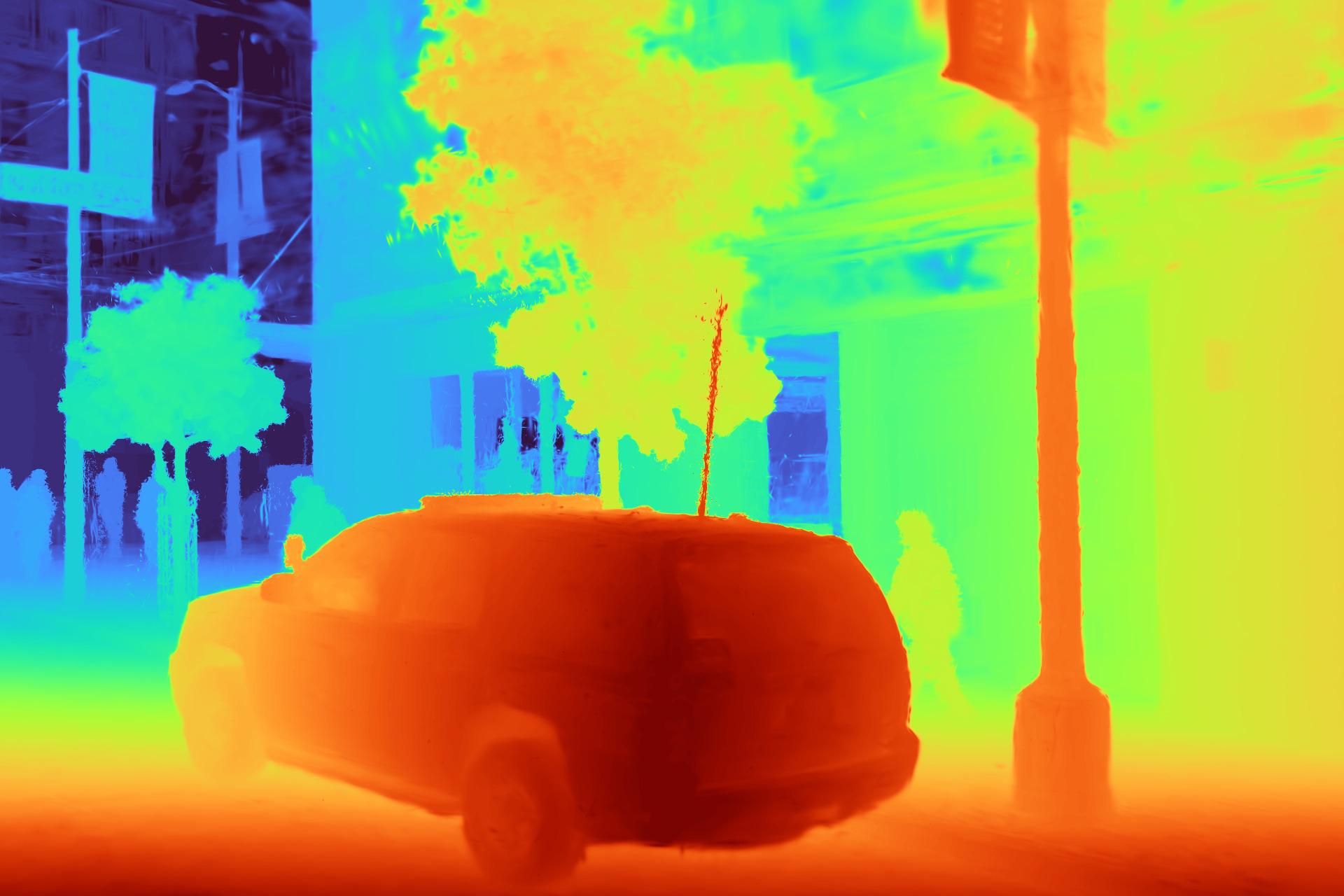} & 
             \includegraphics[width=0.245\linewidth]{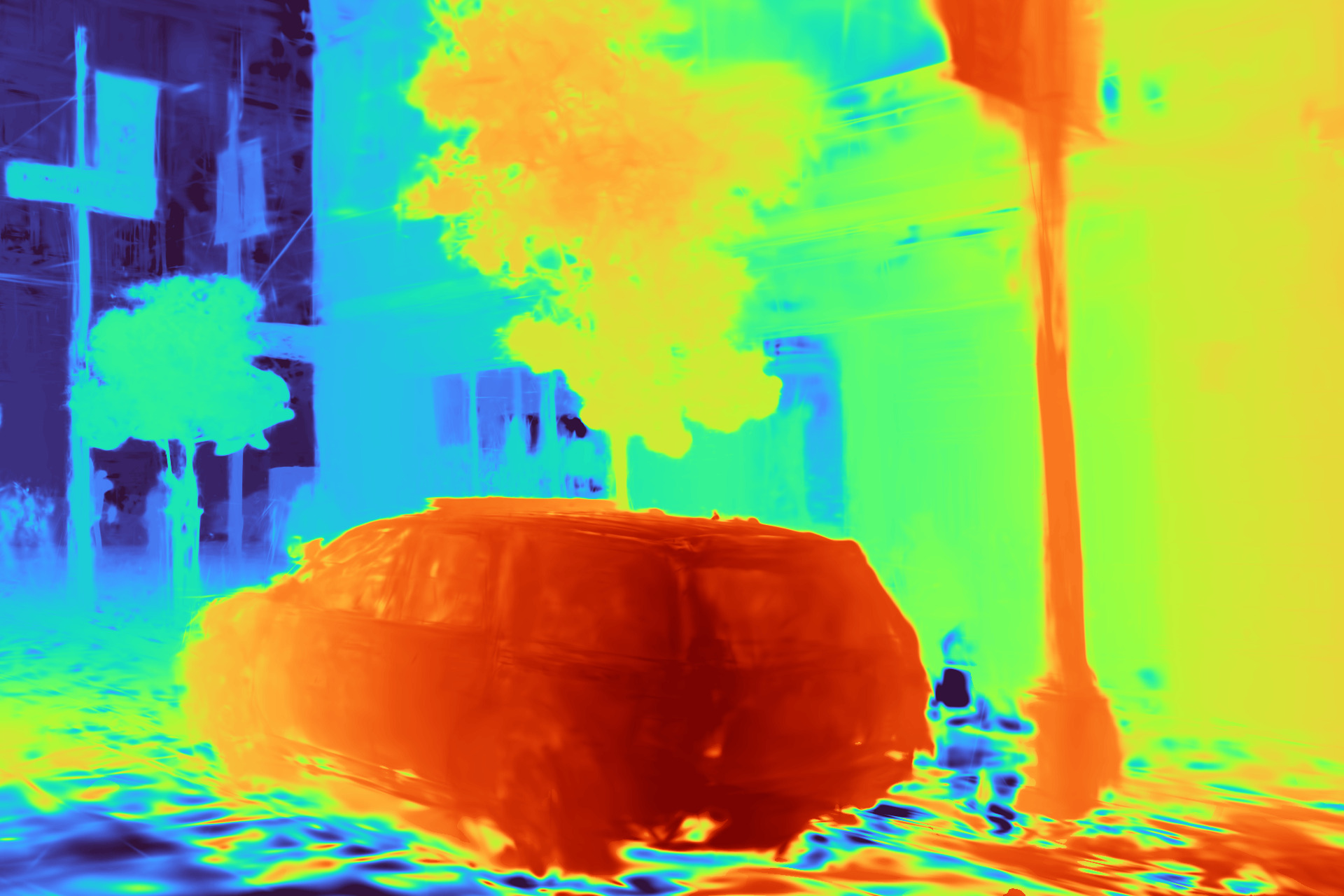} & 
             \includegraphics[width=0.245\linewidth]{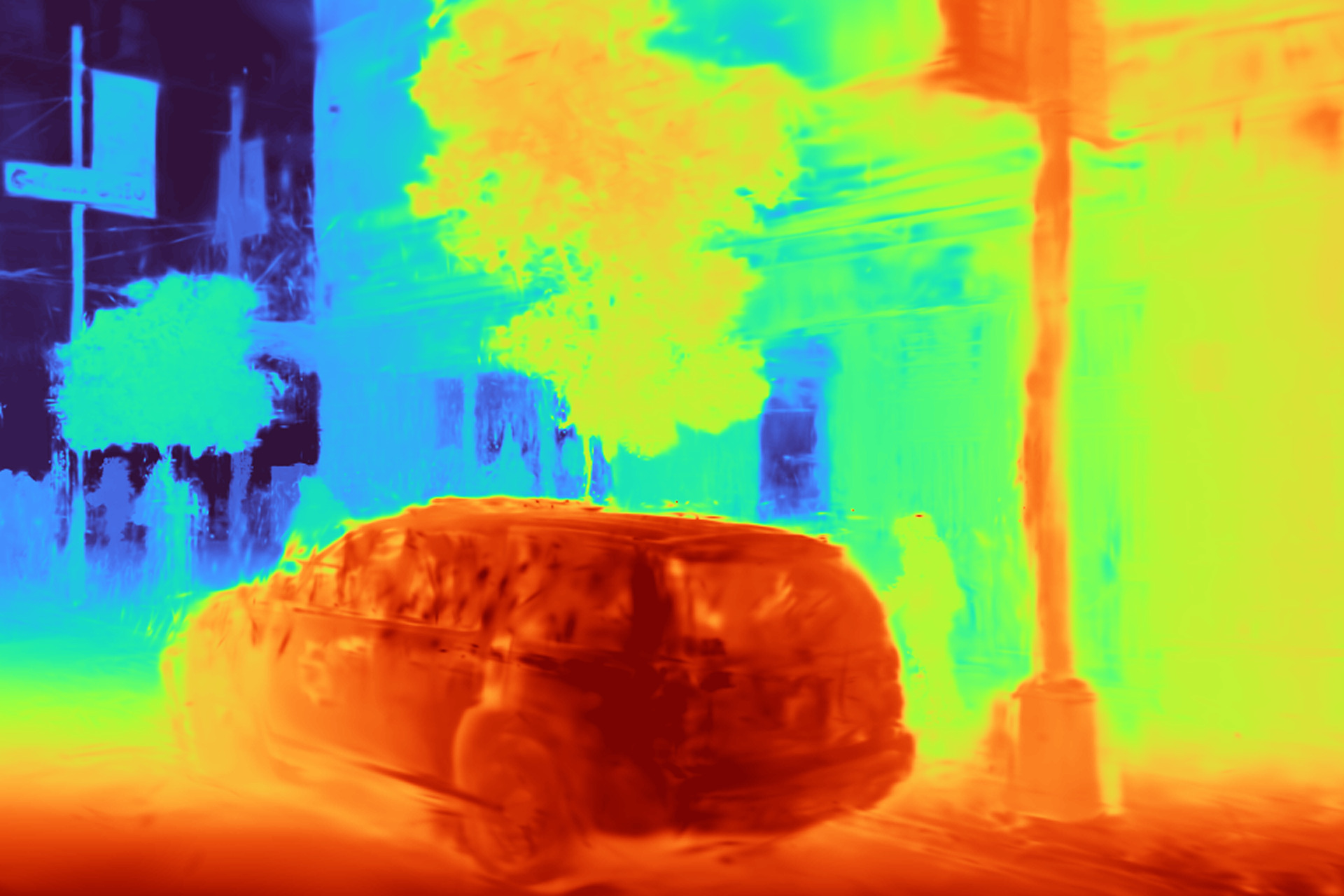} \\
    
            \includegraphics[width=0.245\linewidth]{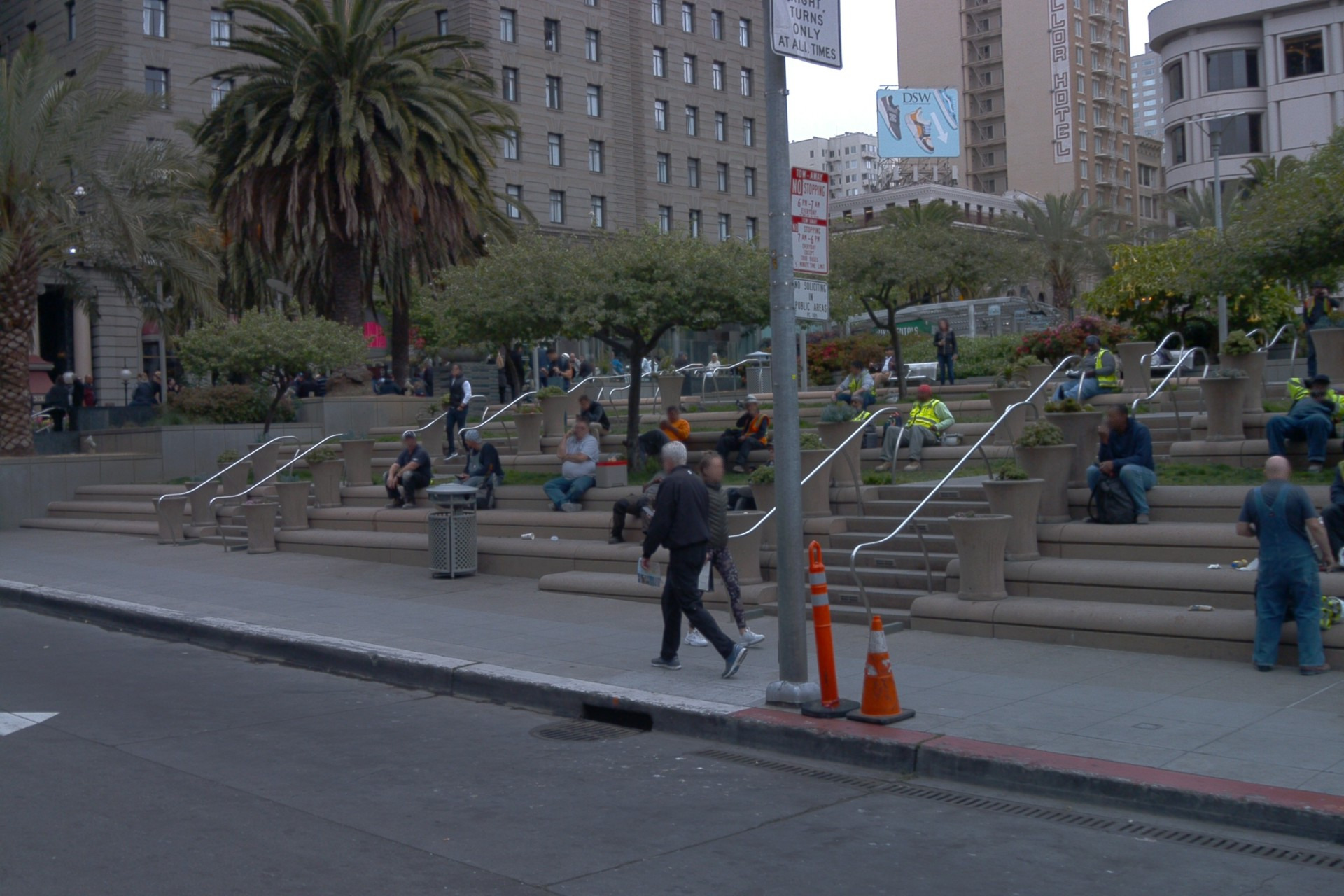} &
             \includegraphics[width=0.245\linewidth]{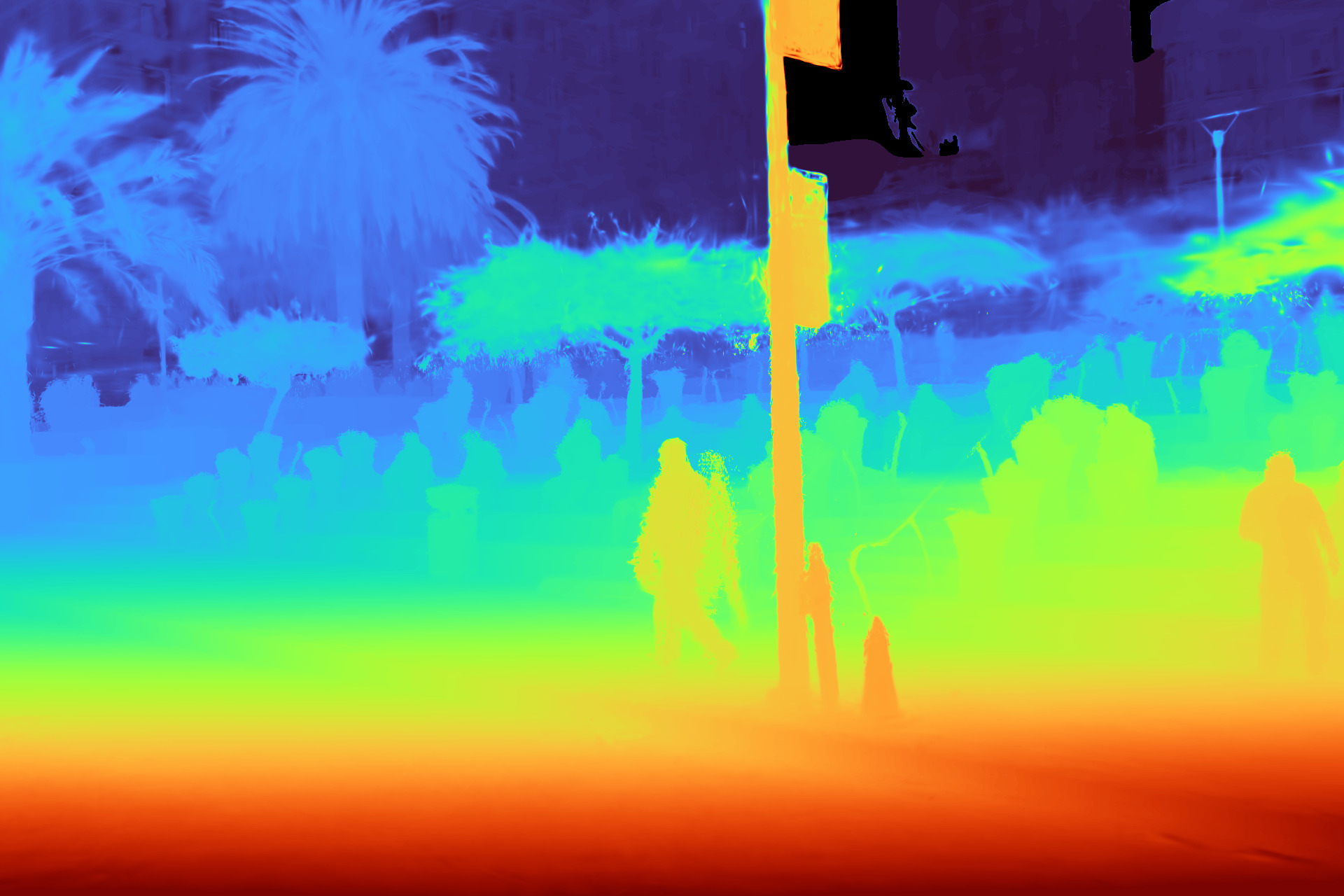} & 
             \includegraphics[width=0.245\linewidth]{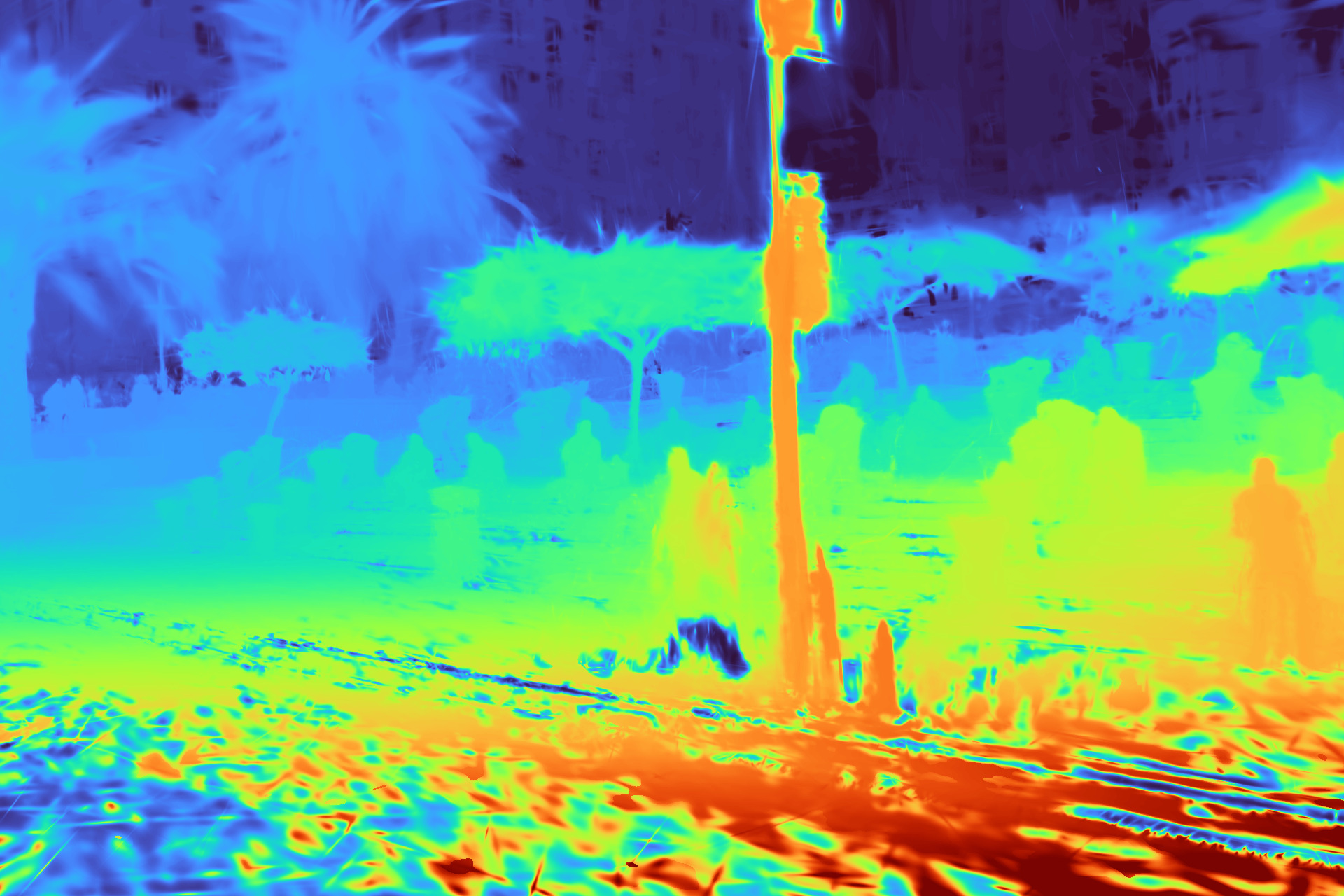} & 
             \includegraphics[width=0.245\linewidth]{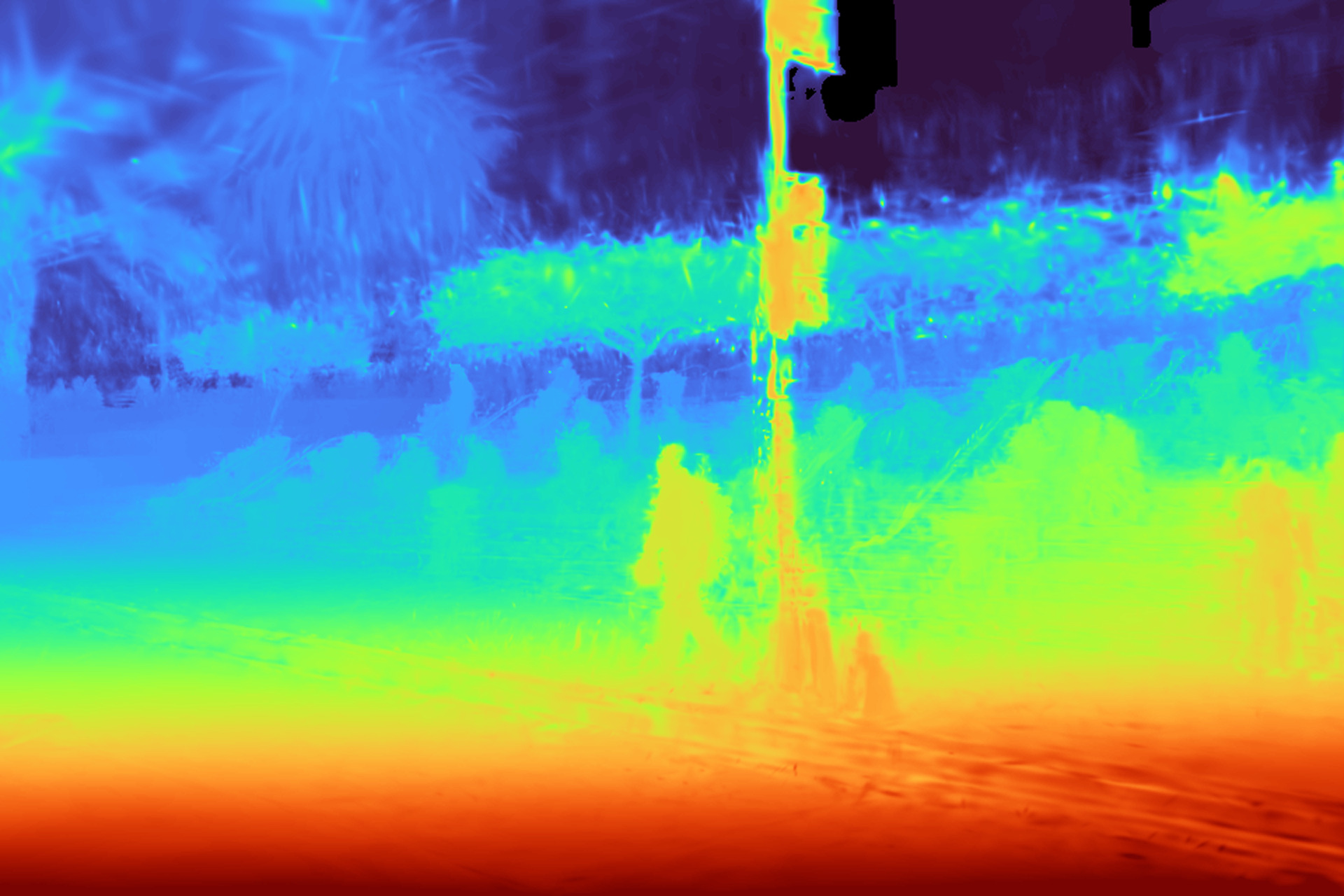} \\

        \end{tabular}
    \normalsize \caption{\normalsize \textbf{Depth map rendering. } Qualitative comparison of depth map rendering on Waymo Open Dataset. Compared to previous methods, our framework provides smooth geometric representations with high level of details.}
    \label{fig:depth}
\end{figure*}

\paragraph{3DGS and NeRF for automotive scenes.}
In automotive applications Neural Radiance Field~(NeRF)~~\cite{mildenhall2020nerf,barron2023zipnerf,mueller2022instant,wu2023mars,xie2023snerf} and 3D Gaussian Splatting~(3DGS)~\cite{kerbl20233dgs,wu20253dgut,hess2024splatad,zhang2024radegs} gathered significant attention since it provide flexible tools for photorealistic sensor modeling.
Initial research efforts~\cite{ost2021nsg,fu2022panopticnerf,kundu2022panopticnerf2,kerbl20233dgs,ml20243dgrt} emphasized the reconstruction of scene semantic and appearance guided by RGB camera images.
Later works enriched scene representation by decoupled modeling of dynamic actors and static layout leveraging either explicit scene graphs~\cite{yang2023unisim,tonderski2024neurad,yang2024emernerf,hess2024splatad,zhou2024drivinggs,yan2024streetgs,zhou2024hugs,zhou2024hugsim,jiang2025realengine} or time-dependent implicit representations~\cite{go2025splatflow,chen2023periodic,peng2025desiregs}.
Recent approaches further extended prior methods to encompass accurate sensor simulations including modeling of complex camera distortions~\cite{xie2023snerf,yang2023unisim,tonderski2024neurad,chen2025salf,zhang2024radegs,wu20253dgut,ml20243dgrt} and LiDAR~\cite{huang2023nfl,wu2024dynfl,yang2023unisim,tonderski2024neurad,chen2025salf,hess2024splatad,zhou2024lidarrt,chen2025omnire}.
3DGS provides a better trade-off between photorealism, physical accuracy and computational efficiency.
However, the conventional 3DGS based on EWA~\cite{zwicker2001ewags} formulation introduces challenges for automotive scenarios limiting the ability to render accurately complex sensors~\cite{huang2024errorgsanal}.
The promising alternative 3DGUT~\cite{wu20253dgut} provides flexibility to render arbitrary cameras without strong approximation.
In this work we introduced the sensor simulation framework based on 3DGUT~\cite{wu20253dgut}.
In contrast to previous 3DGS-based methods~\cite{hess2024splatad,chen2025omnire}, our framework provides a unified rendering formulation for rolling-shutter camera and LiDAR sensors, resulting in improved cross-sensor consistency. 

\section{Method}
\label{sec:method}

We introduce XSIM -- the sensor simulation framework for autonomous driving. Our framework is based on 3DGUT formulation that we overview in \Cref{subsec:3dgut}. In \Cref{subsec:e3dgs} we introduce extended 3D Gaussian representation to alleviate geometry and color distribution mismatches. Finally, we describe our rolling shutter sensor modeling approach along with the proposed phase mode mechanism in \Cref{subsec:senor_modeling}.

\subsection{Gaussian Splatting Preliminary}
\label{subsec:3dgut}

\paragraph{Scene representation}
\label{sec:method:gaussian_splatting}

3DGS~\cite{kerbl20233dgs,zwicker2001ewags,wu20253dgut} represents an arbitrary scene as a set of transparent Gaussian particles. Each particle is parameterized by its mean position $\vmu \in \sR^3$ and a shape encoded by a covariance matrix $\mSigma \in \sR^{3\times 3}$. The contribution of a particle is defined by a Gaussian kernel function:
\begin{equation}
    \label{eq:gaussian}
    \rho(\vx) = \sigma \exp{\bigg(-\dfrac{1}{2}(\vx - \vmu)^\top \mSigma^{-1} (\vx - \vmu)\bigg)}
\end{equation}
Covariances $\mSigma = \mR\mS\mS^\top \mR^\top$ in practice are represented and optimized via decoupled scaling vector $\vs \in \sR^3$ and rotation quaternion $\vq \in \sR^4$.
Each particle is associated with opacity value $\sigma \in \sR$, diffuse color $c_d \in \sR^3$, and view-dependent appearance feature vector $c_s \in \sR^f$. While original 3DGS~\cite{kerbl20233dgs} uses spherical harmonics for encoding view-dependent appearance and evaluate them into color before volumetric integration, recent autonomous driving simulation frameworks~\cite{tonderski2024neurad,hess2024splatad} render both $c_d$ and $c_s$ features into an image. We follow their approach, and use small trainable network consisting of few convolution layers to decode RGB color.   

Dynamic scenes are typically represented as the graph~\cite{chen2025omnire} with static and dynamic actor nodes, consisting of individual Gaussian particles. Dynamic actors are associated with their bounding boxes and an optimizable trajectory (sequence of SE3 poses). As pedestrians and cyclists represent a vulnerable road users category, we follow \cite{chen2025omnire} and represent humans as SMPL~\cite{loper2015smpl} bodies to improve reconstruction quality and provide better controllability. 

\paragraph{Unscented transform.}
\label{sec:method:ray_oriented_rendering}
A key stage that enables efficient rendering in Gaussian splatting is tiling, in which particles are assigned to screen-space tiles based on the 2D projection of their shape. The EWA splatting formulation relies on a linearized projection approximation based on a single point, which makes rendering highly non-linear cameras (e.g. with rolling shutter or optical distortions) challenging. Alternatively, 3DGUT projects particles onto an arbitrary camera via Unscented Transform (UT). Given the mean and covariance of a particle in 3D, UT constructs a set of sigma points, which are individually projected through the camera model. The resulting 2D Gaussian conics are then approximated by weighting projected points. As a result, UT enables tiling for a wide range of complex camera models without requiring any modifications. Finally, particles assigned to each tile are sorted by depth to ensure correct ordering during volumetric rendering.

\paragraph{Volumetric integration.} 

The final stage of rendering is rasterization, which performs volumetric integration. In EWA splatting, 2D Gaussian conics are used explicitly to compute particle responses during volumetric rendering, introducing projection approximation errors into the rendering process. In contrast, 3DGUT uses 2D conics only for tiling, while volumetric rendering is performed by computing particle responses directly in 3D space. Specifically, for a given camera ray with origin $\vo \in \sR^3$ and direction $\vd \in \sR^3$, the particle is evaluated at the point of maximum response along the ray:

\begin{equation}
    \label{eq:max_response}
    \vx_{\text{max}} = \vo + \tau_{\text{max}}\vd, \quad \tau_{\text{max}} = \dfrac{\vd^\top \mSigma^{-1}(\vmu - \vo)}{\vd^\top \mSigma^{-1} \vd}
\end{equation}

Given the maximum response point, the overall volumetric integration follows standard formulation with $\alpha_i=\rho_i(\vx_{\text{max}})$:
\begin{equation}
    \label{eq:blend}
    T_i = \prod_{j<i}(1-\alpha_j), \quad \vc = \sum_i \vc_i\alpha_i T_i 
\end{equation}

Similarly, for depth and range image rendering we use the same equation as above, but replace particle colors $\vc_i$ with maximum response $\tau_{\text{max}}$ distances along the ray. 

\begin{figure}
    \centering
    \includegraphics[width=1.0\linewidth]{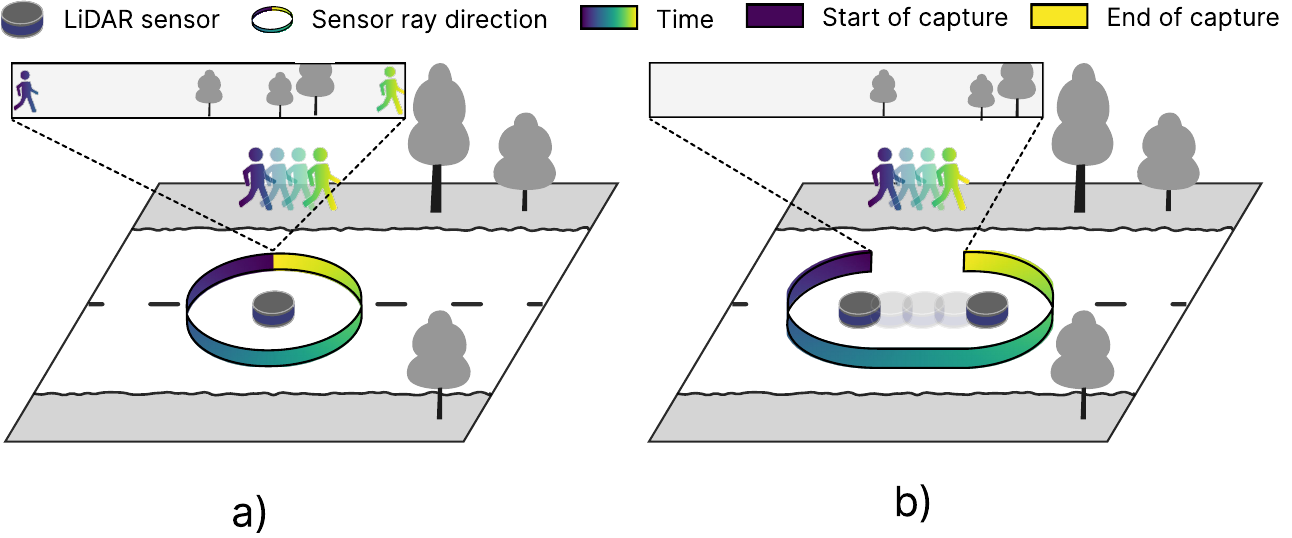}
    \caption{LiDAR discontinuities occurring near azimuth border. a) Even when sensor is stationary and covers exactly 360 degrees, time discontinuity due to the rolling shutter may lead to objects observed twice. b) Ego movement in combination with rolling shutter leads to spatial discontinuities}
    \label{fig:lidar-discontinuity}
\end{figure}

\subsection{Extended 3D Gaussian Representation}
\label{subsec:e3dgs}

While using geometric guidance is generally beneficial for 3DGS scene reconstuction, in some cases the two sensing modalities may impose different requirements on opacity modeling. Accurate geometric modelling requires representing surfaces with precisely located  opaque gaussians, as guided by LiDAR measurements. In contrast, appearance modelling must account for view-dependent effects such as specular reflections and translucency, which often require multiple semi-transparent Gaussians along a viewing ray. Furthermore, surface opacity can be wavelength-dependent, as materials such as glass exhibit different transparency properties for LiDAR and visible light. To accommodate these effects, we augment each Gaussian with separate opacity parameters: $\sigma_c$ for camera and $\sigma_L$ for LiDAR rendering, which are jointly optimized within the unified representation. These two opacity values are then regularized during optimization to ensure consistency. We experimentally demonstrate that this extended representation effectively resolves distribution mismatches and improves both camera and LiDAR modelling for the novel-view synthesis problem. 

\subsection{Sensor Modeling}
\label{subsec:senor_modeling}

Autonomous driving perception and planning algorithms primarily rely on two sensor modalities: cameras and LiDARs. Both sensors typically operate in rolling shutter mode, in which sensor readings are acquired sequentially over time in a row-by-row fashion. As highlited by multiple previous works~\cite{tonderski2024neurad,hess2024splatad}, due to high velocities experienced in autonomous driving scenarios, modeling rolling shutter is essential for accurate reconstruction. Whereas previous work~\cite{hess2024splatad} addressed rolling-shutter effects using separate, sensor-specific models, we describe generalized rolling-shutter modeling approach. We also identify spherical cameras (i.e. LiDARs) as an edge case of 3DGUT approach and propose phase modeling mechanism to mitigate arising rendering issues. 

\begin{figure}
    \centering

    \setlength{\tabcolsep}{1.0pt}
    \begin{tabular}{ccc}
        Ground-truth & Ours, With LO & Ours, Without LO \\
        \includegraphics[width=0.33\linewidth,trim={5cm 0 0 0},clip]{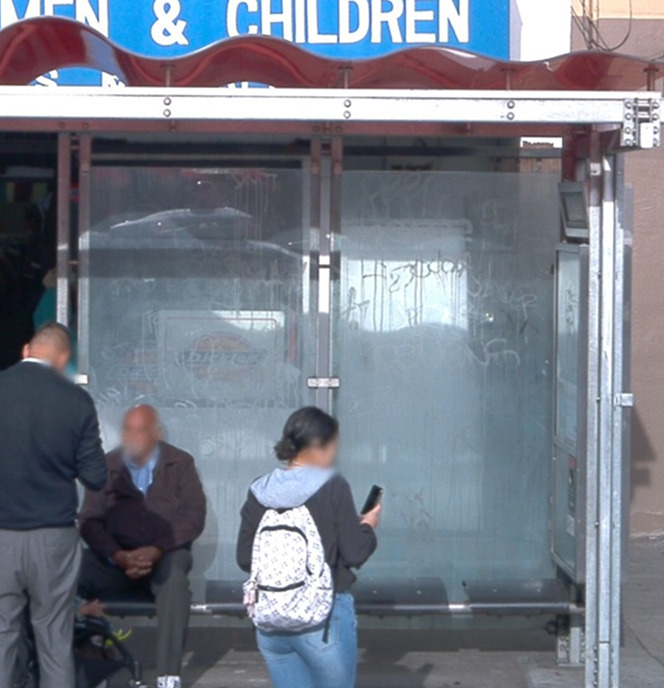} &
        \includegraphics[width=0.33\linewidth,trim={5cm 0 0 0},clip]{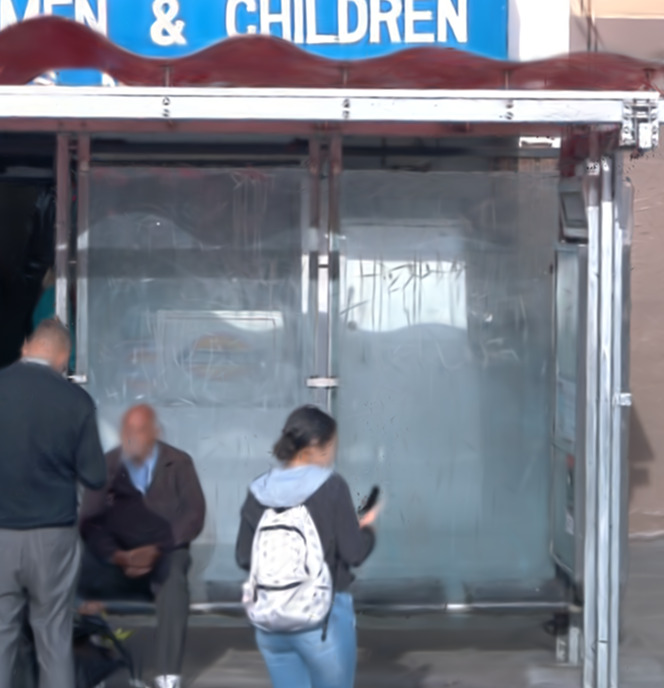} & 
        \includegraphics[width=0.33\linewidth,trim={5cm 0 0 0},clip]{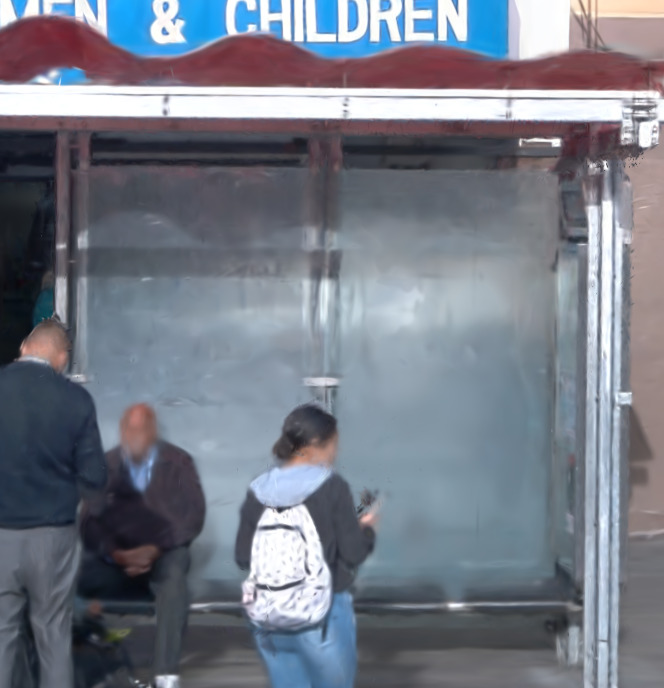}\\
    \end{tabular}
    \caption{Modeling LiDAR opacity (LO) separately resolves geometry and color distributions mismatch, and increases quality of appearance modeling for translucent surfaces and specular reflections. }
    \label{fig:lidar-density}
\end{figure}
\begin{figure*}[!htbp]
    \centering
    \includegraphics[width=1.0\linewidth, trim={0cm 0cm 0cm 0.0cm},clip]{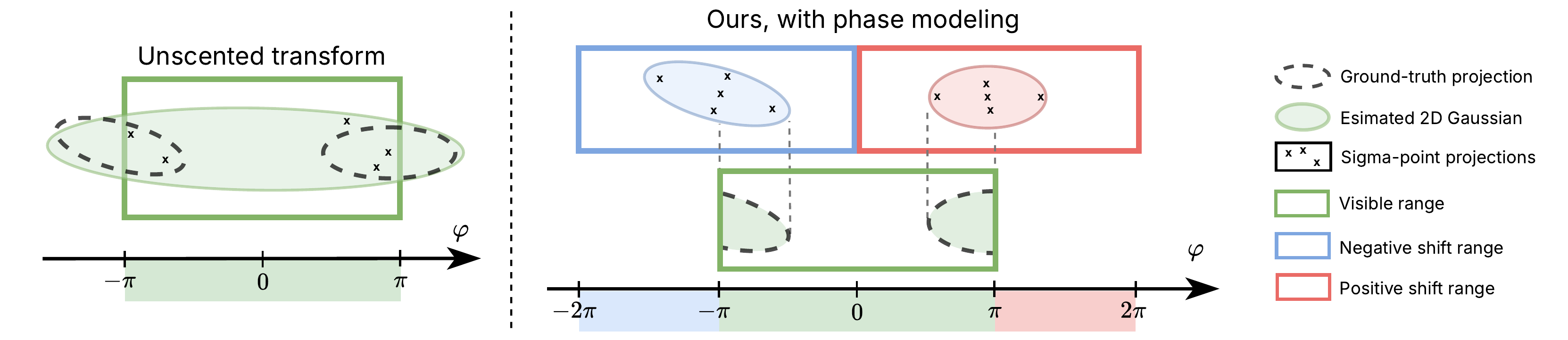}
    \caption{
    Single Gaussian particle spanning the azimuth boundaries ($\varphi= \pm\pi$) of rolling-shutter spherical camera projects into two separate 2D gaussians with different 2D covariances. Unscented Transform provides unimodal posterior approximation of particle projection, leading to a overly large projections with incorrect shapes. Our phase modeling mechanism enables bimodal 2D gaussian projection by considering two additional projections shifted by $\pm \pi$ from visible range. By performing projections shifted by half period, we handle particles which sigma points fall into multiple azimuth periods. 
    }
    \label{fig:phm}

\end{figure*}

\paragraph{General rolling shutter modeling.}
\label{subsec:sm:rolling_shutter}

Due to the rolling-shutter mechanism, sensor readings are not captured instantaneously, and the observation time varies across pixels.
Consequently, each image-space point $(u, v)$ is associated with a distinct capture time $\tau(u,v)$. While in practice the rolling shutter time is linear and aligned with either horizontal or vertical image axis, we can express it as:

\begin{equation}
    \label{eq:rolling-time}
	\tau(u, v) = \tau_{\text{start}} + u \tau_u + v \tau_v
\end{equation}
where $\tau_u, \tau_v$ define the scan direction and speed, and denote middle of exposure time as $\tau_{\text{mid}} = \tau(0.5, 0.5)$.

As time progresses during image acquisition, both the sensor and the scene evolve. We assume that, over the duration of capture, the camera and all dynamic actors move with constant linear and angular velocities. Under this assumption, the position of a 3D world point $\vx_w \in \sR^3$ at time $\eta$ is given by:

\begin{equation}
	\vx_w(\eta) = \vx_w(\tau_\text{mid}) + (\vv_a + \vw_a \times \vr) \eta
\end{equation}
where $\vv_a \in \sR^3$ and $\vw_a \in \sR^3$ denote linear and angular velocities of the dynamic actor, respectively, and $\vr \in \sR^3$ is radius vector defined by point position in actor coordinates.

Similarly, the camera pose evolves over time according to constant-velocity motion. Let $\vq(\eta) \in \sR^4$ and $\vt(\eta) \in \sR^3$ denote the camera orientation and position at time $\eta$. Camera movement is then modeled as: 

\begin{equation}
\label{eq:proj_cam_move}
	\vq(\eta) = e^{\vw_c \eta / 2} \otimes \vq(\tau_\text{mid}), \quad \vt(\eta) = \vt(\tau_\text{mid}) + \eta\vv_c
\end{equation}
where $\vv_c$ and $\vw_c$ are the camera’s linear and angular velocities, and $\otimes$ denotes the Hamilton product. Given the arbitrary static camera projection function $\pi(\cdot): \sR^3 \xrightarrow{} [0, 1]^2$ and known point time $\eta$ projection of world point $x_w \in \sR^3$ into image coordinates $(u, v)$ has the closed form: 

\begin{equation}
    \label{eq:proj_uv}
    \begin{aligned}
    \vx_c(\eta) &= \vq^{-1}(\eta) \otimes (\vx_w(\eta) - \vt(\eta)) \otimes \vq(\eta), \\
    u(\eta),v(\eta) &= \pi(\vx_c(\eta)), 
    \end{aligned}
\end{equation}

However, since the observation time $\eta$ of the point is typically not known prior to projection, the projection problem formulates as estimating world time $\eta$ which is consistent with image space point observation time: 

\begin{equation}
\label{eq:proj_problem}
\eta = \tau(u(\eta), v(\eta))
\end{equation}

Although this dependency is nonlinear and does not admit a closed-form solution, it can be efficiently solved using iterative methods. Following~\cite{sun2020waymoopen}, we use the Newton–Raphson method and observe that only a few iterations are sufficient in practice.

\paragraph{Phase modeling.}
\label{subec:sm:phase_modeling}

Our framework builds upon 3DGUT splatting, which leverages the Unscented Transform (UT) to approximate the projection of a 3D Gaussian. The UT performs a posterior approximation under the assumption that the projection of a 3D Gaussian can be adequately represented by a single 2D Gaussian. While this assumption holds in many practical settings, it breaks down in scenarios involving spherical rolling-shutter sensors. In particular, spinning spherical LiDAR sensors may observe a single Gaussian particle multiple times when it spans the azimuthal boundary or ignore it depending on the world dynamics (see Fig.~\ref{fig:lidar-discontinuity}). Moreover, this can lead to a same particle projecting into two separate shapes near boundaries with different covariances and depths (see ground-truth projections in Fig. \ref{fig:phm}). 
In such cases, the projected distribution becomes multimodal, violating the assumptions inherent to the UT. 

To address this limitation, we introduce a phase modeling mechanism for spherical camera rendering. Under the spherical projection, the azimuth  is a periodic function defined as:
\begin{equation}
    \phi = \atantwo(y,x) + 2\pi k, \quad k \in \sZ    
\end{equation}

While conventional static spherical mappings retain only the principal solution ($k=0$), we explicitly account for phase wrapping by considering $k\in\{-1,0,+1\}$. Since additional projections may differ in covariance and depth to camera, in addition to the standard central projection, we perform auxiliary projections shifted by $\pm\pi$ (see Fig. \ref{fig:phm}). For each interval, we constrain projections of sigma points into it by initializing $\tau$ in Eq. \ref{eq:proj_problem} with an interval middle of exposure time, and shift projected point azimuths by $2\pi k$ if they fall outside of interval. 2D Gaussian conics and depth are individually estimated for each interval based on projected sigma points, and valid projections are passed to the next tiling stage. Compared to original 3DGUT projection, our mechanism results in accurate tiling with no false tile-particle intersections and less depth sorting errors for particles near azimuth boundaries. 

\renewcommand{\arraystretch}{1.2}
\begin{table*}[!ht]
\centering
\begin{tabular}{clcrrrrrrrr} 
\toprule
\multirow{2}{*}{Dataset} & \multirow{2}{*}{Method} & \multirow{2}{*}{Conference} & \multicolumn{4}{c}{Reconstruction} & \multicolumn{4}{c}{Novel-view synthesis} \\ \cmidrule(lr){4-7} \cmidrule(lr){8-11}
                                                         &                         &                             & \multicolumn{1}{c}{PSNR$\uparrow$} & \multicolumn{1}{c}{SSIM$\uparrow$} & \multicolumn{1}{c}{LPIPS$\downarrow$} & \multicolumn{1}{c}{CD$\downarrow$} & \multicolumn{1}{c}{PSNR$\uparrow$} & \multicolumn{1}{c}{SSIM$\uparrow$} & \multicolumn{1}{c}{LPIPS$\downarrow$} & \multicolumn{1}{c}{CD$\downarrow$}\\ \midrule
\multirow{7}{*}{\begin{tabular}[c]{@{}c@{}}Waymo\\ (12 scenes)\end{tabular}} & PVG                     & Arxiv23                     & 25,02                    & 0,8005                   & 0,4408                    & 82,11                   & 24,29                    & 0,7864                   & 0,4451                    & 68,27                  \\
                                                                             & StreetGS                & ECCV24                      & 25,45                    & 0,8123                   & 0,3111                    & 16,16                   & 24,41                    & 0,7827                   & 0,3198                    & 16,87                  \\
                                                                             & OmniRE                  & ICLR25                      & 25,94                    & 0,8159                   & 0,3049                    & 15,68                   & 24,60                    & 0,7765                   & 0,3207                    & 13,76                  \\
                                                                             & HUGS                  & CVPR24                      & 26,90                    & 0,8513                   & 0,3351                    & 44,58                   & 25,95                    & 0,8267                   & 0,3337                    & 37,60                  \\
                                                                             & EmerNerf                & ICLR24                      & 27,15                    & 0,8056                   & 0,4620                    & \underline{0,71}                    & 26,12                    & 0,7962                   & 0,4575                    & 2,54                   \\
                                                                             & SplatAD                 & CVPR25                      & \underline{27,74}                    & \underline{0,8650}                   & \underline{0,2807}                    & 0,82                    & \underline{27,06}                    & \underline{0,8492}                   & \underline{0,2807}                    & \underline{0,82}                   \\
                                                                             & \textbf{XSIM, Ours}              & \textbf{--}                            & \textbf{30,75}           & \textbf{0,9030}          & \textbf{0,2228}           & \textbf{0,08}           & \textbf{29,80}           & \textbf{0,8904}          & \textbf{0,2236}           & \textbf{0,18} \\ \midrule  
\multirow{7}{*}{\begin{tabular}[c]{@{}c@{}}Argoverse \\ (10 scenes)\end{tabular}} & PVG                     & Arxiv23                     & 23,78                    & 0,7164                   & 0,4840                    & 31,36                   & 22,93                    & 0,7031                   & 0,4908                    & 27,20                  \\
                                                                                  & StreetGS                & ECCV24                      & 23,85                    & 0,7223                   & 0,3824                    & 21,62                   & 22,37                    & 0,6975                   & 0,3806                    & 21,01                  \\
                                                                                  & OmniRE                  & ICLR25                      & 23,97                    & 0,7230                   & 0,3822                    & 21,87                   & 22,44                    & 0,6975                   & 0,3815                    & 22,02                  \\
                                                                                  & UniSim                  & CVPR23                      & 23,04                    & 0,6697                   & 0,3986                    & 25,58                   & 23,06                    & 0,6694                   & 0,3962                    & 26,24                  \\
                                                                                  & NeuRad                  & CVPR24                      & 26,46                    & 0,7271                   & 0,3045                    & {\ul 2,43}              & 26,49                    & 0,7271                   & 0,3044                    & {\ul 2,63}             \\
                                                                                  & SplatAD                 & CVPR25                      & {\ul 28,71}              & {\ul 0,8333}             & {\ul 0,2653}              & 2,78                    & {\ul 28,40}              & {\ul 0,8258}             & {\ul 0,2706}              & 2,68                   \\
                                                                                  & \textbf{XSIM, Ours}              & \textbf{--}                 & \textbf{29,44}           & \textbf{0,8431}          & \textbf{0,2563}           & \textbf{0,57}           & \textbf{29,44}           & \textbf{0,8423}          & \textbf{0,2514}           & \textbf{1,26} \\ \midrule 
\multirow{7}{*}{\begin{tabular}[c]{@{}c@{}}Pandaset\\ (10 scenes)\end{tabular}}   & PVG                     & Arxiv23                     & 23,62                    & 0,7066                   & 0,4405                    & 101,33                  & 22,81                    & 0,6885                   & 0,4537                    & 121,54                 \\
                                                                                  & StreetGS                & ECCV24                      & 23,70                    & 0,7192                   & 0,3206                    & 18,68                   & 22,53                    & 0,6866                   & 0,3249                    & 19,90                  \\
                                                                                  & OmniRE                  & ICLR25                      & 23,73                    & 0,7196                   & 0,3246                    & 21,02                   & 22,58                    & 0,6884                   & 0,3262                    & 18,99                  \\
                                                                                  & UniSim                  & CVPR23                      & 23,62                    & 0,6953                   & 0,3291                    & 10,46                   & 23,45                    & 0,6910                   & 0,3300                    & 9,68                   \\
                                                                                  & NeuRad                  & CVPR24                      & 26,54                    & 0,7675                   & 0,2386                    & {\ul 1,45}              & 26,05                    & 0,7589                   & 0,2418                    & {\ul 1,65}             \\
                                                                                  & SplatAD                 & CVPR25                      & {\ul 28,69}              & {\ul 0,8759}             & \textbf{0,1853}           & 1,54                    & {\ul 26,77}              & {\ul 0,8044}             & \textbf{0,1904}           & 1,69                   \\
                                                                                  & \textbf{XSIM, Ours}              & \textbf{--}                 & \textbf{29,05}           & \textbf{0,8839}          & {\ul 0,1872}              & \textbf{0,20}           & \textbf{27,00}           & \textbf{0,8055}          & {\ul 0,1944}              & \textbf{1,23}   \\                                                                               
\bottomrule

\end{tabular}
\caption{Quantitative results on three datasets under scene reconstruction and novel-view synthesis scenarios. We report RGB image quality metrics (PSNR, SSIM, LPIPS) and LiDAR reconstruction accuracy measured by Chamfer Distance (CD). Our framework achieves state-of-the-art performance across all datasets and scenarios, with particularly large error reductions for LiDAR rendering. On PandaSet, LPIPS remains competitive and ranks second, with a minor gap to the best-performing method.}
\label{tab:all-datasets}
\end{table*}

\subsection{Camera and LiDAR supervision}
\label{subsec:ssd}
Our scene representation consisting of multiple object nodes is optimized simultaneously from driving logs by randomly sampling images and closest by time LiDAR sweeps at each iteration. We supervise it using combination of losses:
\begin{equation}
\mathcal{L} = \underbrace{\lambda \mathcal{L}_1 + (1- \lambda)L_{\text{SSIM}}}_{\text{camera guidance}} + \underbrace{\mathcal{L}_{\text{depth}}}_{\text{LiDAR}} + \mathcal{L}_{\text{opacity}} + \mathcal{L}_{reg}
\end{equation}
Following common practice we set $\lambda = 0.2$. LiDAR guidance loss is $\mathcal{L}_1$ loss between rendered and ground-truth ray lengths. To ensure consistency between optimized opacities we impose $L_{\text{opacity}} = \sum_i | \sigma_{c,i} - \sigma_{L,i} |$ regularization. Details on other regularization terms $\mathcal{L}_{reg}$ are listed in appendix. 

\section{Experiments}
\label{sec:exps}

\paragraph{Implementation details.} 
We implement camera and LiDAR rendering using custom CUDA kernels. A unified rendering pipeline allows sharing rasterization forward and backward passes across all camera models. To handle non-uniform LiDAR beam angles during tiling, particle assignment iterates over elevation tile boundaries~\cite{hess2024splatad}. Human modeling follows OmniRE~\cite{chen2025omnire}, using their deformable and SMPL-based scene nodes. We adopt the specular Gaussian configuration and CNN post-processing from \cite{hess2024splatad,tonderski2024neurad}. Hyperparameters largely follow SplatAD~\cite{hess2024splatad}, while Gaussian splitting and densification use the 3DGUT~\cite{wu20253dgut} strategy with minor modifications. Full hyperparameter details are provided in the appendix.

\paragraph{Datasets.}  To evaluate our framework we use three popular datasets -- Waymo Open Dataset~\cite{sun2020waymoopen}, Argoverse2~\cite{Argoverse2} and PandaSet~\cite{xiao2021pandaset}.
Following OmniRE~\cite{chen2025omnire}, we use 12 scenes from Waymo, featuring ego-vehicle motion, dynamic and diverse classes of vehicles and pedestrians.
As for Argoverse2 and PandaSet we adopt SplatAD~\cite{hess2024splatad} partition without modifications. Both training and evaluation are performed using full-resolution images and point clouds.

\paragraph{Baselines.} For experimental evaluation we choose a wide range of baselines, featuring both recent NeRF-based methods (UniSim~\cite{yang2023unisim}, NeuRAD~\cite{tonderski2024neurad},  EmerNerf~\cite{yang2024emernerf}) and 3DGS-based PVG~\cite{chen2023periodic}, StreetGaussians~\cite{yan2024streetgs}, OmniRe~\cite{chen2025omnire}, HUGS~\cite{zhou2024hugs,zhou2024hugsim} and SplatAD~\cite{hess2024splatad}. For PVG, StreetGaussians and OmniRe we use implementation based on $\texttt{drivestudio}$. As UniSim has no official repository, we use reimplementation by $\texttt{neurad-studio}$. 

\paragraph{Evaluation metrics.} For rendered image quality assessment we use standard novel-view synthesis metrics -- PSNR $\uparrow$, SSIM $\uparrow$, and LPIPS$\downarrow$. To measure LiDAR simulation quality we compute Chamfer Distance (CD$\downarrow$) metric between rendered and ground-truth point clouds. 

\subsection{Image rendering}

\begin{figure*}[!t]
    \centering
        \setlength{\tabcolsep}{1pt}
        \Large
        \begin{tabular}{cccc}
            Ground-truth & \textbf{XSIM, Ours} & SplatAD & OmniRE \\
            
            \includegraphics[width=0.245\linewidth]{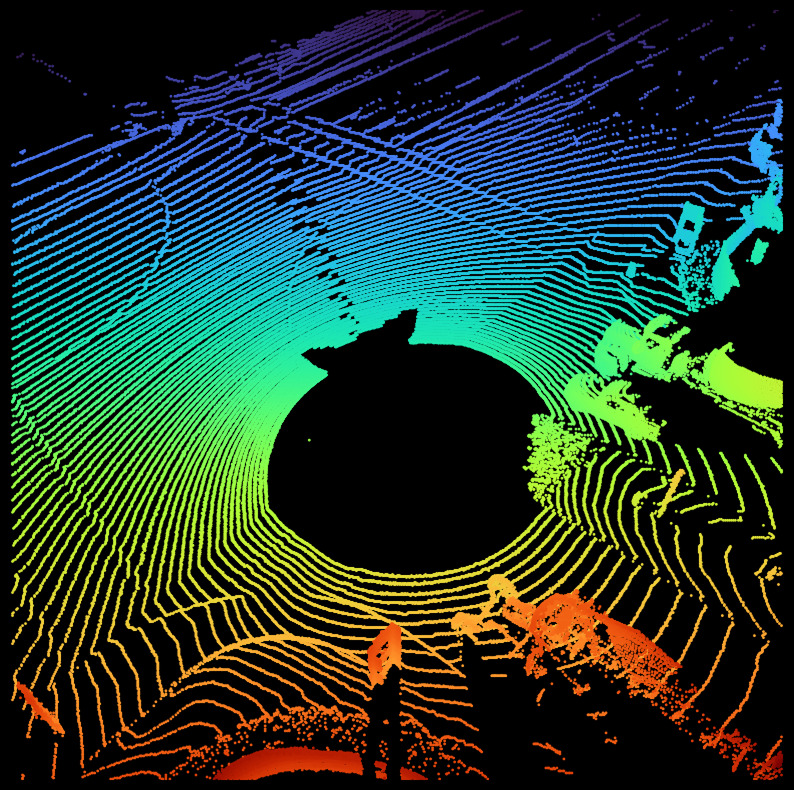} &
             \includegraphics[width=0.245\linewidth]{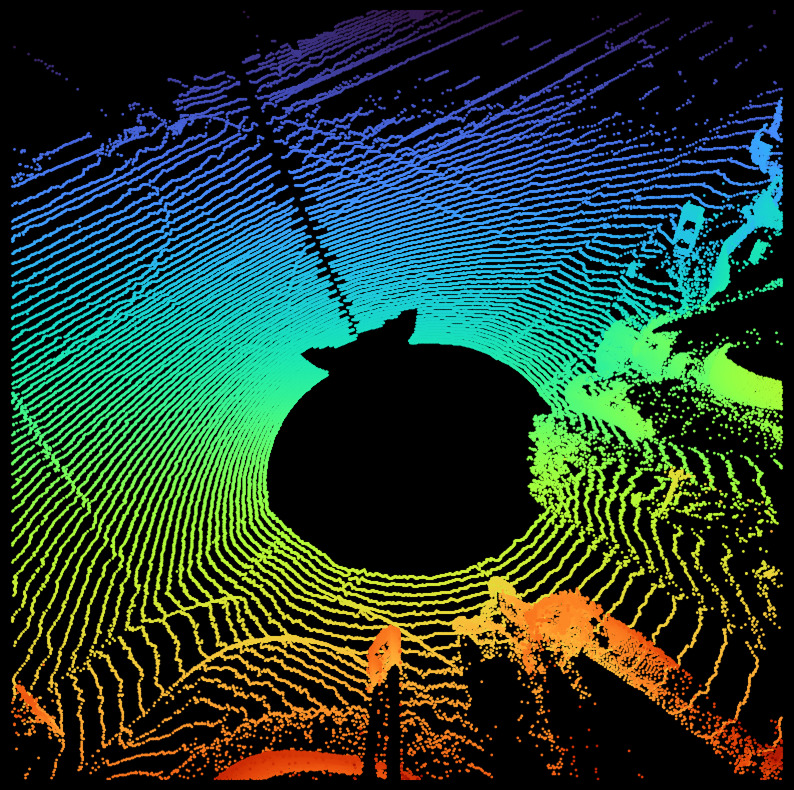} & 
             \includegraphics[width=0.245\linewidth]{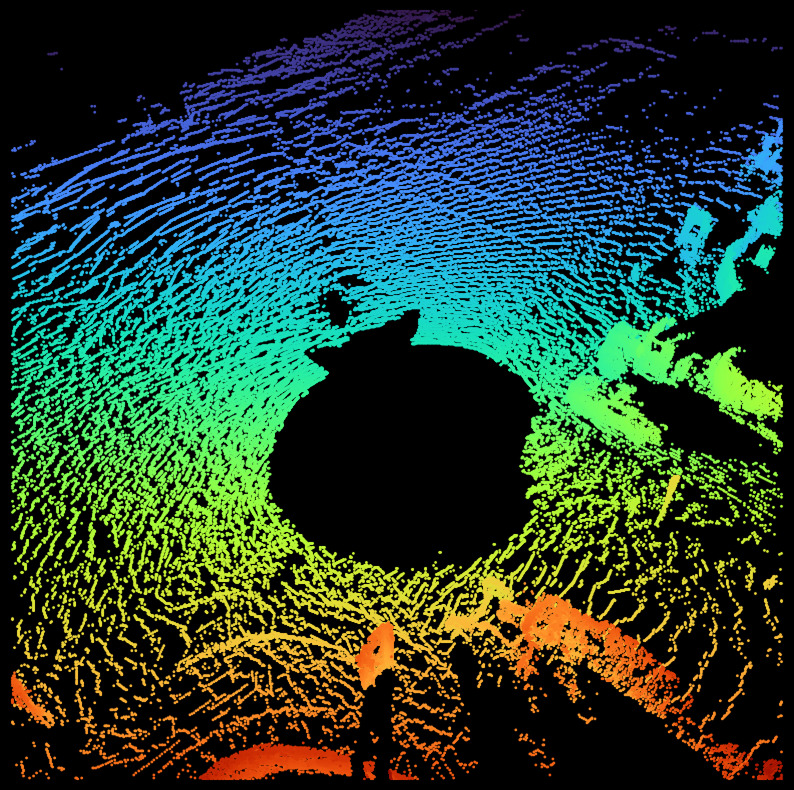} & 
             \includegraphics[width=0.245\linewidth]{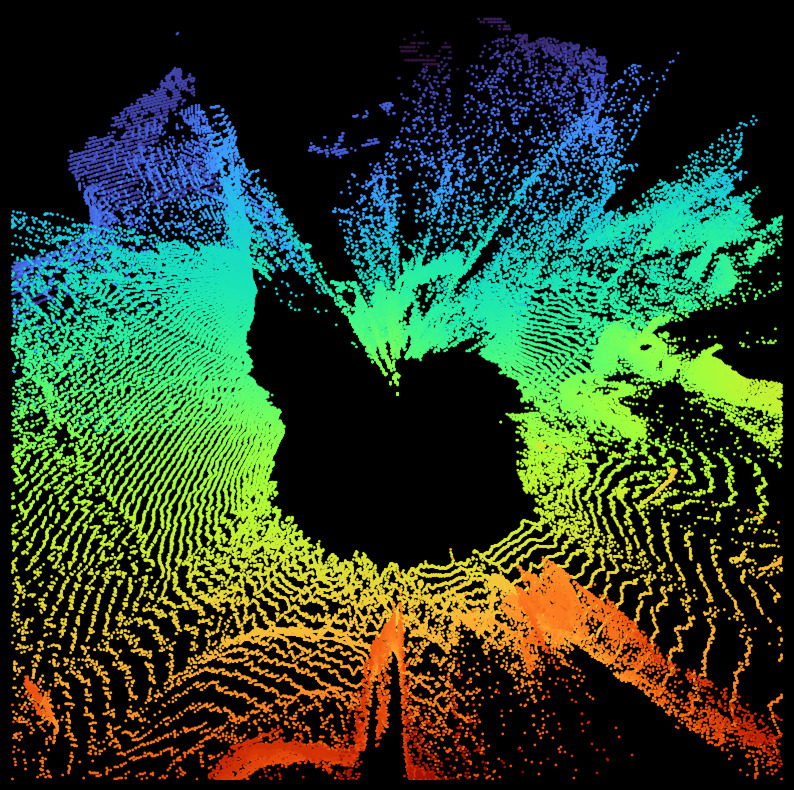} \\
    
            \includegraphics[width=0.245\linewidth]{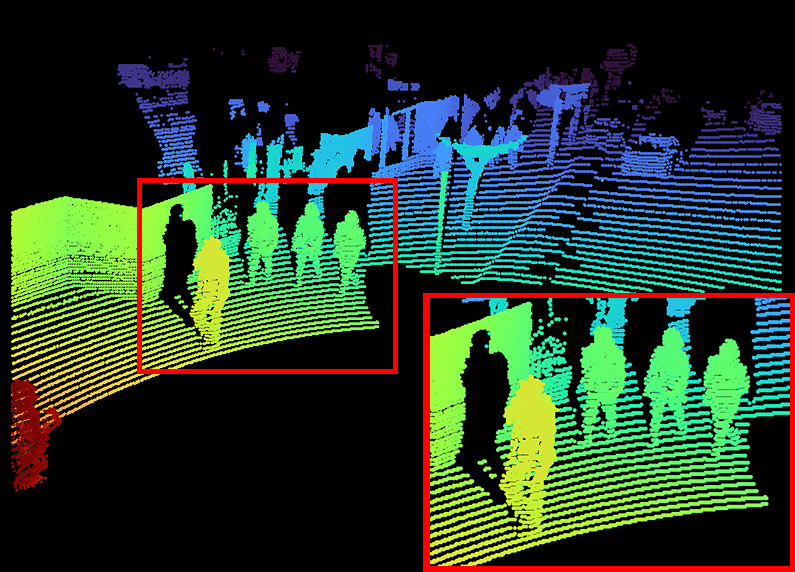} &
             \includegraphics[width=0.245\linewidth]{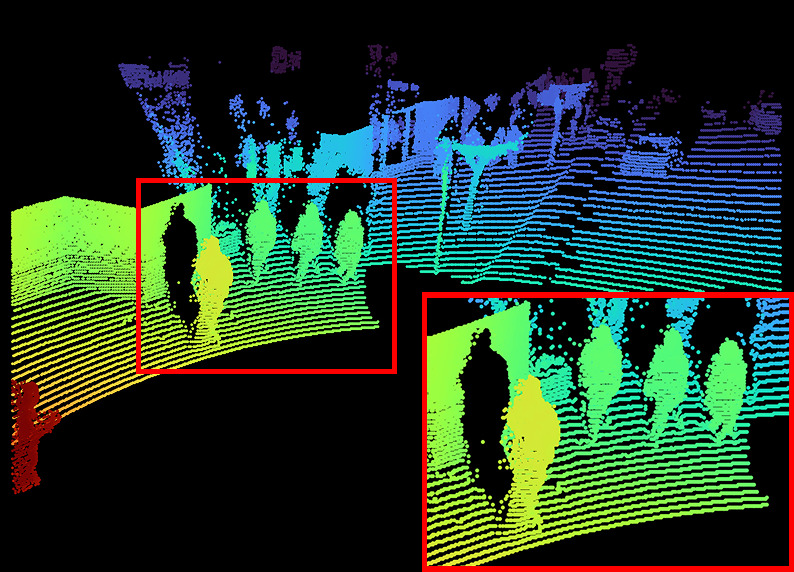} & 
             \includegraphics[width=0.245\linewidth]{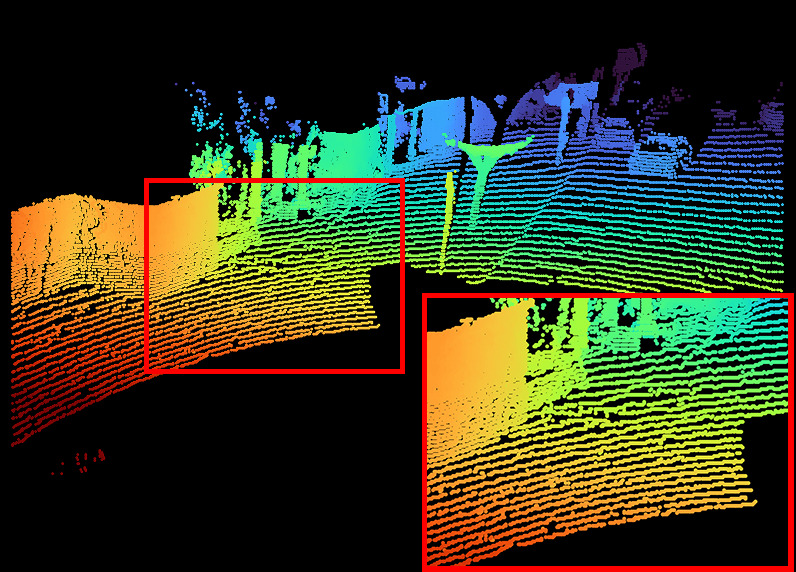} & 
             \includegraphics[width=0.245\linewidth]{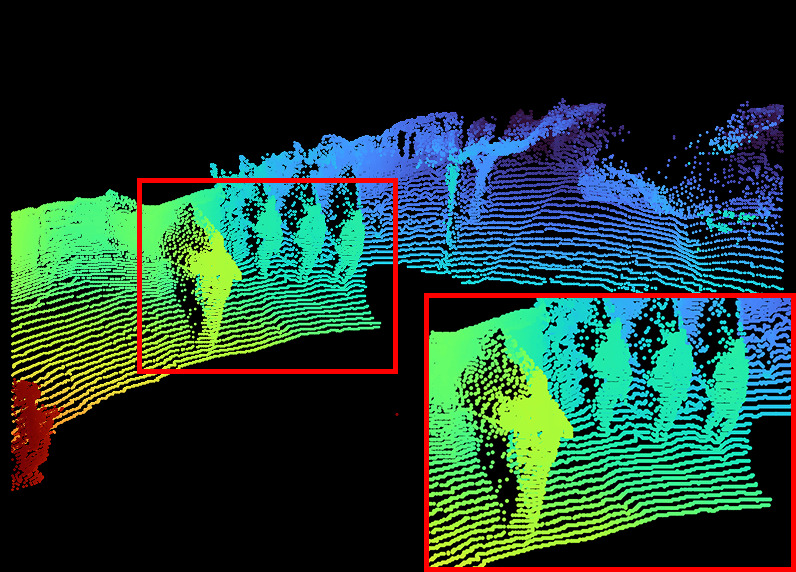} \\

    \end{tabular}
        
    \caption{\textbf{LiDAR Rendering}. Comparison of LiDAR point clouds rendered by different methods. Previous approaches exhibit distorted scan-line patterns and incomplete geometry, including vulnerable road users such as pedestrians. }
    \label{fig:lidar-pc-comparison}
\end{figure*}

\paragraph{Scene Reconstruction.}

Following common practice, we evaluate the upper bound of reconstruction and modeling capacity of frameworks by reconstructing scenes using all available RGB images and LiDAR sweeps. Quantitative results on three datasets are reported in \Cref{tab:all-datasets}. On the Waymo dataset, XSIM achieves substantial improvements over the previous state of the art, SplatAD, with gains of \textcolor{OliveGreen}{+3.01} PSNR and \textcolor{OliveGreen}{+3.8\%} SSIM, while reducing LPIPS by \textcolor{OliveGreen}{20.6\%}. Across all three datasets, XSIM consistently demonstrates superior RGB image reconstruction quality compared to prior baselines. On PandaSet, performance remains competitive across all metrics, with LPIPS showing a minor performance drop.

\paragraph{Novel-view synthesis.}

As our framework targets simulation of novel scenarios and trajectories, we evaluate rendering quality on views unseen during reconstruction (\Cref{tab:all-datasets}). We train on every second sensor frame and evaluate on the remaining frames. In this setting, XSIM achieves gains of \textcolor{OliveGreen}{+2.74} PSNR on Waymo and \textcolor{OliveGreen}{+1.04} PSNR on Argoverse over previous state-of-the-art method. We further assess extrapolation capability in \Cref{fig:nvs-lane-shift} by rendering ego-vehicle cameras under a lateral shift of 3 meters. Compared to prior methods, our framework produces more consistent renderings.

\subsection{LiDAR and depth rendering}

We evaluate LiDAR rendering quality on three datasets in both scene reconstruction and novel-view synthesis settings using Chamfer Distance (CD) (\Cref{tab:all-datasets}). XSIM achieves state-of-the-art results on all datasets, with an 
\textcolor{OliveGreen}{x8.8} CD error reduction on Waymo reconstruction and an \textcolor{OliveGreen}{x4.5} error reduction on Waymo NVS. As shown in \Cref{fig:lidar-pc-comparison}, XSIM better preserves characteristic LiDAR ring patterns and accurately renders pedestrians. We further assess geometric quality via RGB-camera depth rendering (\Cref{fig:depth}), where our method produces smooth, dense depth maps. 

\subsection{Ablation study}

We illustrate the effect of our phase modeling mechanism via synthetic example~(\cref{fig:projection-ablation}) with a single car mesh captured by rolling shutter LiDAR camera. As the LiDAR moves along the vehicle, it becomes visible twice near the azimuthal boundaries of the range image due to rolling shutter. While 3DGUT default projection produces artifacts at the boundaries, our framework with phase modeling precisely reconstructs and renders the scene.
We further ablate our framework components in a novel-view synthesis setting on six scenes from Waymo dataset by individually disabling features (Tab. \ref{tab:ablations}). Camera rolling-shutter modeling (\texttt{c} vs. \texttt{a}) leads to significant gains in PSNR. Removing the separate LiDAR opacity parameter (\texttt{c} vs. \texttt{a}) degrades both image and LiDAR metrics, with qualitative effects shown in \Cref{fig:lidar-density}. Modeling LiDAR rolling shutter (\texttt{d} vs. \texttt{c}) with our phase modeling mechanism (\texttt{e} vs. \texttt{c}) further improves image and LiDAR rendering demonstrating effectiveness of our contributions.

\begin{table}[!htbp]
\setlength{\tabcolsep}{3.5pt}
\begin{tabular}{lcccc} 
\toprule 
\renewcommand{\arraystretch}{1.1}
Component & \multicolumn{1}{l}{PSNR$\uparrow$} & \multicolumn{1}{l}{SSIM$\uparrow$} & \multicolumn{1}{l}{LPIPS$\downarrow$} & \multicolumn{1}{l}{CD$\downarrow$} \\ \midrule
\footnotesize{(a)} XSIM, full & \textbf{30,03}           & \textbf{0,8945}          & \textbf{0,2122}           & \textbf{0,21}          \\
\footnotesize{(b)} -- Camera rolling shutter                       & 28,95                    & 0,8761                   & 0,2407                    & 0,21                   \\
\footnotesize{(c)} -- Lidar opacity                                & 29,55                    & 0,8888                   & 0,2189                    & 0,25                   \\
\footnotesize{(d)} \hspace{0.2cm} -- Lidar rolling shutter                        & 29,05                    & 0,8816                   & 0,2296                    & 0,31                   \\
\footnotesize{(e)} \hspace{0.2cm} -- Phase modelling                              & 29,32                    & 0,8824                   & 0,2245                    & 0,28                   \\
\bottomrule
\end{tabular}
\caption{Ablation studies on novel-view synthesis on Waymo dataset (half of split). Each row corresponds to disabling of a single component relative to the parent configuration. }\label{tab:ablations}
\end{table}
\section{Conclusion}
\label{sec:conclusion}
In this paper, we presented XSIM, a unified sensor simulation framework for autonomous driving that extends 3DGUT splatting with generalized rolling-shutter modeling, enabling modeling of complex sensors in dynamic environments in a unified manner. 
Our phase modeling mechanism enables robust rendering for spherical rolling-shutter sensors by explicitly accounting for azimuthal discontinuities. 
Extensive experiments on multiple autonomous driving benchmarks demonstrate that our approach consistently improves geometric accuracy and photorealism, outperforming strong previous baselines.

\bibliographystyle{named}
\bibliography{ijcai26}

\appendix
\begin{appendices}

\section{Generic rolling-shutter camera projection}

We provide full listing of our generalized rolling shutter projection in \Cref{alg:projection}. Given the static camera projection function $\pi(x) \mapsto (u,v)$, rolling-shutter time function $\tau(u,v)$, sensor velocities $v_c, w_c$, torque-corrected actor point velocity $v_a'$, middle of exposure camera SE3 camera pose $q_0, t_0$, \Cref{alg:projection} allows to project world-space point $x_w$ directly into a rolling-shutter image. Due to unknown point observation time $\eta$, this algorithm iteratively finds solution for $\eta = \tau(u(\eta), v(\eta))$ equation. Here we use Newton-Raphson method for finding equation root, that requires computing Jacobian of discrepancy function $\Delta\eta = \eta - \tau(u(\eta),v(\eta))$. 

For linear rolling shutter time function, Jacobian is defined as:

\begin{equation}
    \frac{d(\Delta\eta)}{d\eta} = 1 - \tau_u \frac{du}{d\eta} - \tau_v \frac{dv}{d\eta}
\end{equation}

In its turn, screen-space coordinate derivatives with respect to time can be computed as:

\begin{equation}
    \frac{du}{d\eta} = \frac{d\pi_u}{dx_c} \frac{dx_c}{d\eta} \quad \quad  \frac{dv}{d\eta} = \frac{d\pi_v}{dx_c} \frac{dx_c}{d\eta}
\end{equation}
where $x_c$ -- point position in camera coordinate system. Here, $\frac{dx_c}{d\eta}$ does not depend on the camera model and is derived once and computed analytically in implementation. 
Jacobian of camera $\frac{d\pi}{dx_c}$ here is the same as Jacobian used in EWA splatting for approximating projection. 
For instance, given a perspective camera projection function $\pi_{\text{perspective}}(x) = (f_x \frac{x}{z} + c_x, f_y \frac{y}{z} + c_y)$ Jacobian is defined as:

\begin{equation}
    \frac{d\pi_{\text{perspective}}}{dx_c} = \begin{pmatrix}
        \frac{f_x}{z} & 0 & -\frac{f_x x}{z^2} \\
        0 & \frac{f_y}{z} & -\frac{f_y y}{z^2}
    \end{pmatrix}
\end{equation}
For LiDAR rendering, spherical projection is defined as:
\begin{equation}
    \pi_{\text{spherical}}(x,y,z) = (\atantwo(y,x), \arcsin\frac{y}{\sqrt{x^2+y^2+z^2}})
\end{equation}
and has Jacobian matrix (provided by \cite{hess2024splatad}):
\begin{equation}
    \frac{d\pi_{\text{spherical}}}{dx_c} = \begin{pmatrix}
        -\frac{y}{x^2+y^2} & \frac{x}{x^2+y^2} & 0 \\
        -\frac{xz}{r^2 \sqrt{x^2+y^2}} & -\frac{yz}{r^2\sqrt{x^2+y^2}} & \frac{\sqrt{x^2+y^2}}{r^2}
    \end{pmatrix}
\end{equation}

Note, that in general differentiability of projection function $\pi(x)$ is not required. In practice, Newton-Raphson method can be swapped with fixed-point iteration method by assuming $\frac{d\Delta\eta}{d\eta} = 1$. While both methods typically converge to precise solutions, we observe that in practice Newton-Raphson method converges 1-2 iterations faster than fixed-point iteration.

\begin{algorithm}[tb]
    \caption{$\pi_{\text{rolling}}$: Rolling-shutter point projection}
    \label{alg:projection}
    \textbf{Input}: $\pi(x), x_w, \eta_0$\\
    \textbf{Parameters}: $\tau(u,v), v_c, w_c, v_a', q_0, t_0, \Delta\eta_{\text{thr}},N$\\
    \textbf{Output}: $u,v,d,\eta,\text{isValid}$
    \begin{algorithmic}[1] 
        \STATE Let $\eta=\eta_0$.
        \STATE Let $i=0$
        \WHILE{($\Delta\eta > \Delta\eta_{\text{thr}}$) \OR ($i\leq N$)}
        \STATE \textcolor{MidnightBlue}{// According to Equation 7 of the main paper}
        \STATE $u,v,d$ = $\pi(q^{-1}(\eta)) \otimes (x_w(\eta) - t(\eta)) \otimes q(\eta))$
        \STATE \textcolor{MidnightBlue}{// $(u,v) \in [0;1]^2$ -- screen space point coordinates}
        \STATE \textcolor{MidnightBlue}{// $d$ -- projected point depth}
        \STATE $\Delta \eta$ = $\eta - \tau(u, v)$
        \STATE $\frac{d(\Delta\eta)}{d\eta} = 1 - \tau_u \frac{du}{d\eta} - \tau_v \frac{dv}{d\eta}$
        \STATE \textcolor{MidnightBlue}{// Newton-Raphson iteration:}
        \STATE $\eta = \eta - \Delta\eta / \frac{d(\Delta\eta)}{d\eta} $
        \STATE ++i
        \ENDWHILE
        \STATE isValid = ($d > 0$) \AND ($\Delta\eta < \Delta\eta_{\text{thr}}$)
        \RETURN{$u,v,d,\eta$, isValid} 
    \end{algorithmic}
\end{algorithm}

\section{Phase modeling}

Complete listing of Unscented Transform with phase modeling mechanism for spherical rolling-shutter camera is provided in \Cref{alg:phm}. For simplicity of description, here we assume that projection function return values in radians, rather than normalized coordinates. As $\text{UT}(\cdot)$ function we denote standard Unscented Transform, which constructs sigma points, projects them via given function $\pi(\cdot)$, and returns $\mu_{2D} = (\mu_\varphi, \mu_\theta)$, particle 2D extent $(\varphi_{ext}, \theta_{ext})$ based on estimated 2D covariance, depth $d$, and flag $I$ that specifies if projection is sucessful and projected particle extent intersects with visible image range. By considering additional projections $\pi_{\text{negative}}$ and $\pi_{\text{positive}}$, our mechanism handles cases where particle 2D projection becomes bimodal.

\begin{algorithm}[!t]
    \caption{Unscented transform with phase modeling}
    \label{alg:phm}
    \textbf{Input}: $\mu_{3D}, \Sigma_{3D}, \pi_{\text{rolling}}(x)$ \\
    \textbf{Parameters}: $ \tau(u,v), v_c, w_c, v_a', q_0, t_0$ \\
    \textbf{Output}: $\mu_{2D} \in \sR^{K\times 3}$, $\varphi_{\text{ext}} \in \sR^{K}$, $\theta_{\text{ext}} \in \sR^{K}$, $d \in \sR^{K}$\\
    \begin{algorithmic}[1]
        \STATE {\textcolor{MidnightBlue}{// Define visible range projection function}}
        \STATE {\textcolor{MidnightBlue}{// Initialize solver with $\tau_{mid}$ time.}}
        \STATE {$\pi_{\text{central}} \coloneq \pi_{\text{rolling}} (x, \tau_{\text{mid}})$ }
        \STATE {\textcolor{MidnightBlue}{// Perform standard UT}}
        \STATE{$\mu_{\varphi}^C, \mu_{\theta}^C, \varphi_{ext}^C, \theta_{ext}^C, d^C, I^C$ = UT($\mu_{3D}, \Sigma_{3D}, \pi_{\text{central}})$}
        \STATE{\textcolor{MidnightBlue}{// If projection does not intersect with visible range}}
        \IF {\textbf{not} $I^C$}
            \RETURN {$\emptyset$}
        \ENDIF
        \STATE{\textcolor{MidnightBlue}{// Even if central projection is valid, particle still can be seen twice}}
        \STATE{\textcolor{MidnightBlue}{// Define negative shift projection shifted by $-\pi$}}
        \STATE{$\pi_{\text{negative}}(x) \coloneq \pi_{\text{rolling}}(x, \tau_{\text{start}}) - 2\pi [ \pi_{\text{rolling}}(x, \tau_{\text{start}}) \geq 0]$}
        \STATE{$\mu_{\varphi}^L, \mu_{\theta}^L, \varphi_{ext}^L, \theta_{ext}^L, d^L, I^L$ = UT($\mu_{3D}, \Sigma_{3D}, \pi_{\text{negative}})$}
        \STATE{\textcolor{MidnightBlue}{// Define positive shift projection shifted by $+\pi$}}
        \STATE{$\pi_{\text{positive}}(x) \coloneq \pi_{\text{rolling}}(x, \tau_{\text{end}}) + 2\pi [ \pi_{\text{rolling}}(x, \tau_{\text{start}}) < 0]$}
        \STATE{$\mu_{\varphi}^R, \mu_{\theta}^R, \varphi_{ext}^R, \theta_{ext}^R, d^R, I^R$ = UT($\mu_{3D}, \Sigma_{3D}, \pi_{\text{positive}})$}

        \STATE{\textcolor{MidnightBlue}{// If both projections valid, return both}}
        \IF {$I^L$ \AND $I_R$ \AND $\varphi_{ext}^L < \pi$ \AND $\varphi_{ext}^R < \pi$}
            \RETURN {$(\mu_{2D}^L, \mu_{2D}^R), (\varphi_{ext}^L, \varphi_{ext}^R), (\theta_{ext}^L, \theta_{ext}^R), (d^L, d^R)$}
        \ENDIF
        \STATE{\textcolor{MidnightBlue}{// Only one projection is valid, return it}} \\
        \IF {$I_L$ \AND $\varphi_{ext}^L < \pi$} 
            \RETURN {$\mu_{\varphi}^L, \mu_{\theta}^L, \varphi_{ext}^L, \theta_{ext}^L, d^L$}
        \ENDIF
        \IF {$I_R$ \AND $\varphi_{ext}^R < \pi$} 
            \RETURN {$\mu_{\varphi}^R, \mu_{\theta}^R, \varphi_{ext}^R, \theta_{ext}^R, d^R$}
        \ENDIF
        \RETURN {$\mu_{\varphi}^C, \mu_{\theta}^C, \varphi_{ext}^C, \theta_{ext}^C, d^C$}
    \end{algorithmic}
\end{algorithm}

\section{Framework details}

\paragraph{Initialization} We initialize the Gaussian scene representation using LiDAR sweeps, object bounding box annotations, and camera data available in the driving logs. For each LiDAR sweep, we use the corresponding bounding box annotations to separate points into static background and individual dynamic actors. Points are colored by projecting them onto the corresponding camera images. Points that are not visible in any camera and therefore lack color are colored using the three nearest neighbors with known color. For dynamic actors known to be symmetric (e.g., vehicles), we additionally symmetrize points along the longitudinal axis. Background and actor points are then randomly downsampled to meet node-specific thresholds. Additional points are allocated using inverse-distance sphere sampling to cover regions of the scene outside LiDAR coverage.

\paragraph{Scene nodes} Our framework is fully modular, consisting of multiple nodes that form the scene representation, render it, and compute loss functions given a camera input. We begin by estimating dynamic object SE(3) poses and velocities at the sensor’s mid-exposure time. Object poses are initialized from dataset annotations and refined using learnable additive corrections. Poses and velocities at arbitrary times are obtained by interpolating the trajectory. Following \cite{chen2025omnire}, the scene Gaussian representation comprises static, rigid, deformable, and SMPL scene nodes. The SMPL node represents humans detected by a pretrained pose and shape estimation network, while the deformable node is used for undetected far-range pedestrians and cyclists. The deformable node employs a learnable MLP with instance embeddings to deform actor-frame Gaussians according to sensor time. The concatenated world-space Gaussians are then rendered using our 3DGUT-based procedure. The resulting color feature maps are transformed into RGB renderings by a small CNN post-processor~\cite{hess2024splatad}. Finally, the RGB rendering is alpha-composited with a learnable environment cube map to represent the sky.

\paragraph{Loss functions}

Our scene representation consisting of multiple object nodes is optimized simultaneously from driving logs by randomly sampling images and closest by time LiDAR sweeps at each iteration. We supervise it using combination of losses:
\begin{equation}
\mathcal{L} = \underbrace{\lambda \mathcal{L}_1 + (1- \lambda)L_{\text{SSIM}}}_{\text{camera guidance}} + \underbrace{\mathcal{L}_{\text{depth}}}_{\text{LiDAR}} + \mathcal{L}_{\text{opacity}} + \mathcal{L}_{\text{reg}}
\end{equation}

Specifically, $L_{\text{reg}}$ loss function is defined as $0.01L_{\text{mask}} + 0.01L_{\text{pose}} + L_{\text{SMPL}}$. Following OmniRE~\cite{chen2025omnire} we use sky segmentation masks and use $L_{mask}$ to penalize rendered image alpha for pixels assigned by segmentation mask to sky. $L_{\text{pose}}$ loss function is a $L_2$ penalty on actor trajectory adjustments that prevents drift of actors inside their coordinate system. $L_{\text{SMPL}}$ is also derived from OmniRE, consisting of multiple regularizations that constrain Gaussians onto SMPL body shape. 

\paragraph{Gaussian strategy}
We use densification and splitting strategy proposed by 3DGUT~\cite{wu20253dgut} with minimal modifications. While vanilla 3DGS~\cite{kerbl20233dgs} uses 2D position gradients as a criteria for particle splitting and cloning, 3DGUT framework uses particle positions in 3D directly and does not produce 2D positional gradients directly. Instead, 3D position gradients norm multiplied by distance to camera is used as a direct criteria replacement. When supervised by RGB camera rendering, high 3D positional gradients are typically correspond to underrepresented scene regions, causing densification by design. This intuition falls apart in case of LiDAR supervision, which produces strong 3D positional gradients along the rays, causing excessive particle densification. We solve this issue by accumulating gradients for criteria only based on RGB cameras supervision. To unify densification strategy with original 3DGS~\cite{kerbl20233dgs}, we also add the same densification criteria based on 2D scales. To enforce proper scene decomposition, we prune dynamic nodes particles which fall outside of their bounding boxes. We also modify opacity-based pruning criteria to be based on maximum value out of lidar and camera particle opacities. 

\paragraph{Optimization} We optimize our scene representation simultaneously with an Adam optimizer for 40000 iterations. Gaussian particles learning rates match SplatAD~\cite{hess2024splatad}, while most other scene node-specific learning rates are derived from~\cite{chen2025omnire}. We detail specific learning rate values in \Cref{tab:lr}. All learning rates are scheduled with 500 iterations warm-up and exponential decay if multiple LR-s are specified. 

\begin{table}[!h]
    \centering
    \begin{tabular}{l|c|c}
    \toprule
         Parameter & Initial LR & Final LR \\ \midrule
         Positions & 1.6e-4 & 1.6e-6 \\
         Scale & \multicolumn{2}{c}{5e-3} \\
         Rotation & \multicolumn{2}{c}{1e-3} \\
         Camera opacity & \multicolumn{2}{c}{0.05}\\
         LiDAR opacity & \multicolumn{2}{c}{0.05}\\
         Diffuse color & \multicolumn{2}{c}{2.5e-3}\\
         Specular features & \multicolumn{2}{c}{2.5e-3}\\
         \midrule
         Actors quaternion corr-s & 1e-5 & 5e-6 \\
         Actors translation corr-s & 5e-4 & 1e-4\\
         Deformable actor embeds & 1e-3 & 1e-4 \\
         Deformable MLP & 8e-3 & 8e-4 \\
         Post-process CNN & \multicolumn{2}{c}{1e-3}\\
         Env light texture & \multicolumn{2}{c}{0.01}\\
    \bottomrule
    \end{tabular}
    \caption{Optimization learning rates}
    \label{tab:lr}
\end{table}

\section{Datasets}

This section lists details related to datasets we use in our experimental evaluation. Overall, we use three datasets -- Waymo Open Dataset~\cite{sun2020waymoopen}, Argoverse 2~\cite{Argoverse2}, PandaSet~\cite{xiao2021pandaset}.

\paragraph{Waymo Open Dataset.} Dataset features $\approx 19$ seconds long sequences with camera rig consisting of five RGB cameras (front, front left, front right, left and right). We use full resolution (1920x1080 for front and 1920x886 for side cameras) for training and evaluation. We choose 12 scenes from the split used by OmniRE~\cite{chen2025omnire} that feature ego-vehicle movement, diverse range of dynamic objects (vehicles, buses, heavy trucks, construction vehicles) and vulnerable road users (pedestrians, cyclicts). Specifically, we use the following scenes: 
\texttt{10231929575853664160\_1160\_000\_1180\_000 (16), 10391312872392849784\_4099\_400\_4119\_400 (21), 12027892938363296829\_4086\_280\_4106\_280 (94), 12251442326766052580\_1840\_000\_1860\_000 (102), 13254498462985394788\_980\_000\_1000\_000 (149),
    1382515516588059826\_780\_000\_800\_000 (172),
    16801666784196221098\_2480\_000\_2500\_000 (323),
17388121177218499911\_2520\_000\_2540\_000 (344),
1918764220984209654\_5680\_000\_5700\_000 (402),
 4487677815262010875\_4940\_000\_4960\_000 (552),
454855130179746819\_4580\_000\_4600\_000 (555),
9653249092275997647\_980\_000\_1000\_000 (788)}
    

For ablations we used half of this scenes list: 21, 94, 344, 552, 555, 788. For modeling humans as SMPL~\cite{loper2015smpl} bodies, we use poses and shape parameters provided by ~\cite{chen2025omnire}.

\paragraph{Argoverse 2.} Dataset consists of $\approx 15.5$ seconds long sequences with seven RGB cameras (2048x1550 resolution for six cameras and frontal camera with 1550x2048 resolution). Following ~\cite{hess2024splatad} we crop bottom 250 pixels of three cameras that contain ego-vehicle. We reuse the same 10 sequences split used by previous works ~\cite{tonderski2024neurad,hess2024splatad}:
\texttt{05fa5048-f355-3274-b565-c0ddc547b315,
0b86f508-5df9-4a46-bc59-5b9536dbde9f,
185d3943-dd15-397a-8b2e-69cd86628fb7,
25e5c600-36fe-3245-9cc0-40ef91620c22,
27be7d34-ecb4-377b-8477-ccfd7cf4d0bc,
280269f9-6111-311d-b351-ce9f63f88c81,
2f2321d2-7912-3567-a789-25e46a145bda,
3bffdcff-c3a7-38b6-a0f2-64196d130958,
44adf4c4-6064-362f-94d3-323ed42cfda9,
5589de60-1727-3e3f-9423-33437fc5da4b}

\paragraph{PandaSet. } Provides sequences of $\approx 8$ seconds each with an RGB camera rig consisting of six cameras.  All cameras have 1920x1080 resolution. We also crop bottom 260 pixels from the back camera to remove ego-vehicle. As in previous works~\cite{tonderski2024neurad,hess2024splatad}, we use 10 sequences: \texttt{1, 11, 16, 53, 63,84,106, 123,158}.

\paragraph{Synthetic example for projection comparison (main paper).} To illustrate the effect of our phase modeling mechanism, we constructed a synthetic scene with a single vehicle mesh (VW Golf MK4 by Jay-Artist, https://www.blendswap.com/blend/3976). We simulate rolling-shutter LiDAR by rendering ground-truth range images using a custom ray tracer. We sample ground-truth LiDAR images with resolution 1024x256 (uniform elevation beams from $-20^{\circ}$ to $20^{\circ}$) emulating moving along static vehicle at 30-60km/h speeds with vehicle seen near azimuth discontinuity boundaries. We optimize 3DGS using our rendering method with phase modelling mechanism and demonstrate the same representation rendered with standard unscented transform projection. Optimization with standard UT from scratch produces poor result due to excessive particles splitting caused by high positional gradients of boundary particles.

\section{Qualitative comparisons}

\begin{figure*}
    \centering
    \setlength{\tabcolsep}{0pt}
    \renewcommand{\arraystretch}{0.0}
    \begin{tabular}{ccccc}
        
         & \Large Ground-truth & \Large\textbf{XSIM, Ours} & \Large SplatAD & \Large OmniRE \\
        \normalsize
        
        \raisebox{2.7cm}{\rotatebox[origin=r]{90}{Argoverse}} & 
            \includegraphics[width=0.245\linewidth]{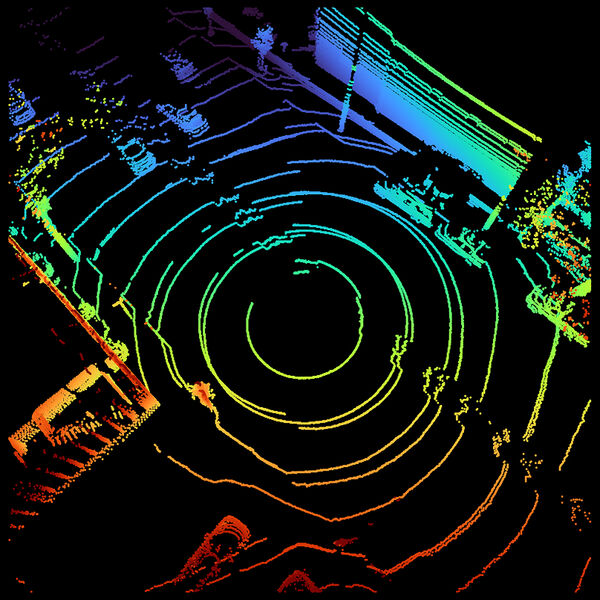} & 
            \includegraphics[width=0.245\linewidth]{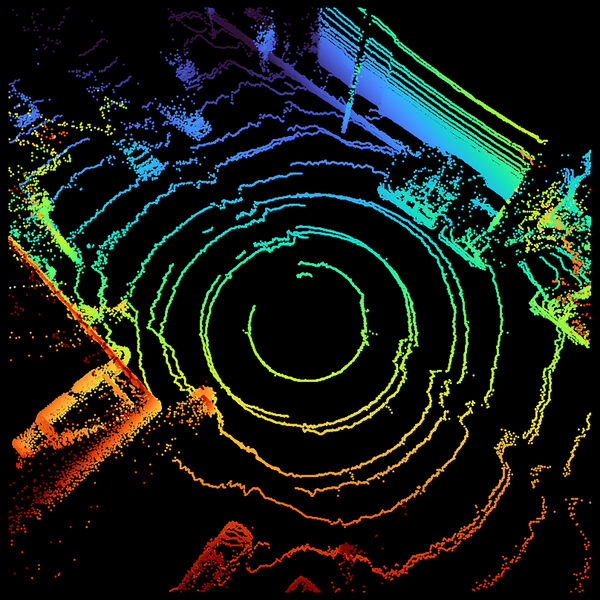} & 
            \includegraphics[width=0.245\linewidth]{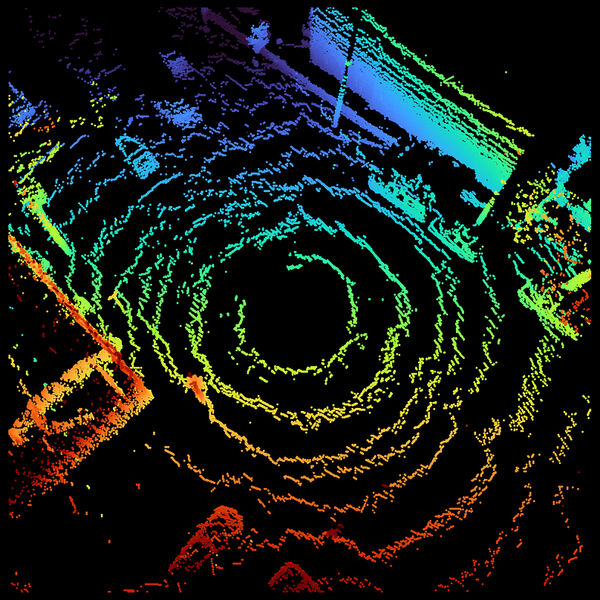} & 
            \includegraphics[width=0.245\linewidth]{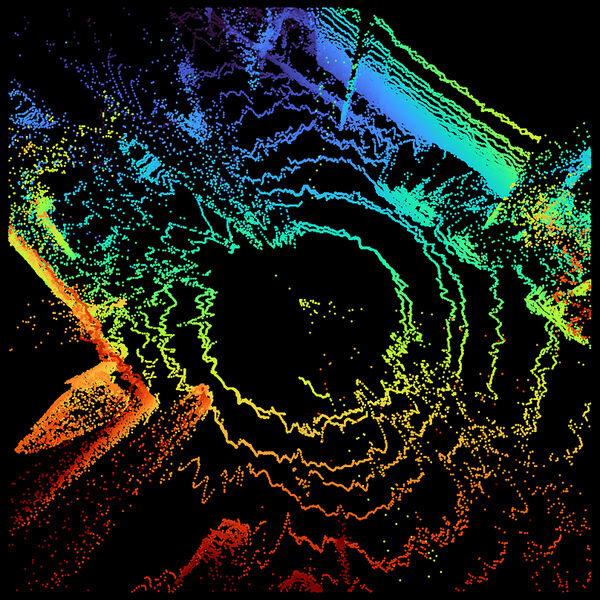} \\
        \raisebox{2.7cm}{\rotatebox[origin=r]{90}{Argoverse}} &
            \includegraphics[width=0.245\linewidth]{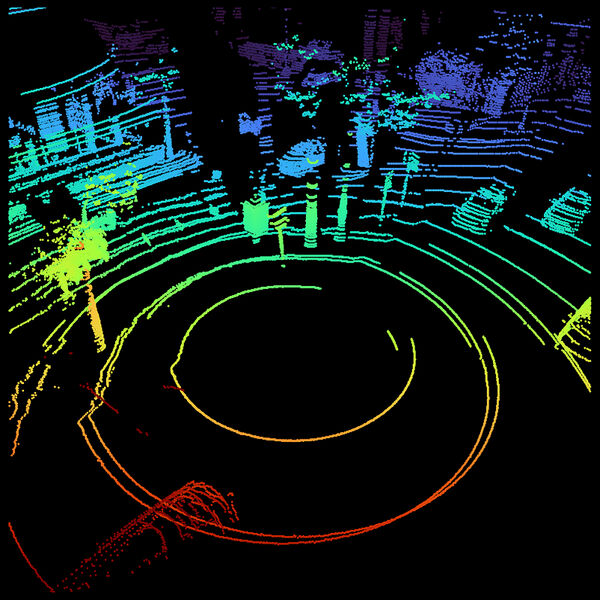} & 
            \includegraphics[width=0.245\linewidth]{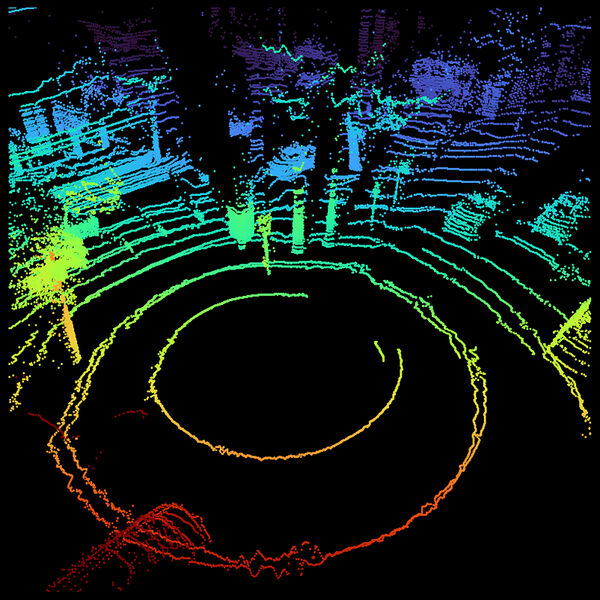} & 
            \includegraphics[width=0.245\linewidth]{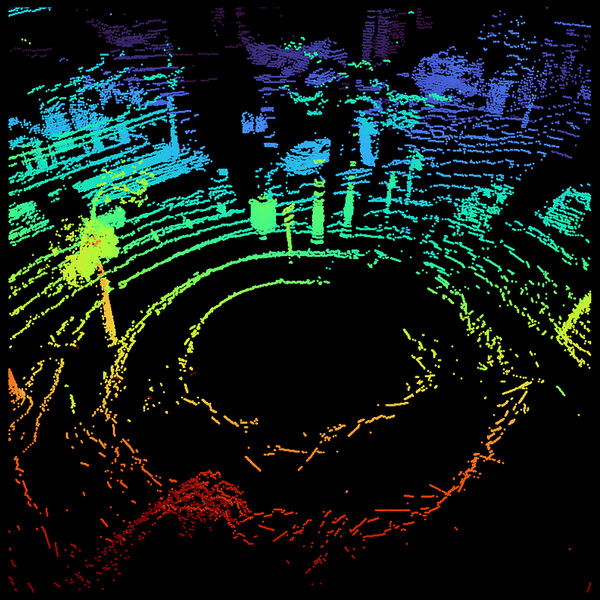} & 
            \includegraphics[width=0.245\linewidth]{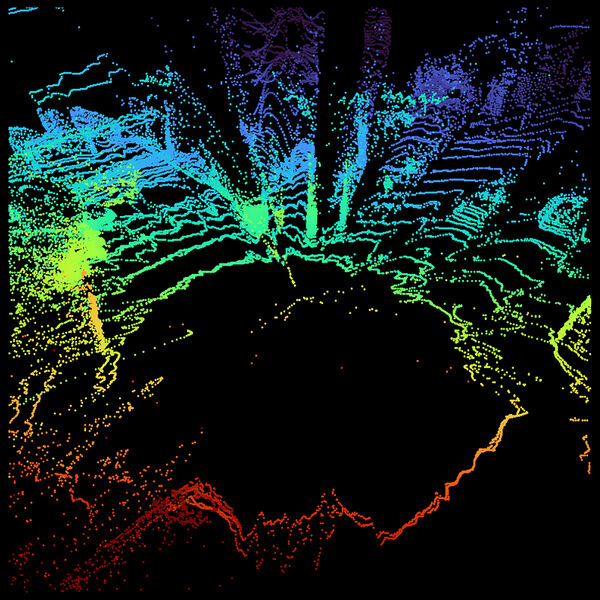} \\
        \raisebox{2.7cm}{\rotatebox[origin=r]{90}{Argoverse}} &
            \includegraphics[width=0.245\linewidth]{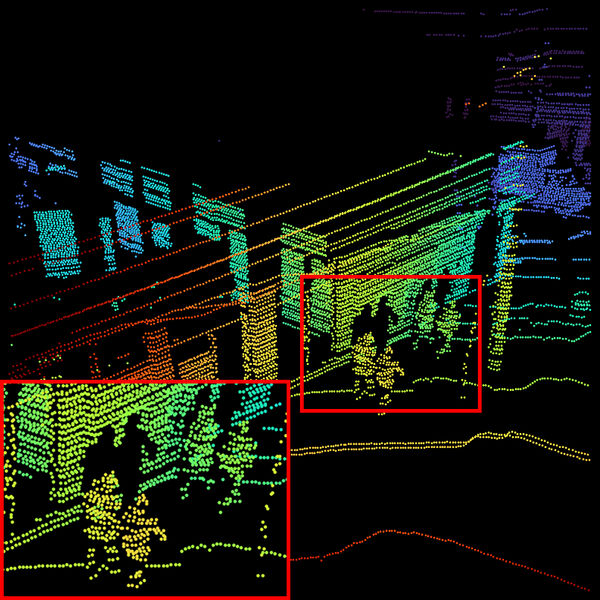} & 
            \includegraphics[width=0.245\linewidth]{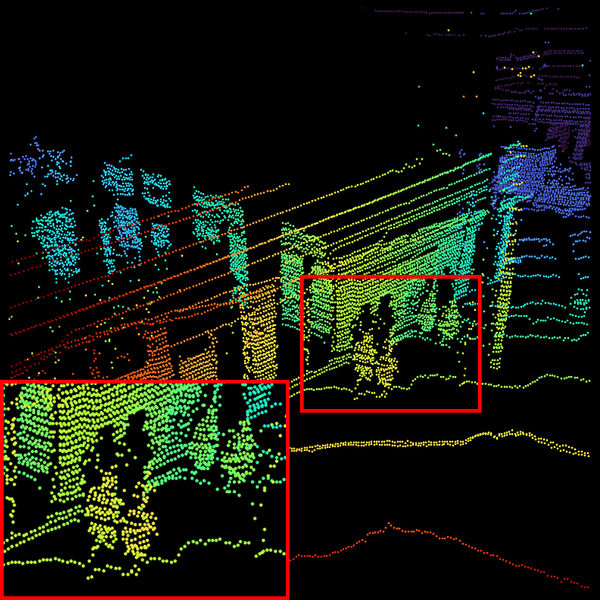} & 
            \includegraphics[width=0.245\linewidth]{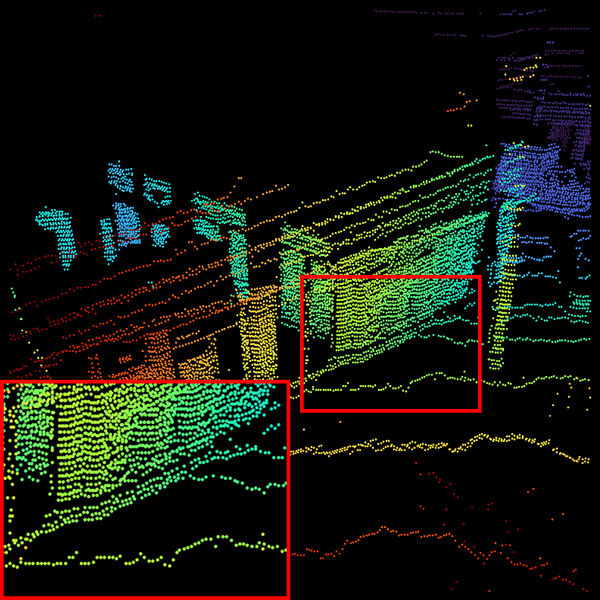} & 
            \includegraphics[width=0.245\linewidth]{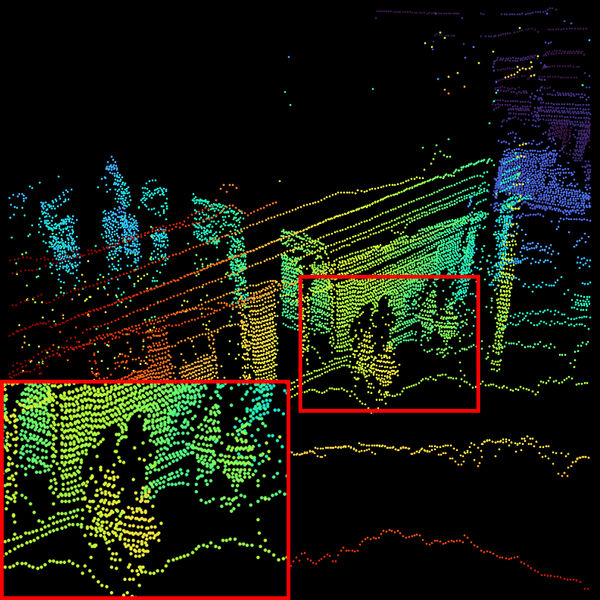} \\
        \raisebox{2.7cm}{\rotatebox[origin=r]{90}{Pandaset}} &
            \includegraphics[width=0.245\linewidth]{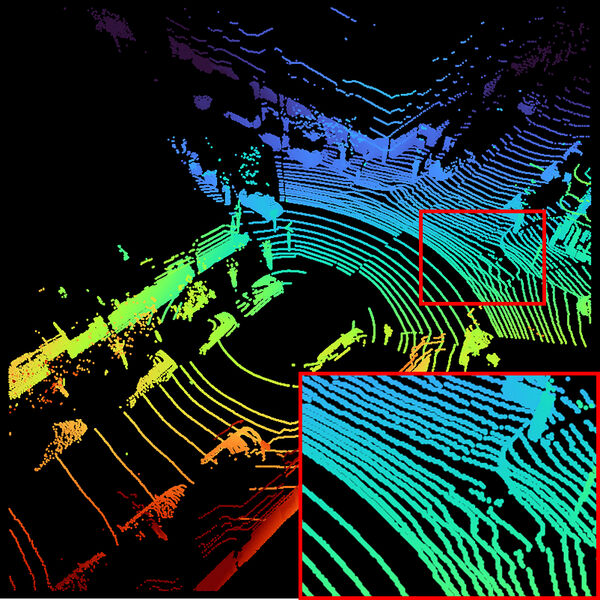} & 
            \includegraphics[width=0.245\linewidth]{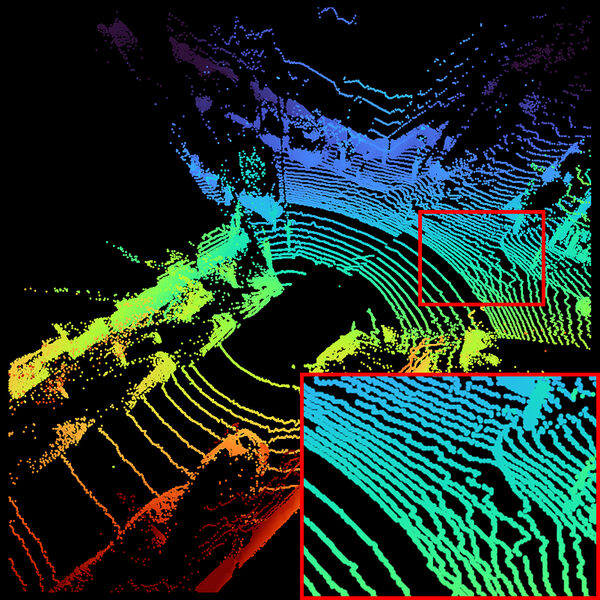} & 
            \includegraphics[width=0.245\linewidth]{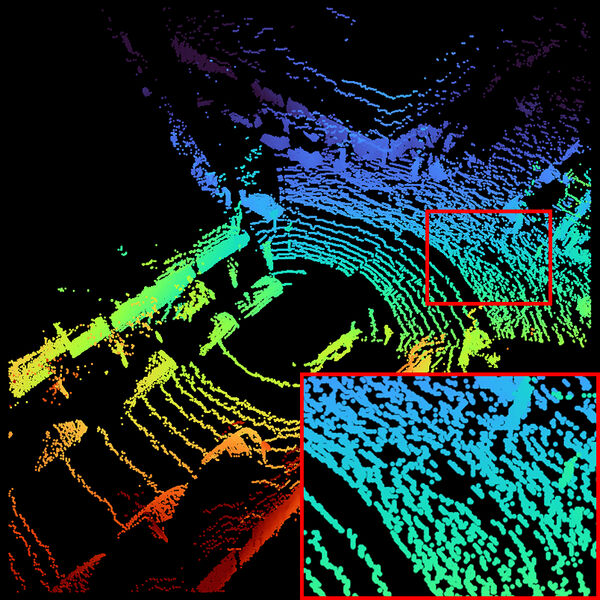} & 
            \includegraphics[width=0.245\linewidth]{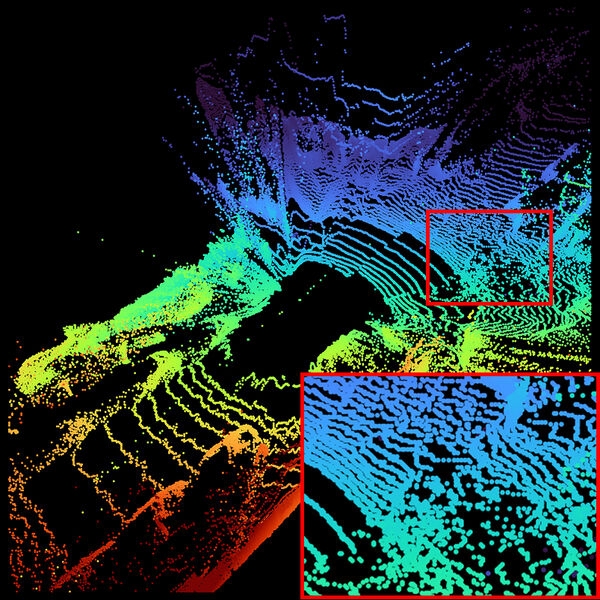} \\
        \raisebox{2.7cm}{\rotatebox[origin=r]{90}{Pandaset}} &
            \includegraphics[width=0.245\linewidth]{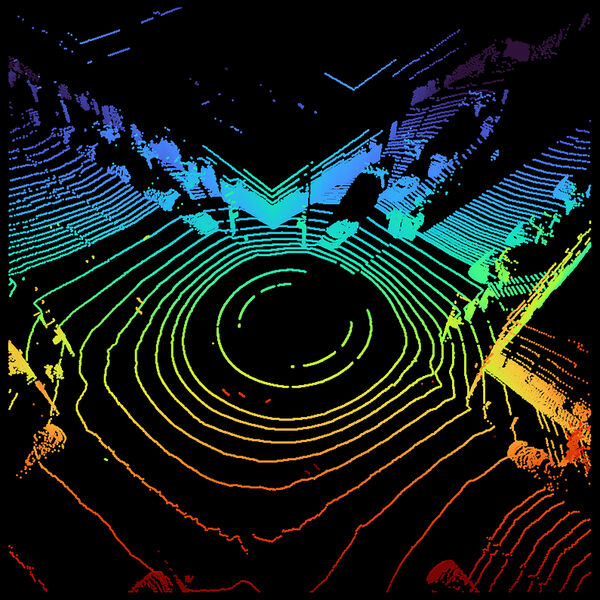} & 
            \includegraphics[width=0.245\linewidth]{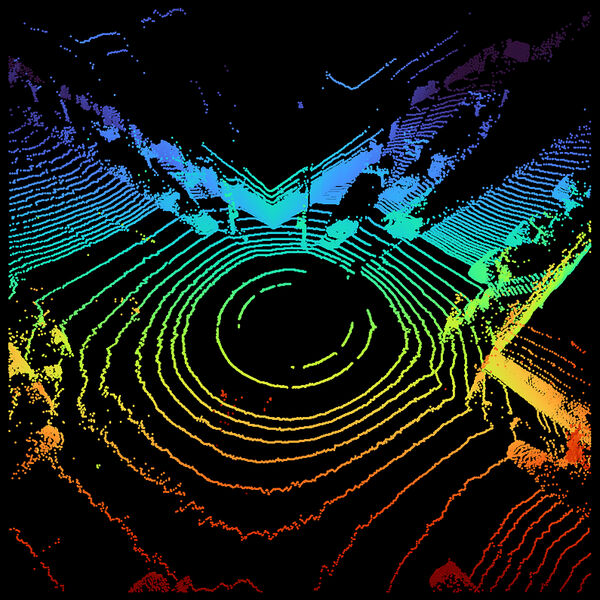} & 
            \includegraphics[width=0.245\linewidth]{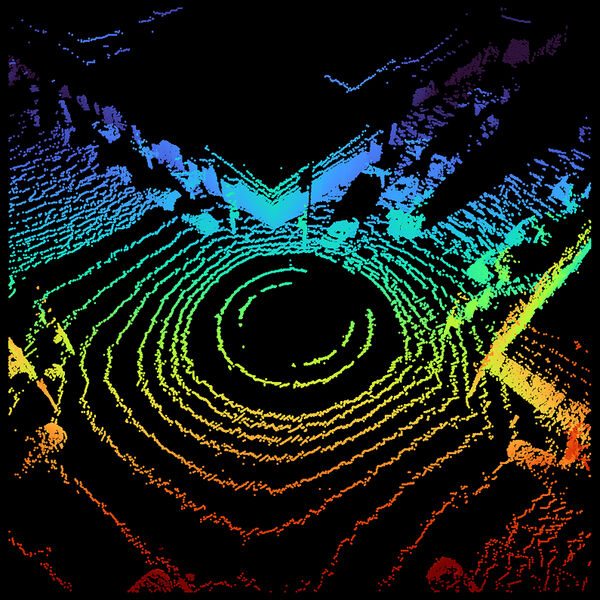} & 
            \includegraphics[width=0.245\linewidth]{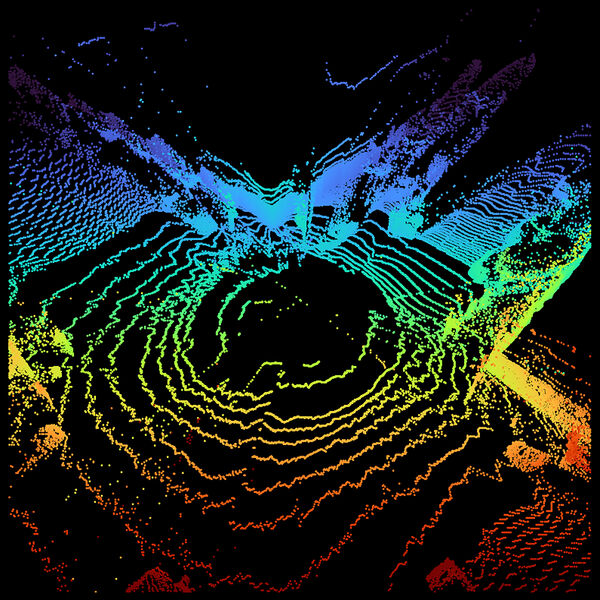} \\
        
    \end{tabular}
    \caption{Qualitative comparison of LiDAR point cloud rendering with previous methods on Argoverse and Pandaset datasets.}\label{fig:lidars}
\end{figure*}

\begin{figure*}
    \centering
    \setlength{\tabcolsep}{1pt}
    \renewcommand{\arraystretch}{1.0}
    \begin{tabular}{cccc}
        
        \Large Ground-truth & \Large\textbf{XSIM, Ours} & \Large SplatAD & \Large OmniRE \\
        \normalsize
        
        \includegraphics[width=0.245\linewidth]{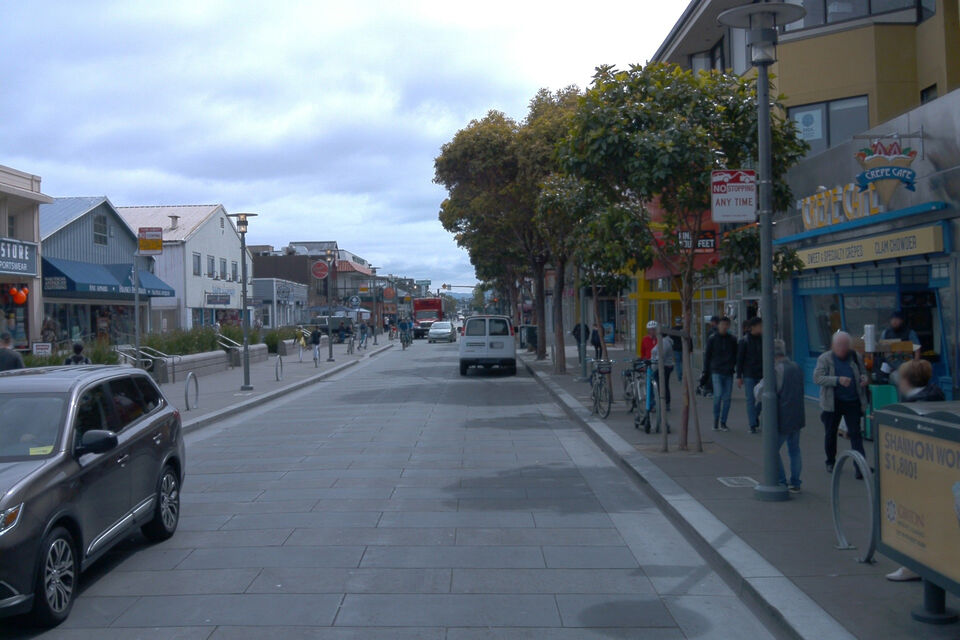} & 
            \includegraphics[width=0.245\linewidth]{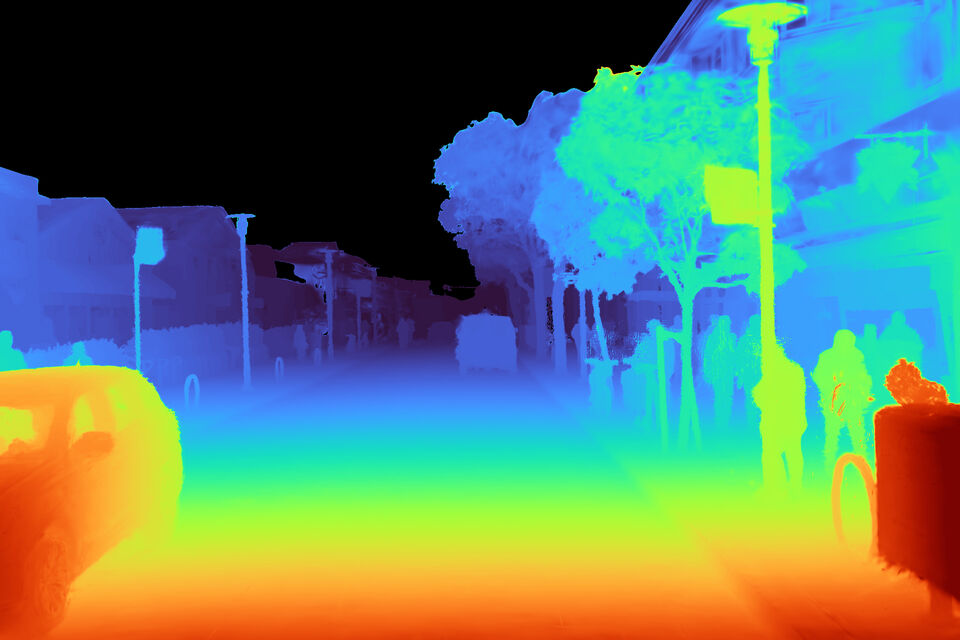} & 
            \includegraphics[width=0.245\linewidth]{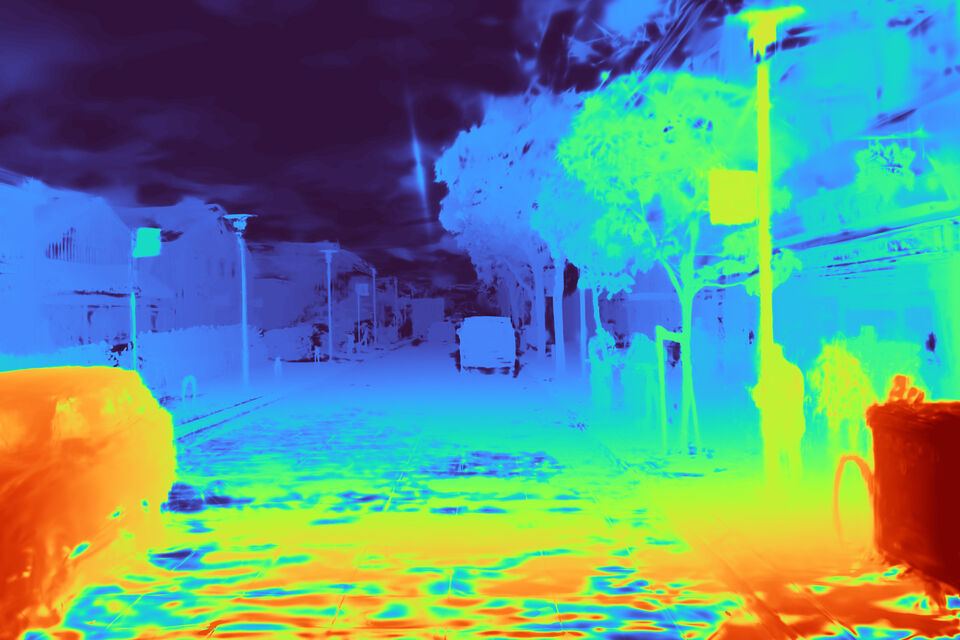} & 
            \includegraphics[width=0.245\linewidth]{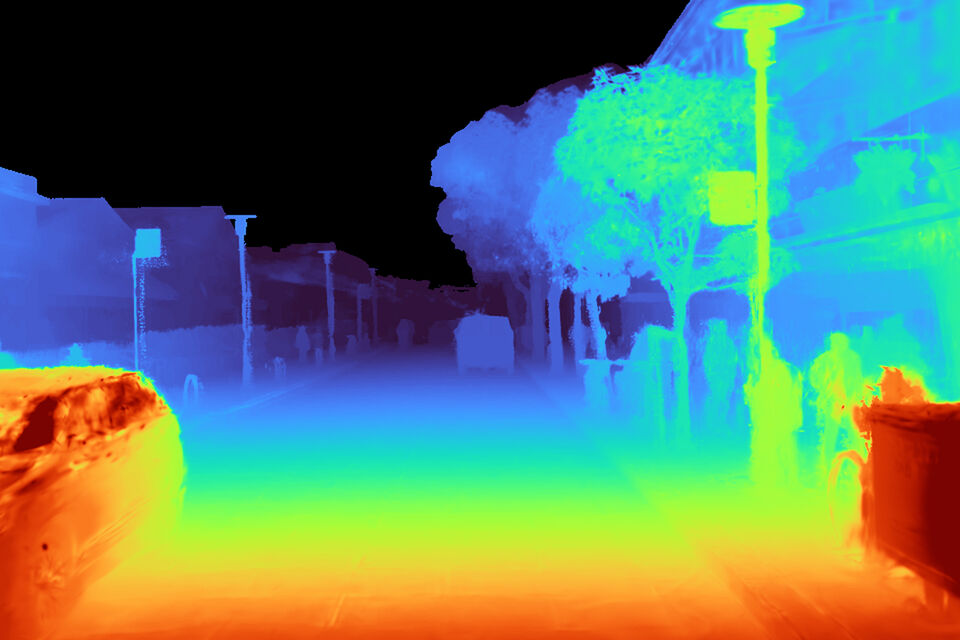} \\

        \includegraphics[width=0.245\linewidth]{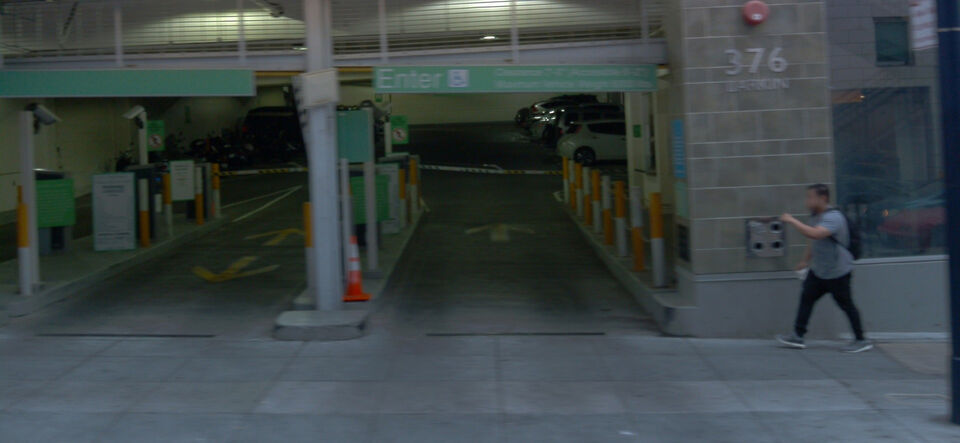} & 
            \includegraphics[width=0.245\linewidth]{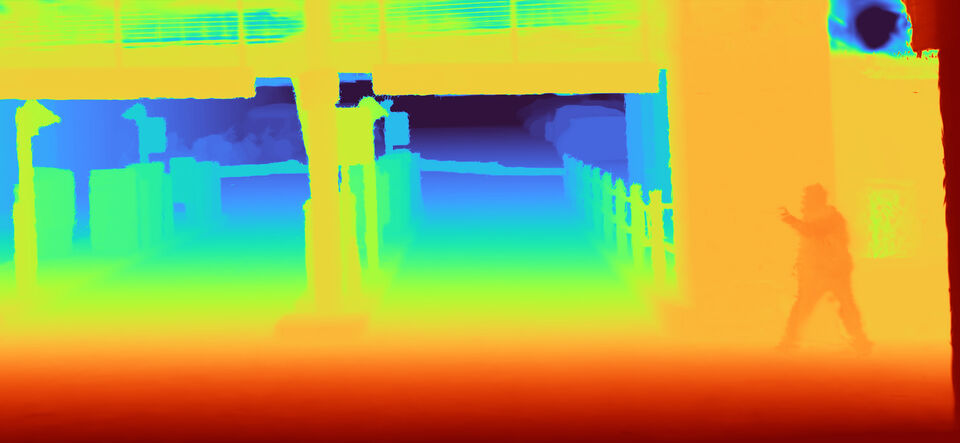} & 
            \includegraphics[width=0.245\linewidth]{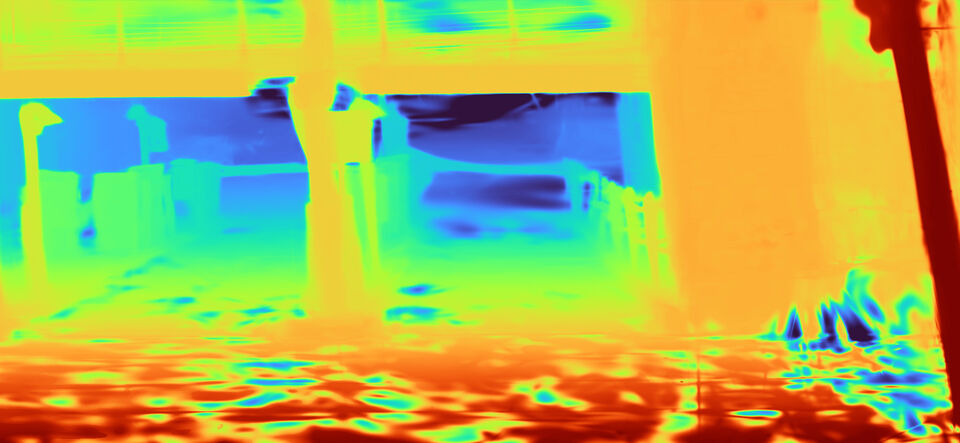} & 
            \includegraphics[width=0.245\linewidth]{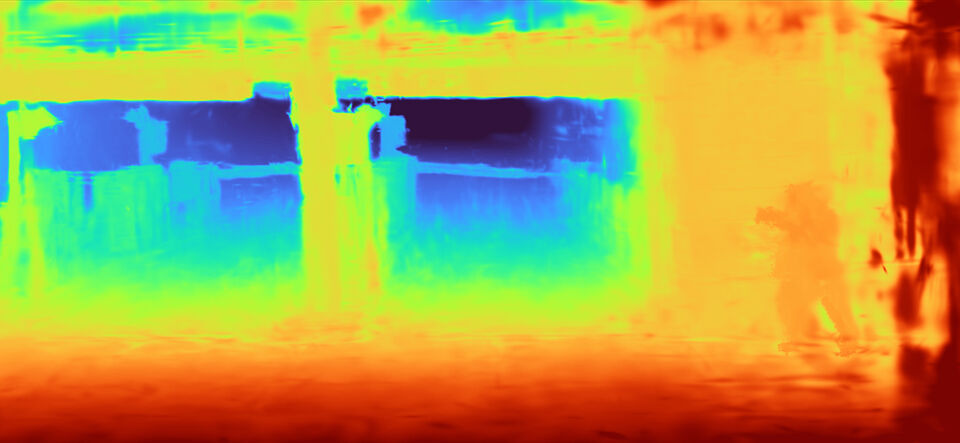} \\

        \includegraphics[width=0.245\linewidth]{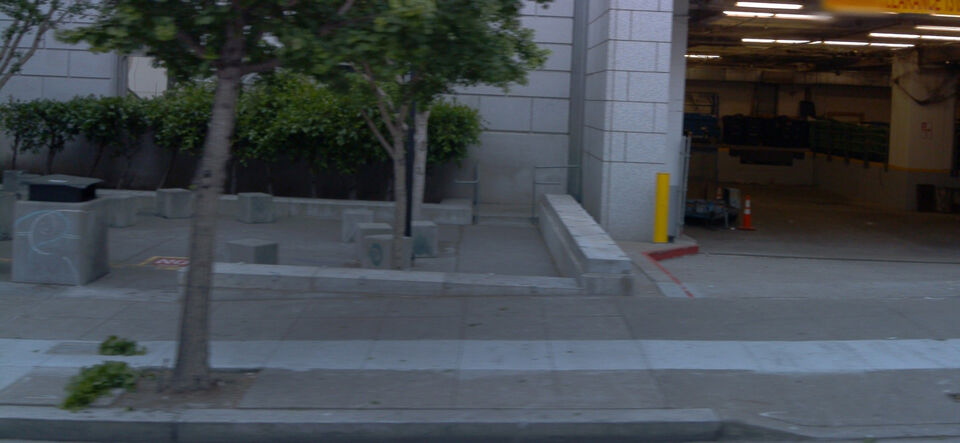} & 
            \includegraphics[width=0.245\linewidth]{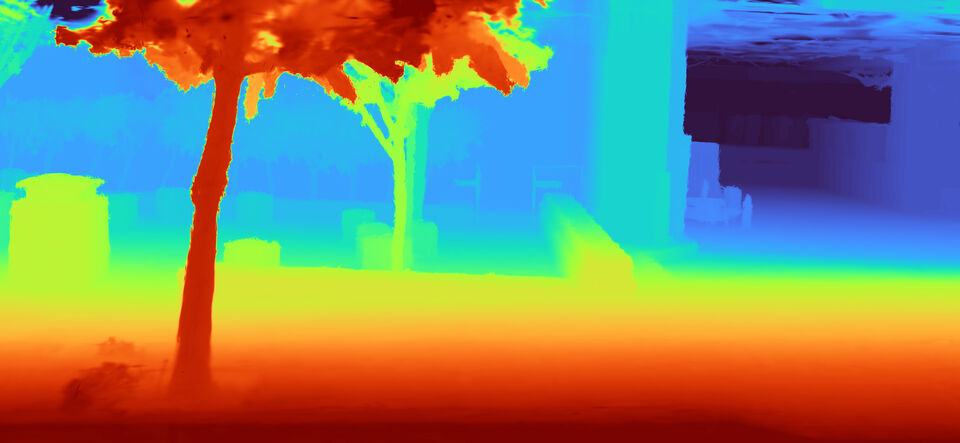} & 
            \includegraphics[width=0.245\linewidth]{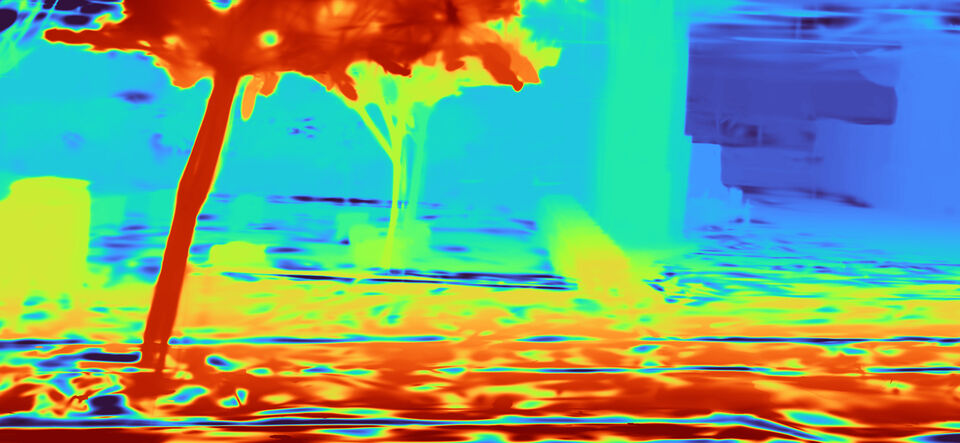} & 
            \includegraphics[width=0.245\linewidth]{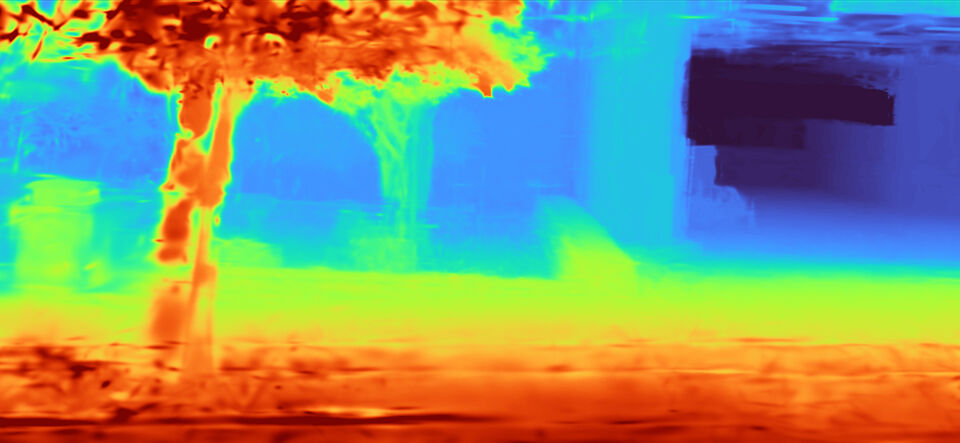} \\

        \includegraphics[width=0.245\linewidth]{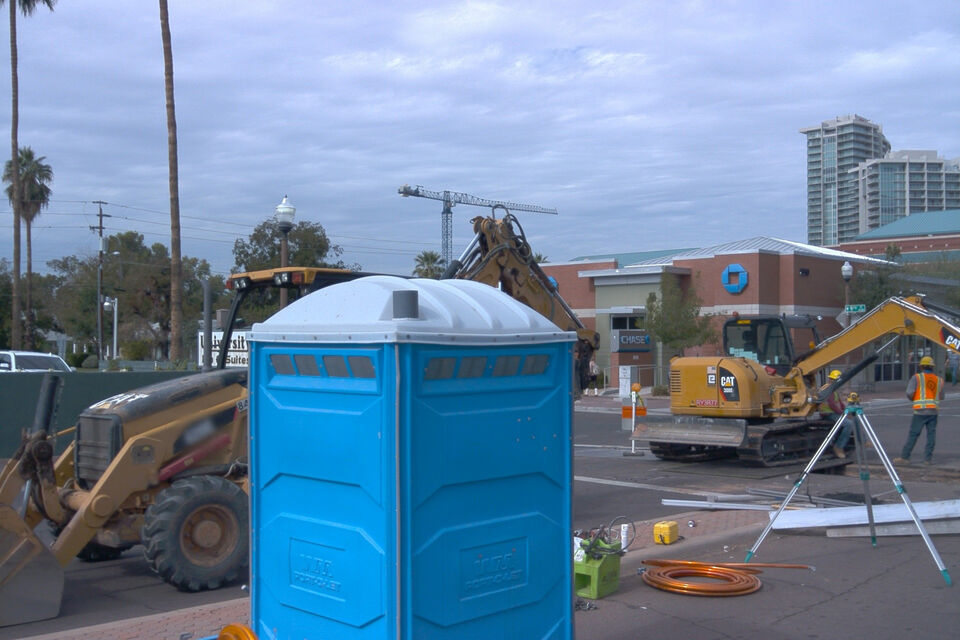} & 
            \includegraphics[width=0.245\linewidth]{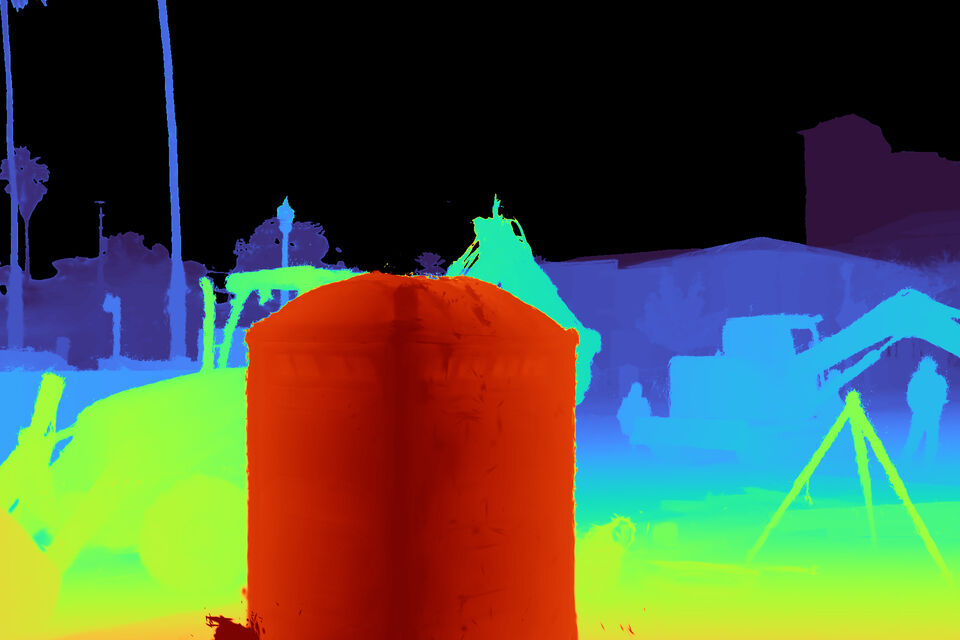} & 
            \includegraphics[width=0.245\linewidth]{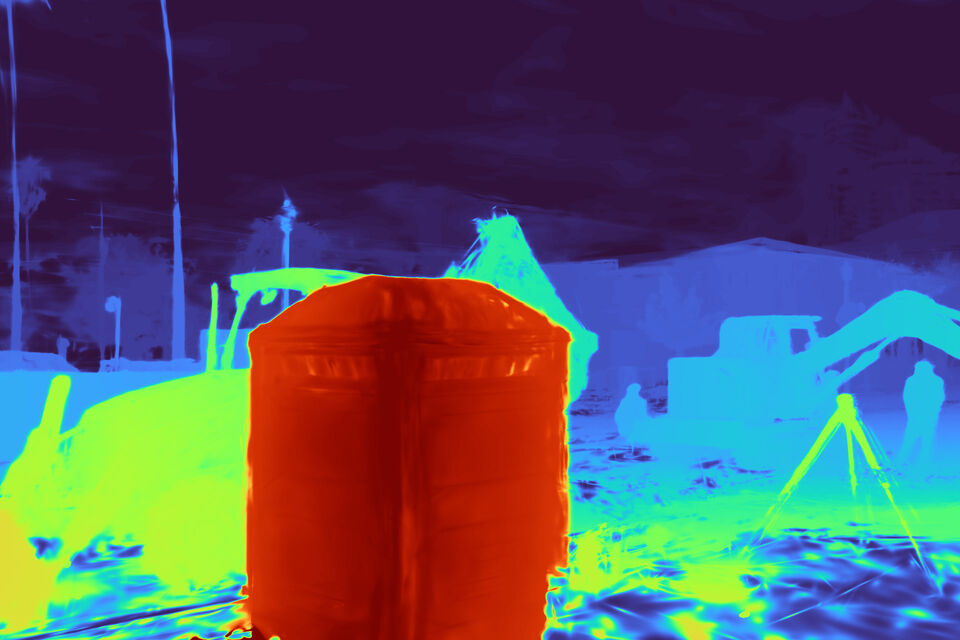} & 
            \includegraphics[width=0.245\linewidth]{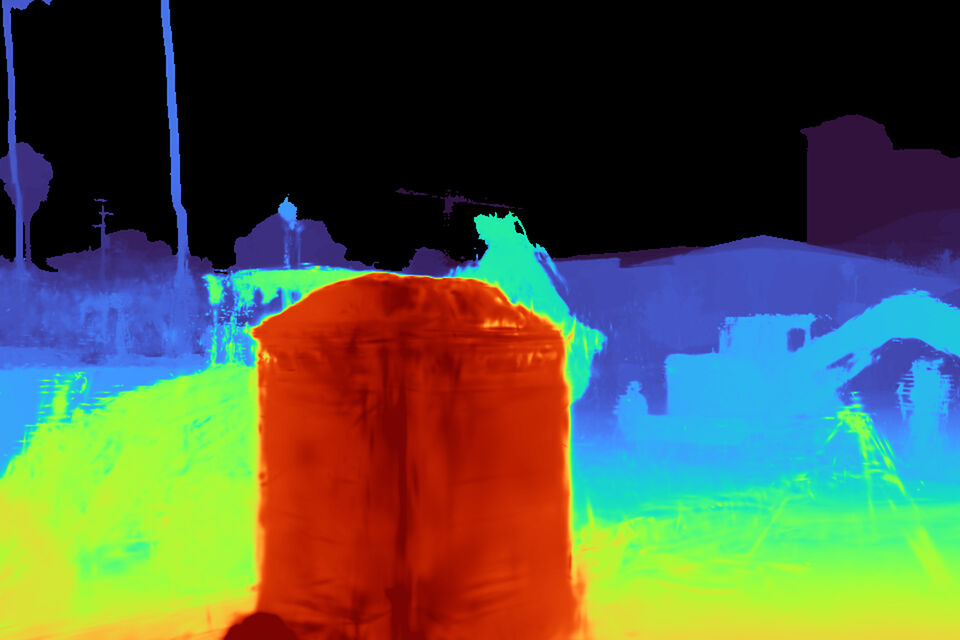} \\

        \includegraphics[width=0.245\linewidth]{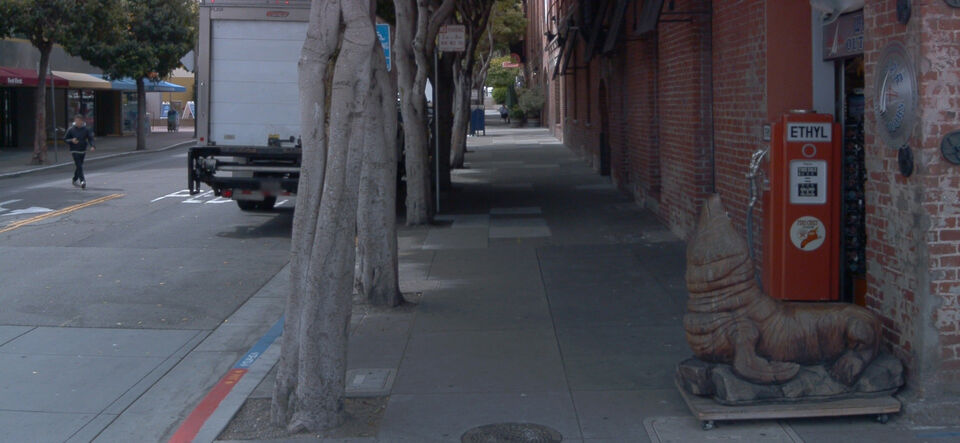} & 
            \includegraphics[width=0.245\linewidth]{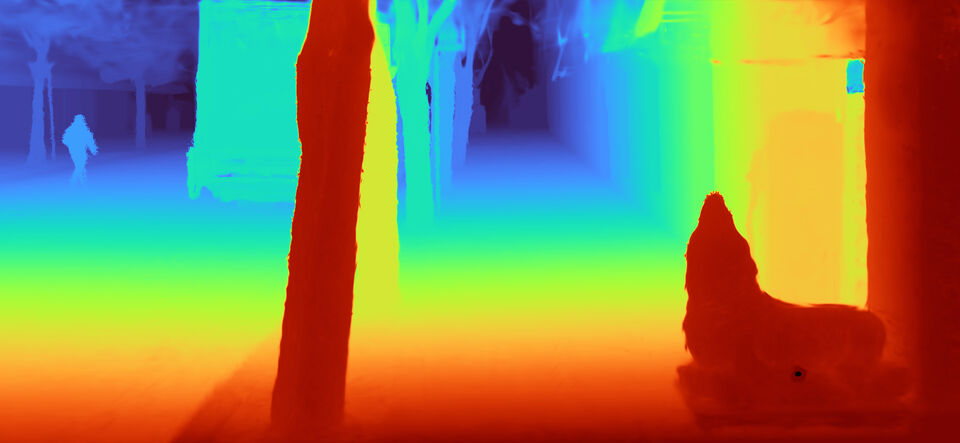} & 
            \includegraphics[width=0.245\linewidth]{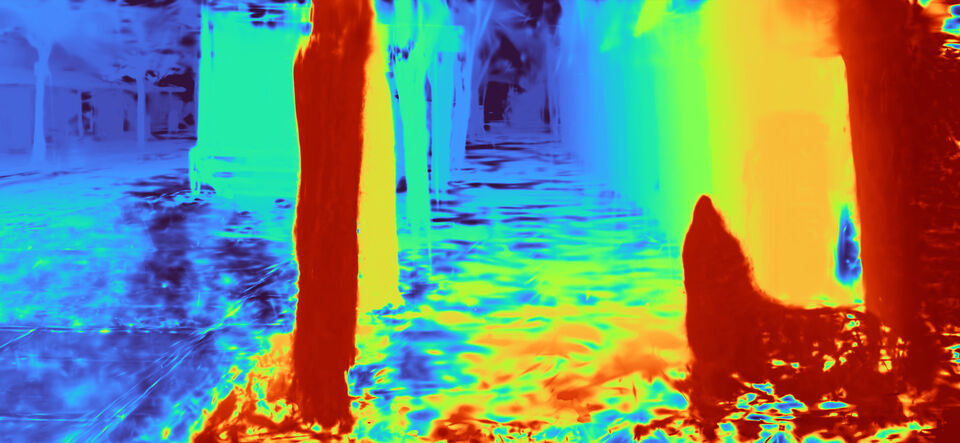} & 
            \includegraphics[width=0.245\linewidth]{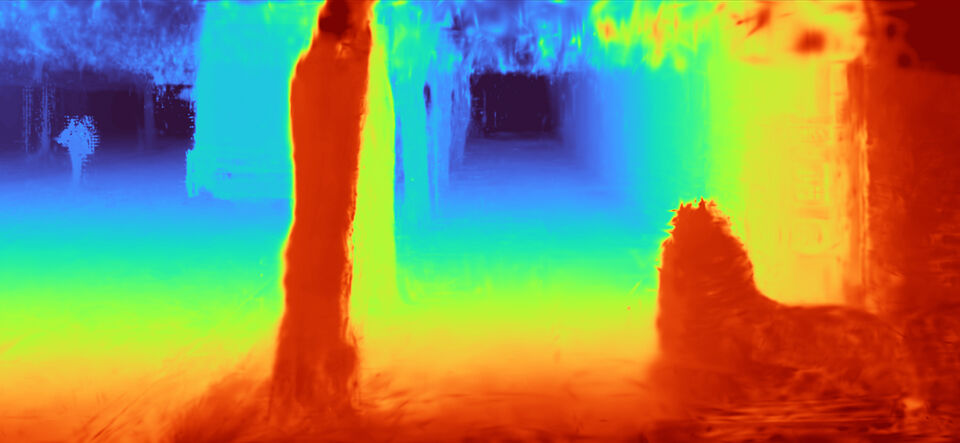} \\

        \includegraphics[width=0.245\linewidth]{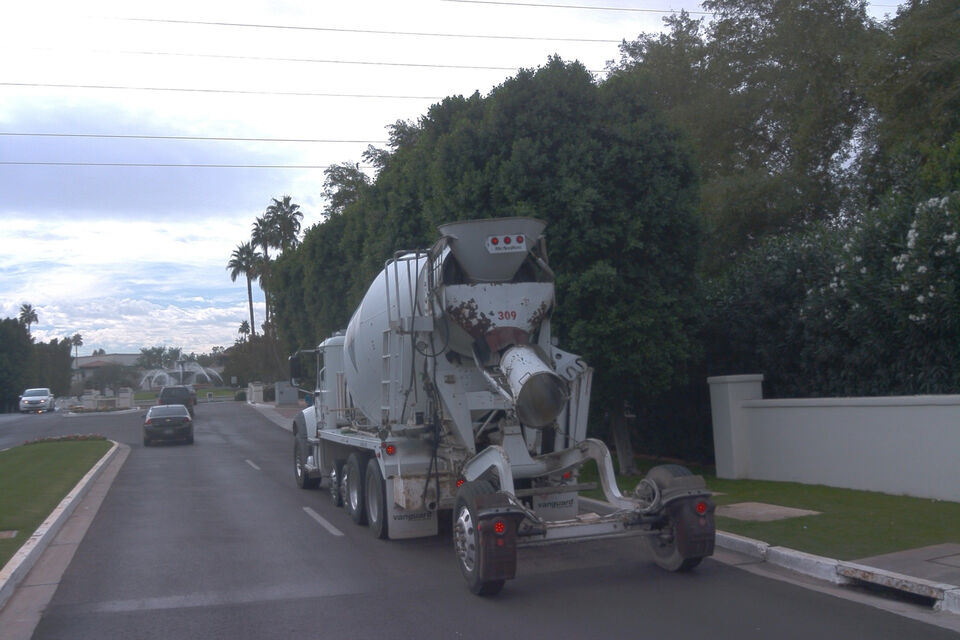} & 
            \includegraphics[width=0.245\linewidth]{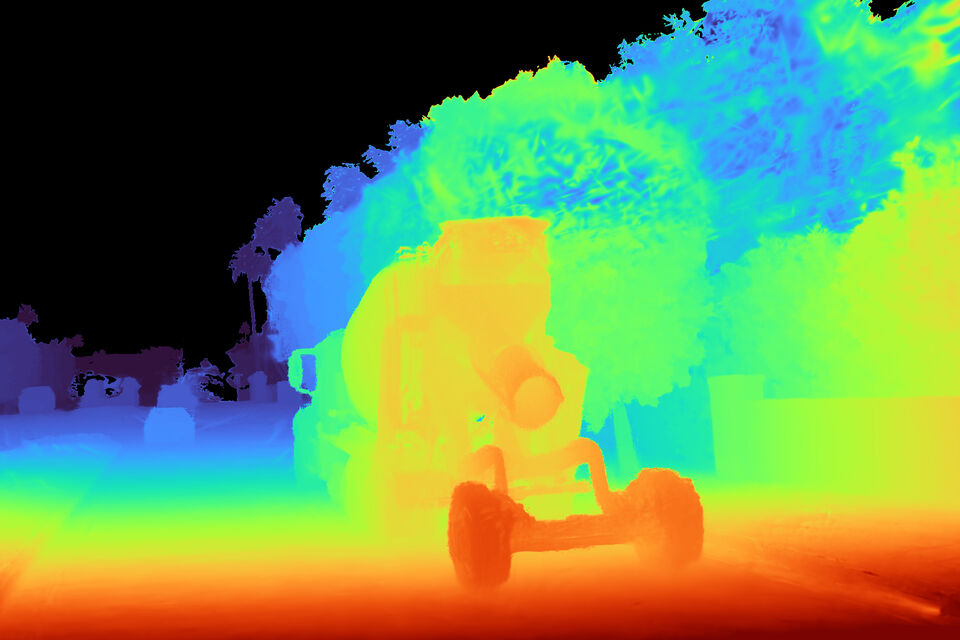} & 
            \includegraphics[width=0.245\linewidth]{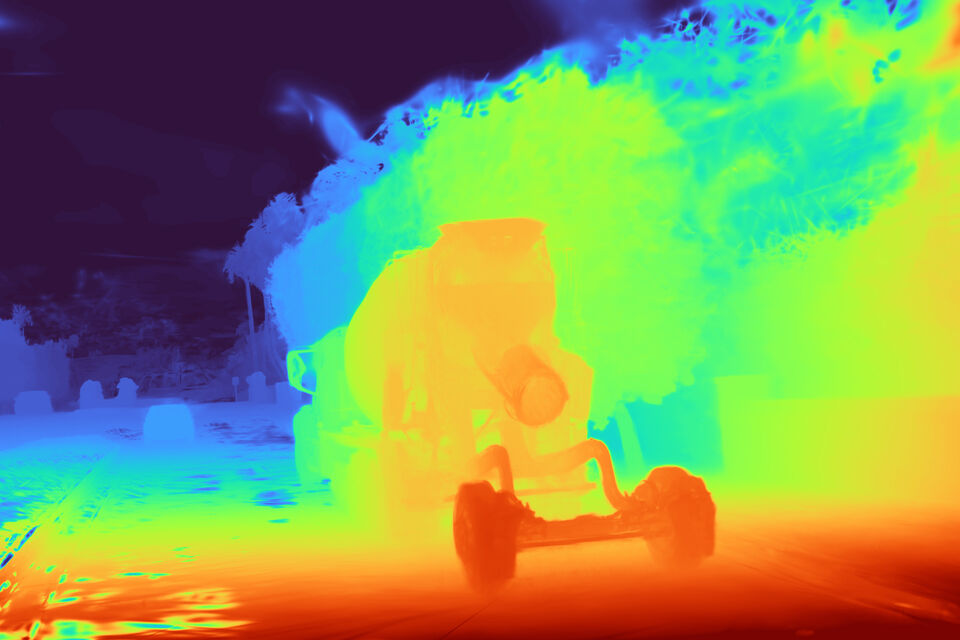} & 
            \includegraphics[width=0.245\linewidth]{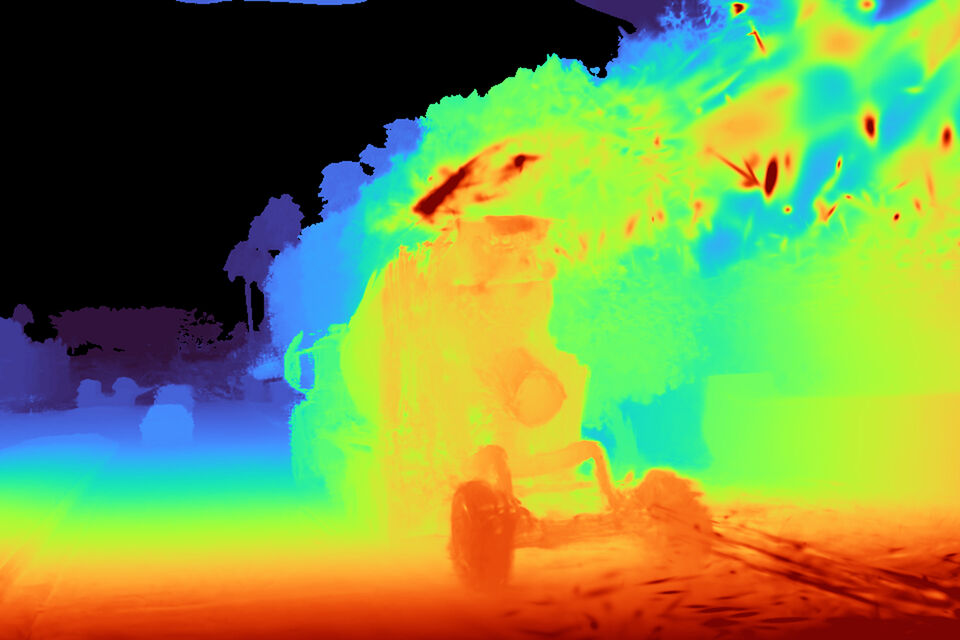} \\

        \includegraphics[width=0.245\linewidth]{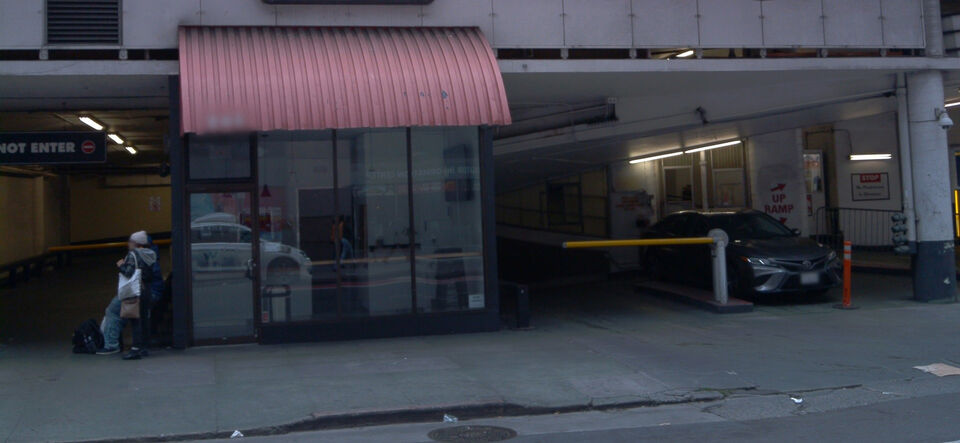} & 
            \includegraphics[width=0.245\linewidth]{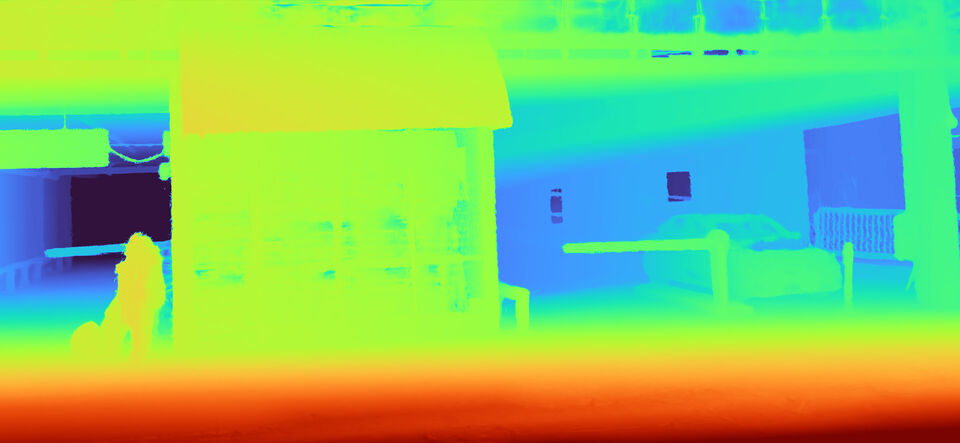} & 
            \includegraphics[width=0.245\linewidth]{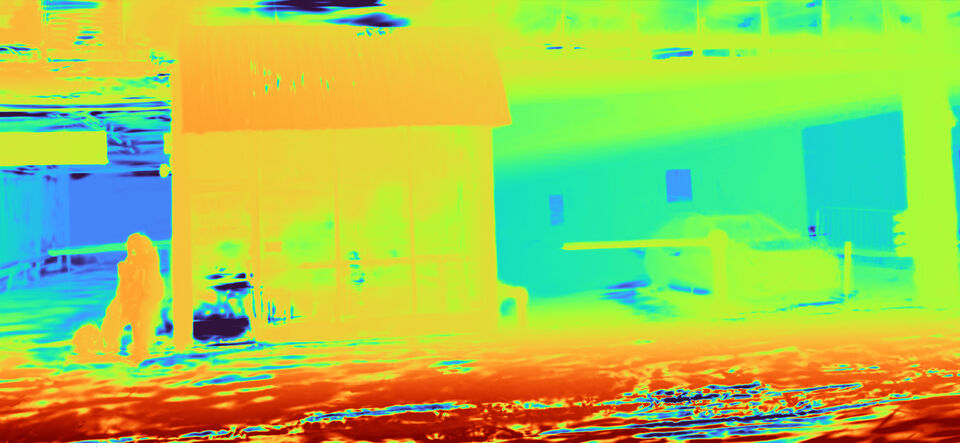} & 
            \includegraphics[width=0.245\linewidth]{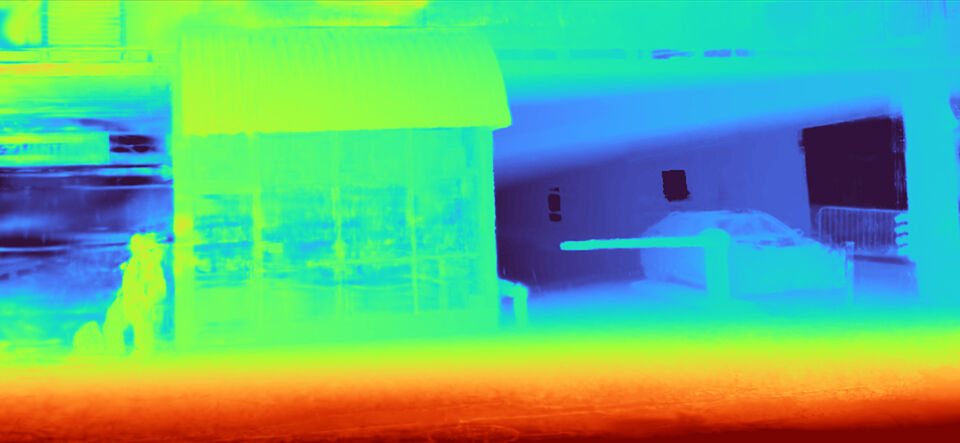} \\
    
    \end{tabular}
    \caption{Qualitative comparison of rendered depth maps with previous methods on Waymo dataset.}\label{fig:depth2}
\end{figure*}

\begin{figure*}
    \centering
    \setlength{\tabcolsep}{1pt}
    \renewcommand{\arraystretch}{1.0}
    \begin{tabular}{ccc}
        
        \Large\textbf{XSIM, Ours} & \Large SplatAD & \Large OmniRE \\
        \normalsize
        
        \includegraphics[width=0.33\linewidth]{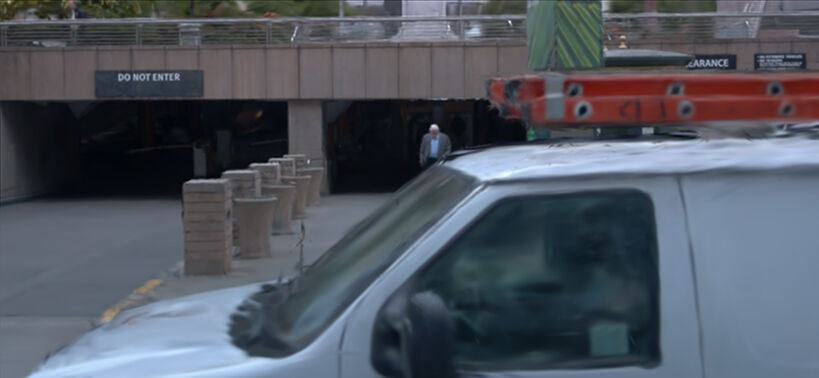} & 
            \includegraphics[width=0.33\linewidth]{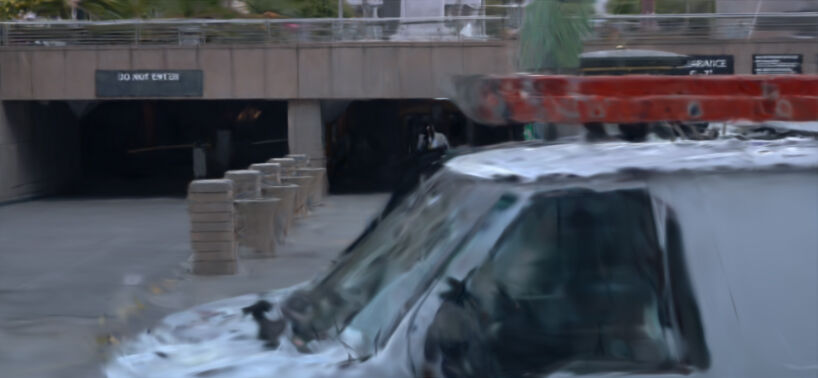} & 
            \includegraphics[width=0.33\linewidth]{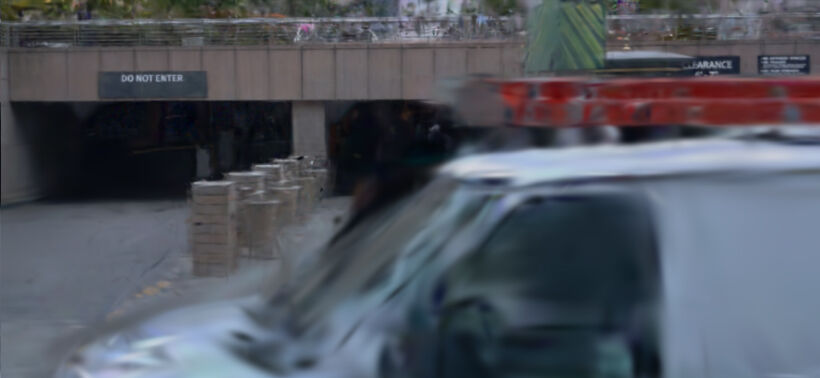} \\

        \includegraphics[width=0.33\linewidth]{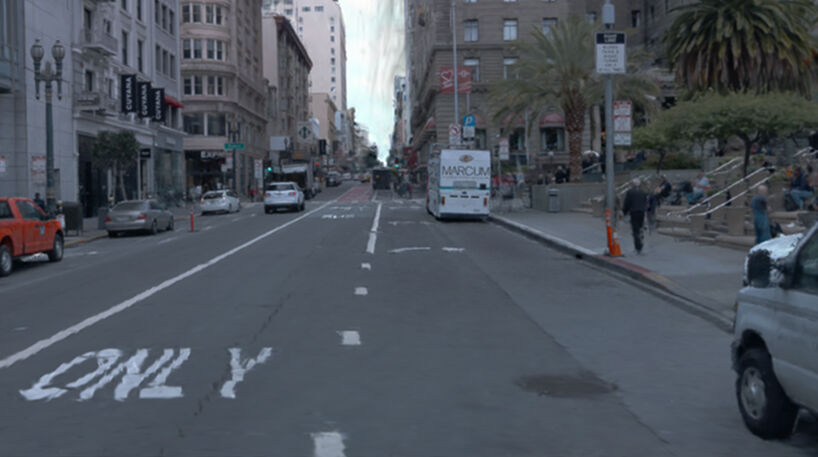} & 
            \includegraphics[width=0.33\linewidth]{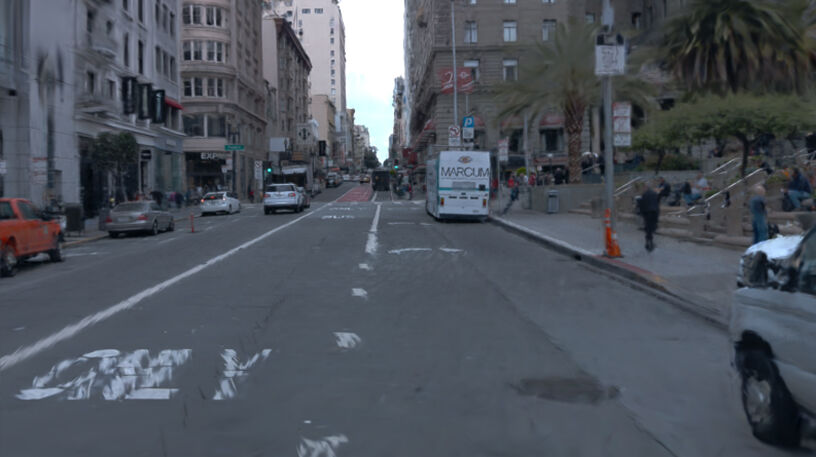} & 
            \includegraphics[width=0.33\linewidth]{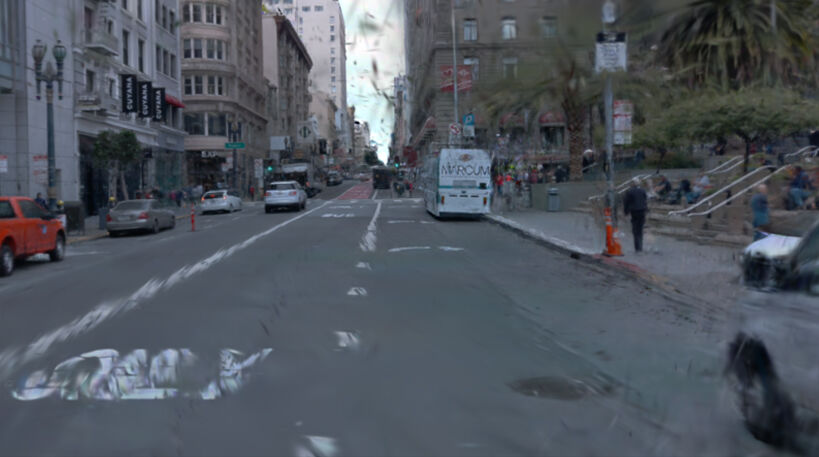} \\

        \includegraphics[width=0.33\linewidth]{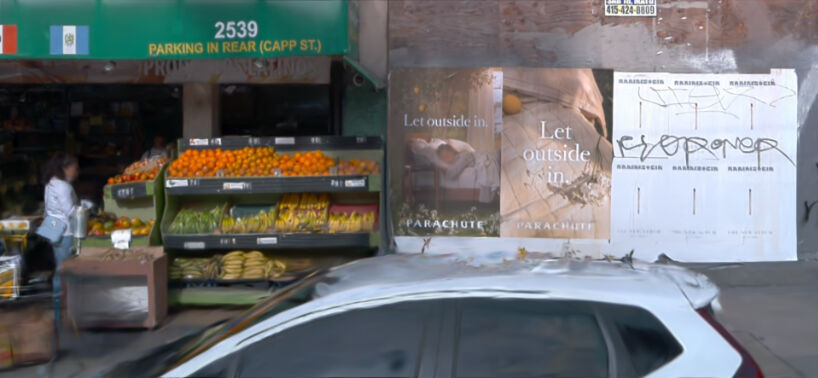} & 
            \includegraphics[width=0.33\linewidth]{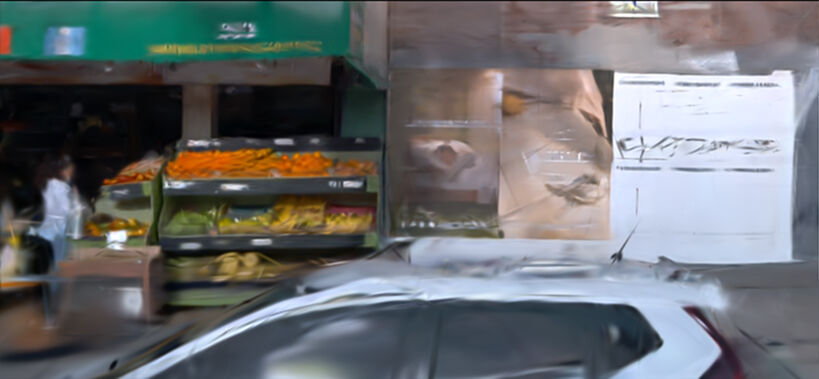} & 
            \includegraphics[width=0.33\linewidth]{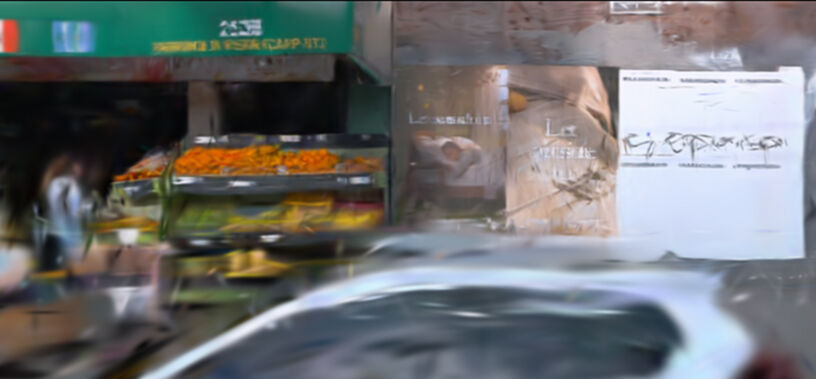} \\

        \includegraphics[width=0.33\linewidth]{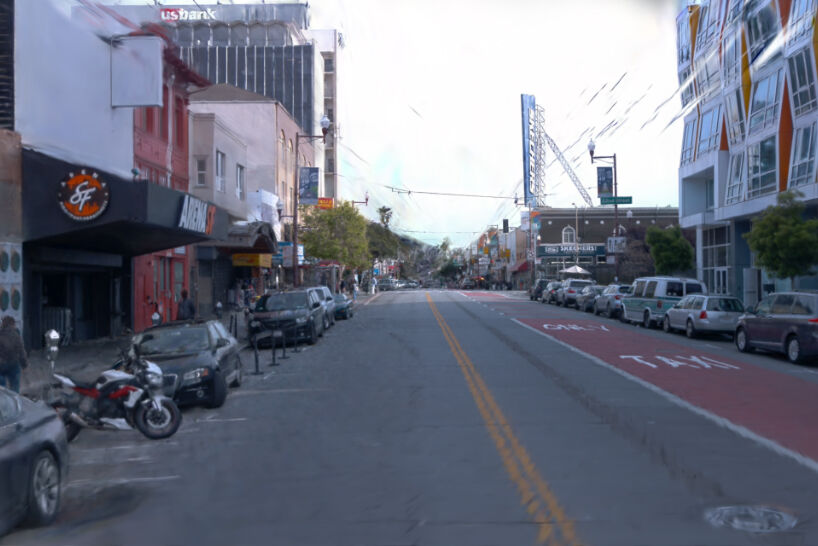} & 
            \includegraphics[width=0.33\linewidth]{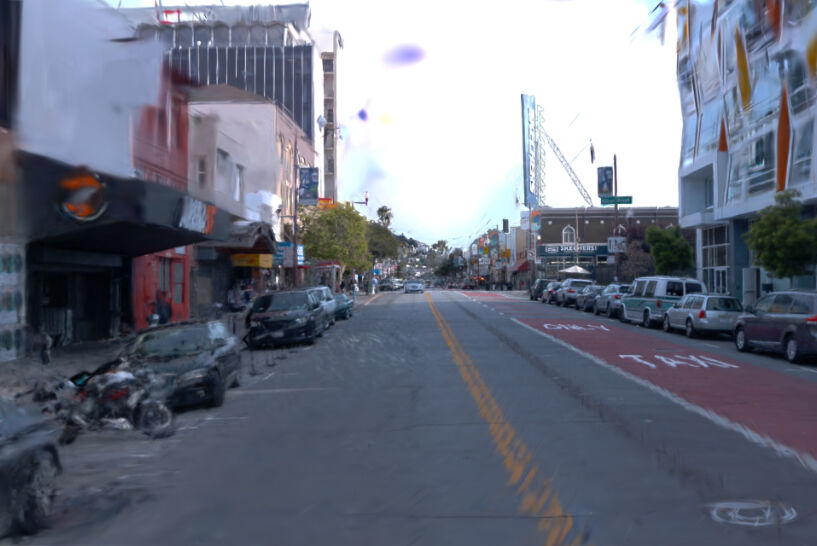} & 
            \includegraphics[width=0.33\linewidth]{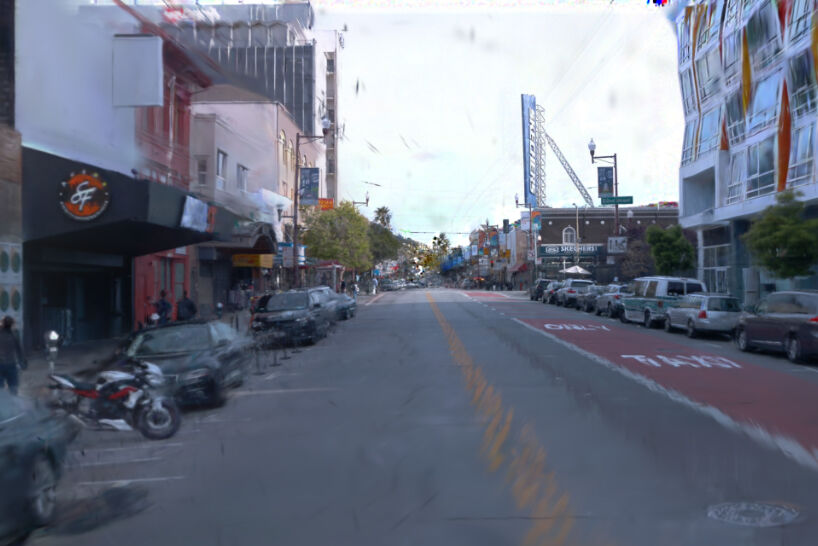} \\

        \includegraphics[width=0.33\linewidth]{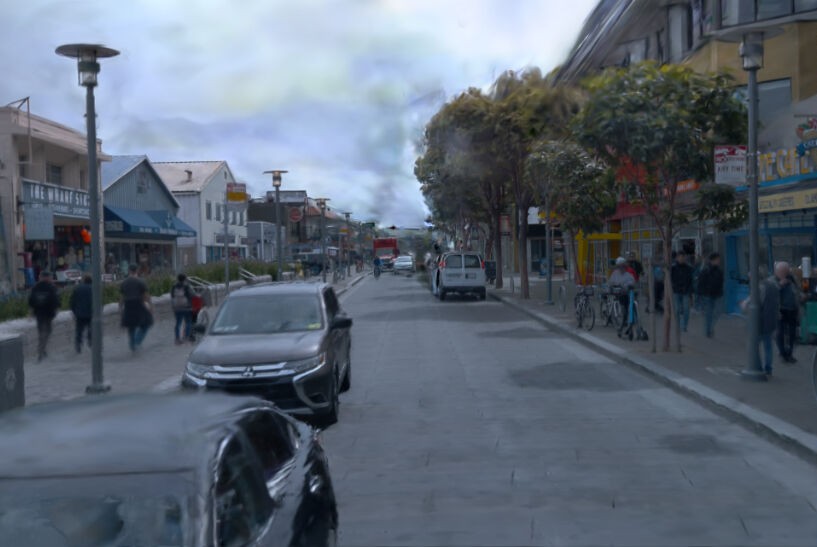} & 
            \includegraphics[width=0.33\linewidth]{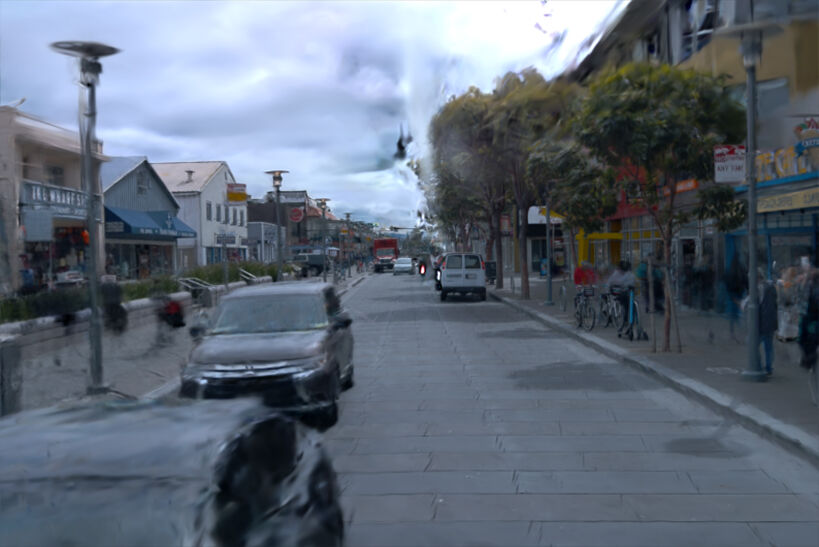} & 
            \includegraphics[width=0.33\linewidth]{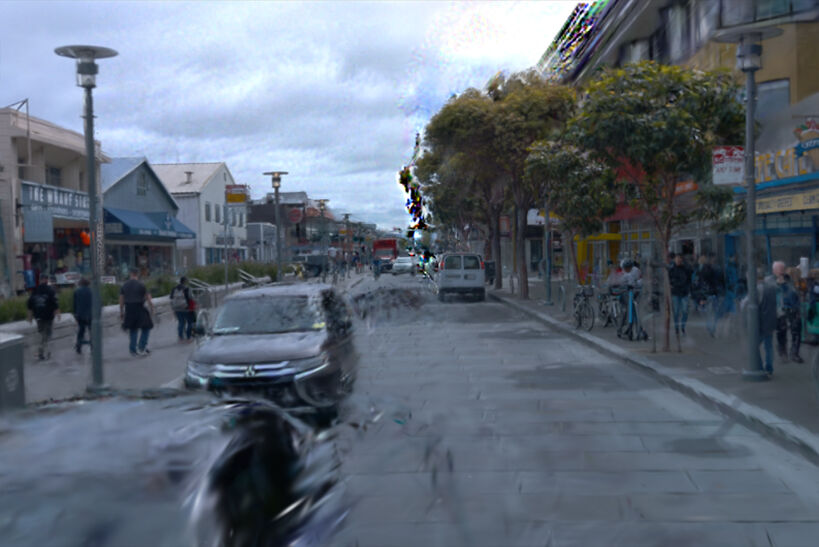} \\

        \includegraphics[width=0.33\linewidth]{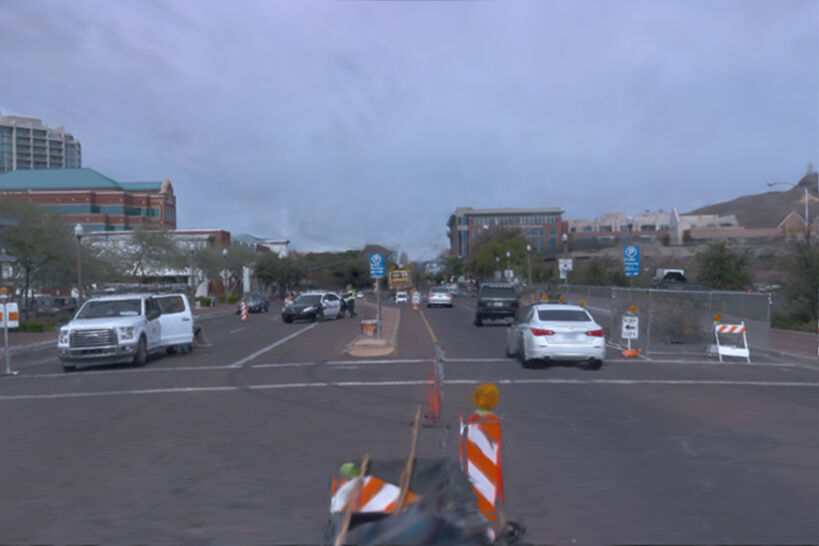} & 
            \includegraphics[width=0.33\linewidth]{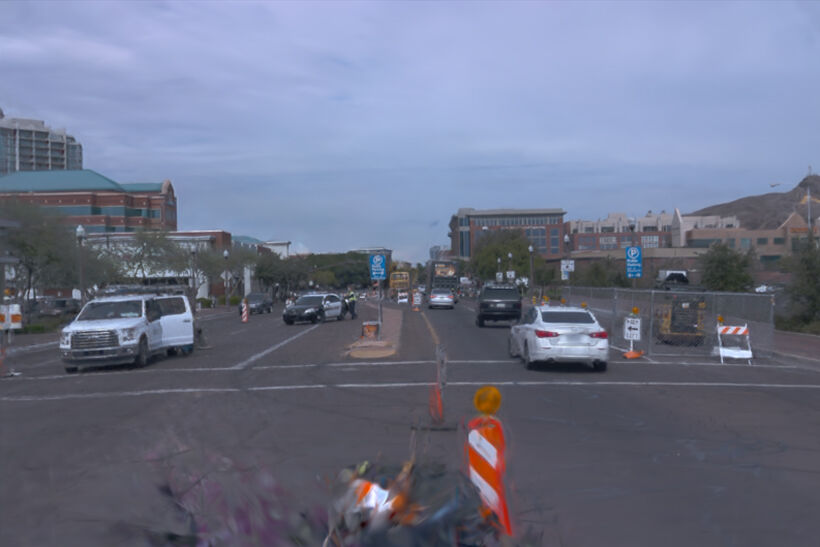} & 
            \includegraphics[width=0.33\linewidth]{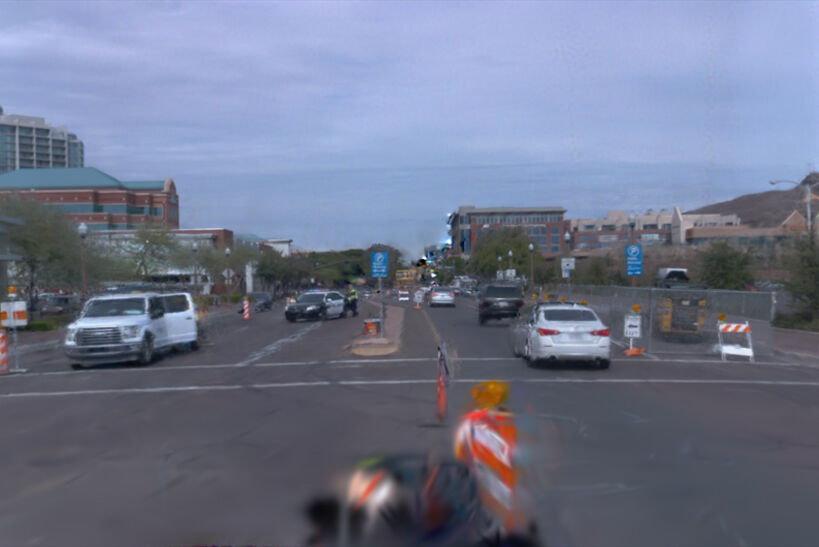} \\
        
    \end{tabular}
    \caption{Lane shift 3m visualizations on Waymo dataset. XSIM provides consistent appearance renderings without floating and blurring artifacts on both static and dynamic objects.}\label{fig:nvs}
\end{figure*}

\begin{figure*}
    \centering
    \setlength{\tabcolsep}{1pt}
    \renewcommand{\arraystretch}{1.0}
    \begin{tabular}{cccc}
        
        \Large Ground-truth & \Large\textbf{XSIM, Ours} & \Large SplatAD & \Large OmniRE \\
        \normalsize

        \includegraphics[width=0.245\linewidth]{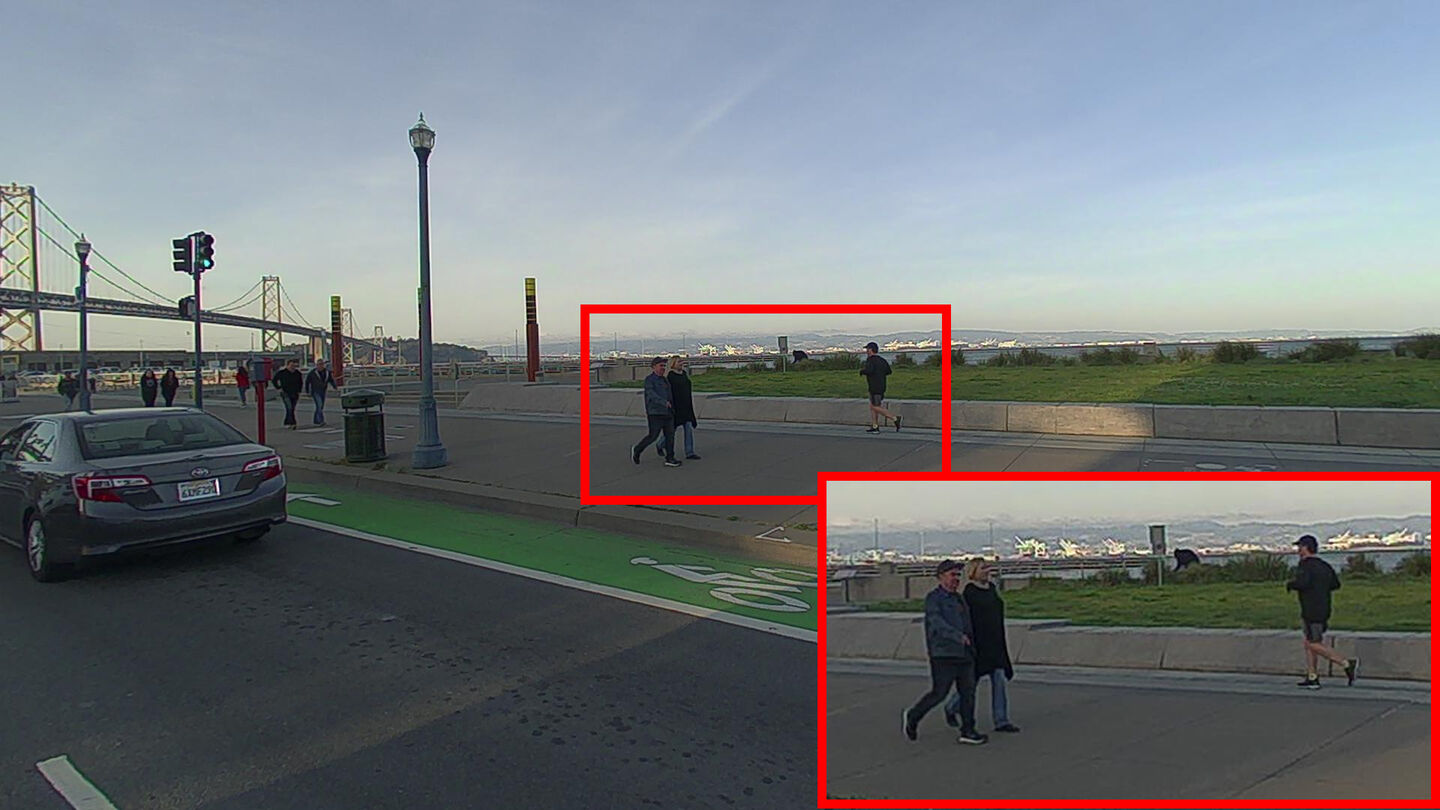} & 
            \includegraphics[width=0.245\linewidth]{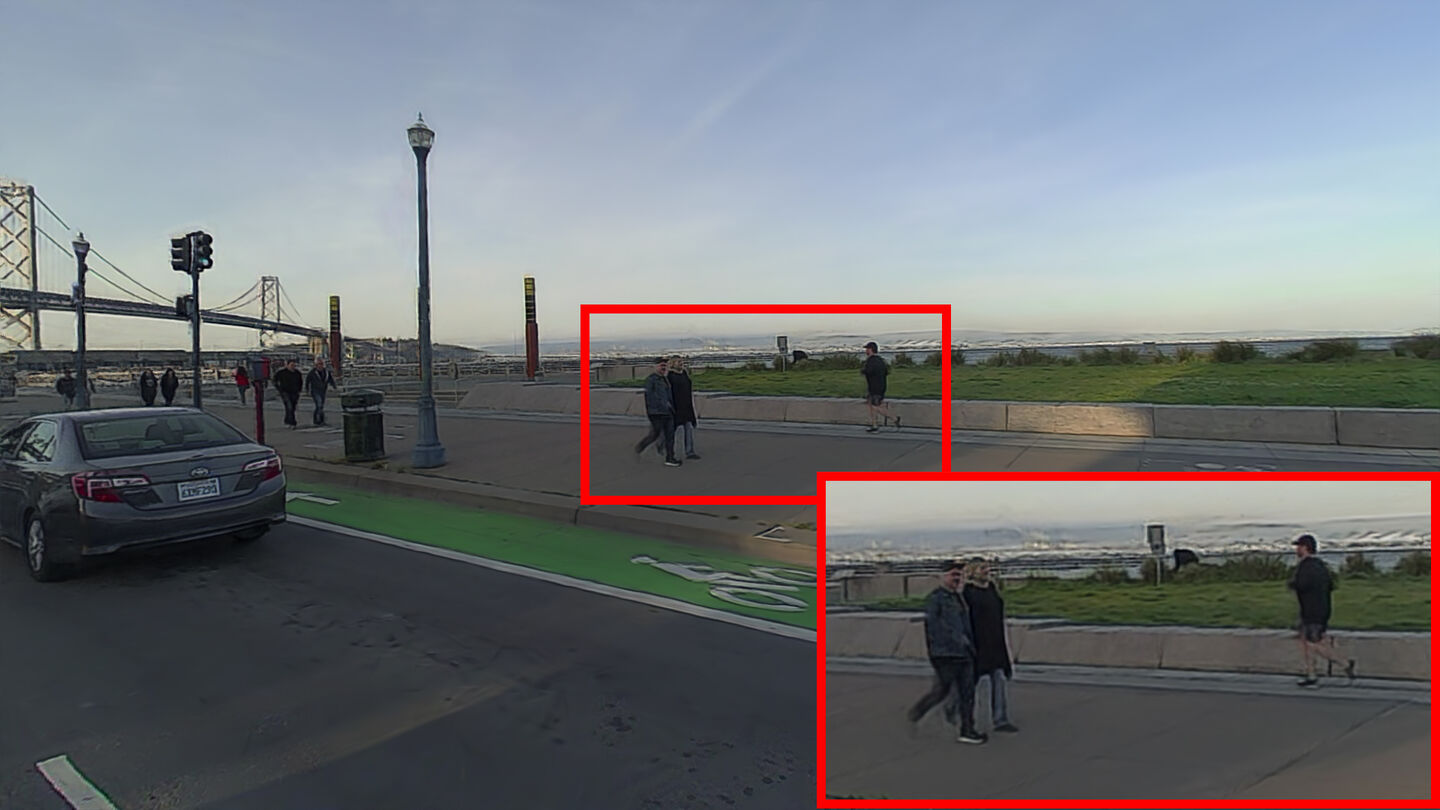} & 
            \includegraphics[width=0.245\linewidth]{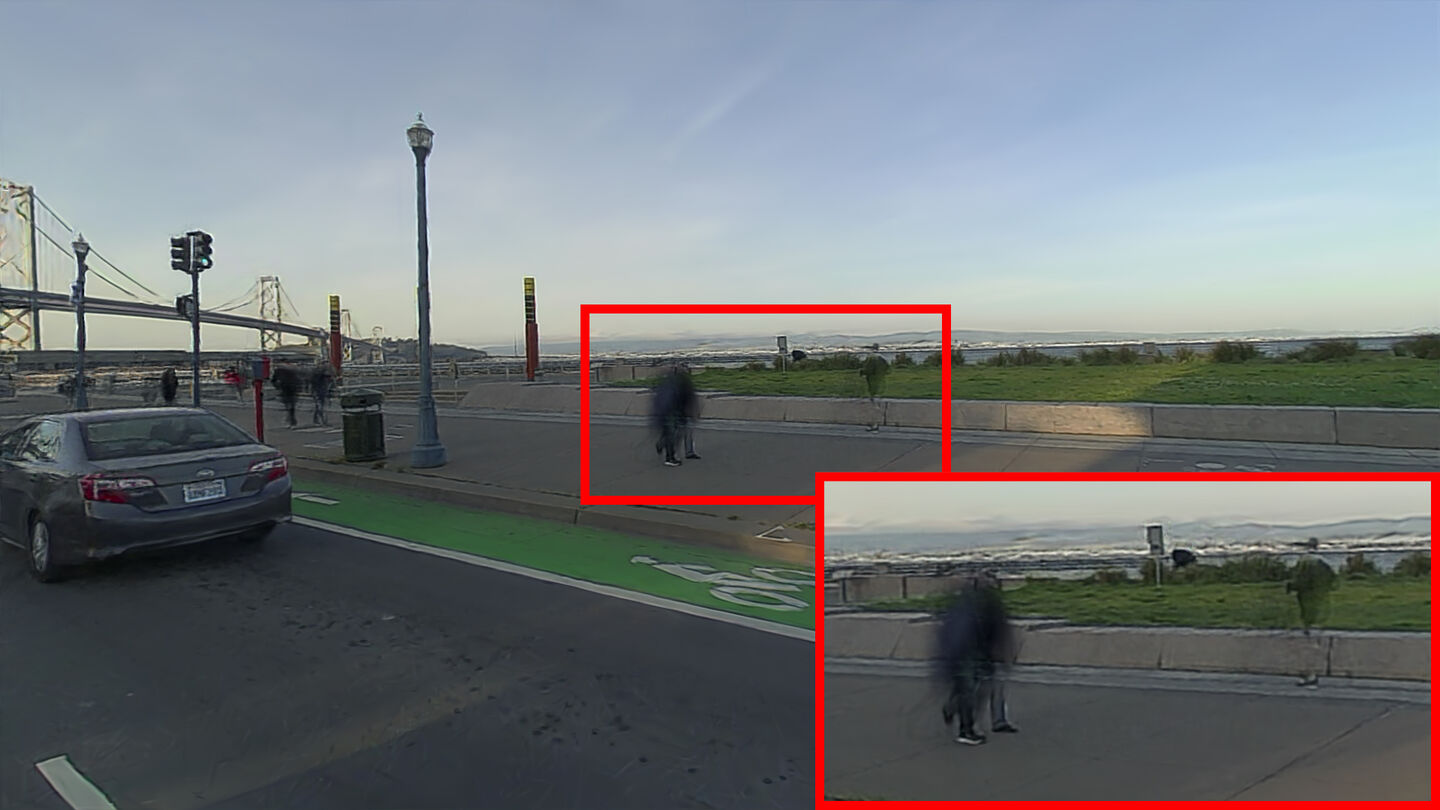} & 
            \includegraphics[width=0.245\linewidth]{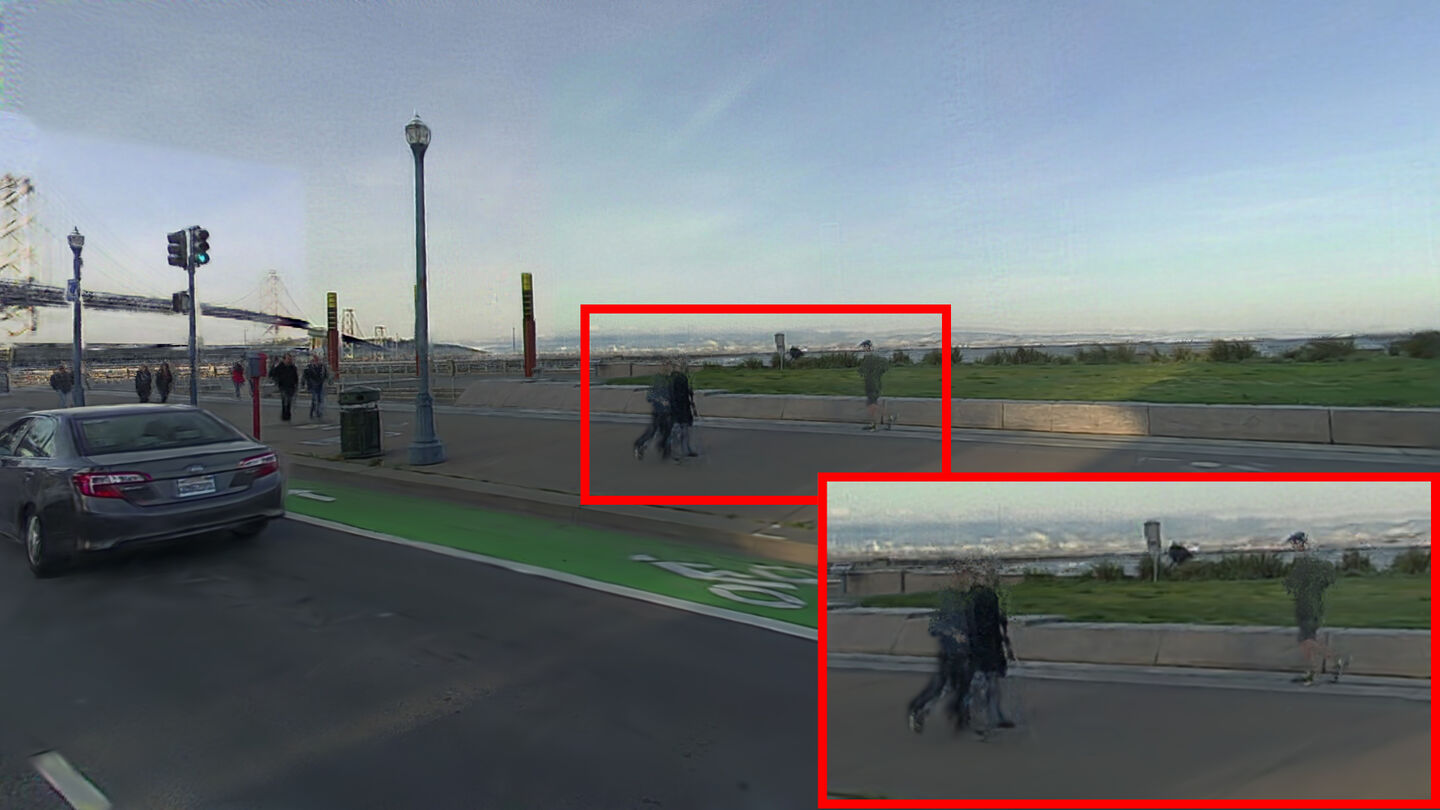} \\

        \includegraphics[width=0.245\linewidth]{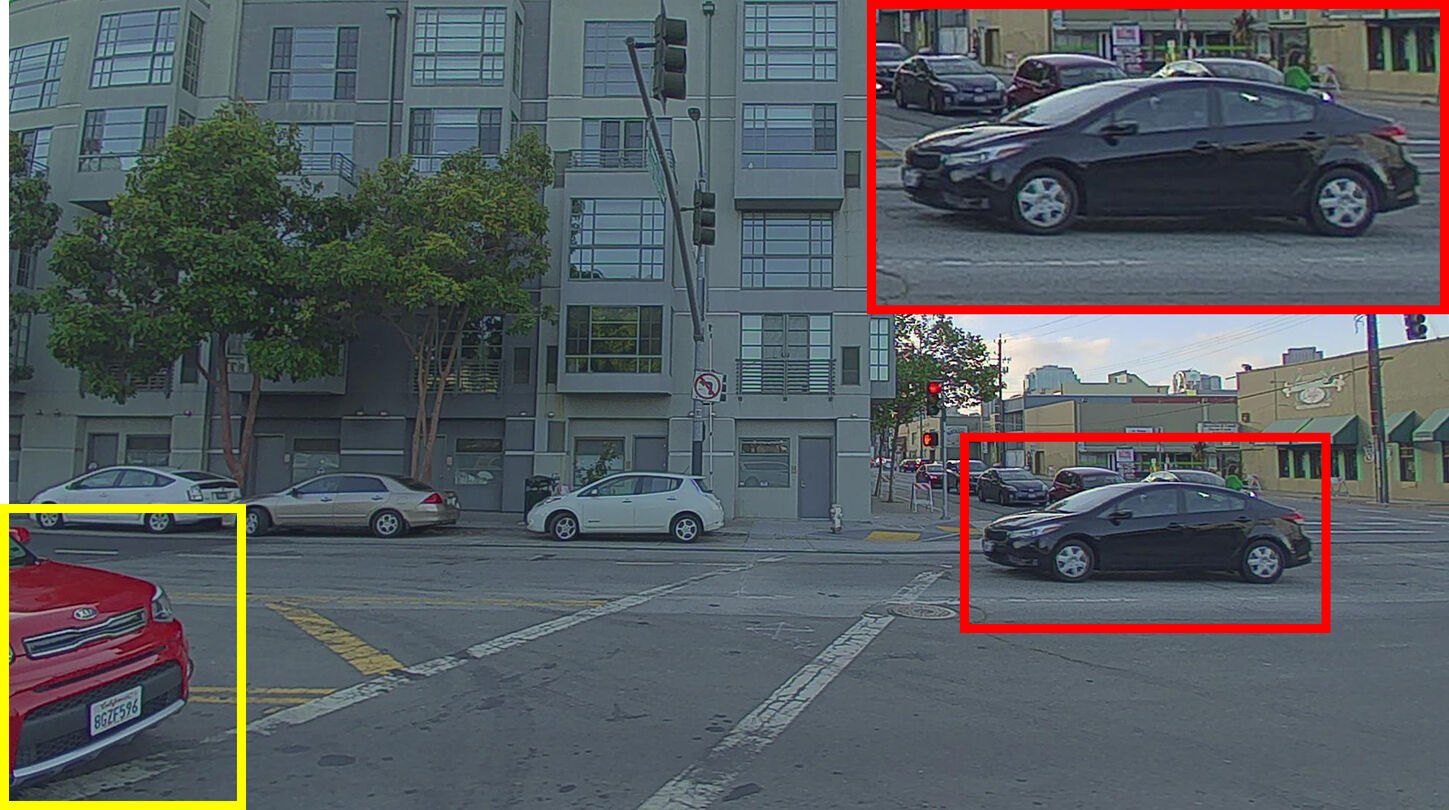} & 
            \includegraphics[width=0.245\linewidth]{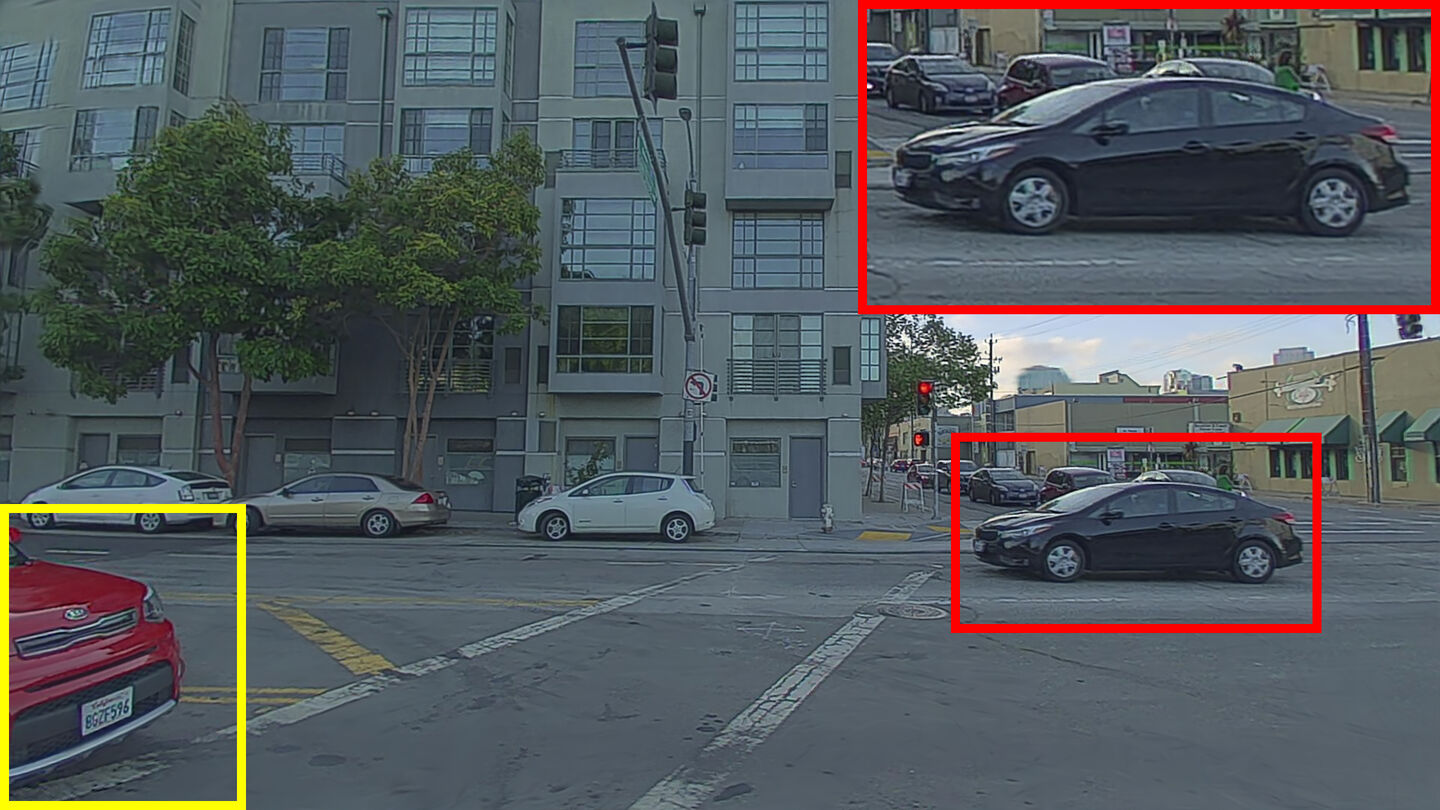} & 
            \includegraphics[width=0.245\linewidth]{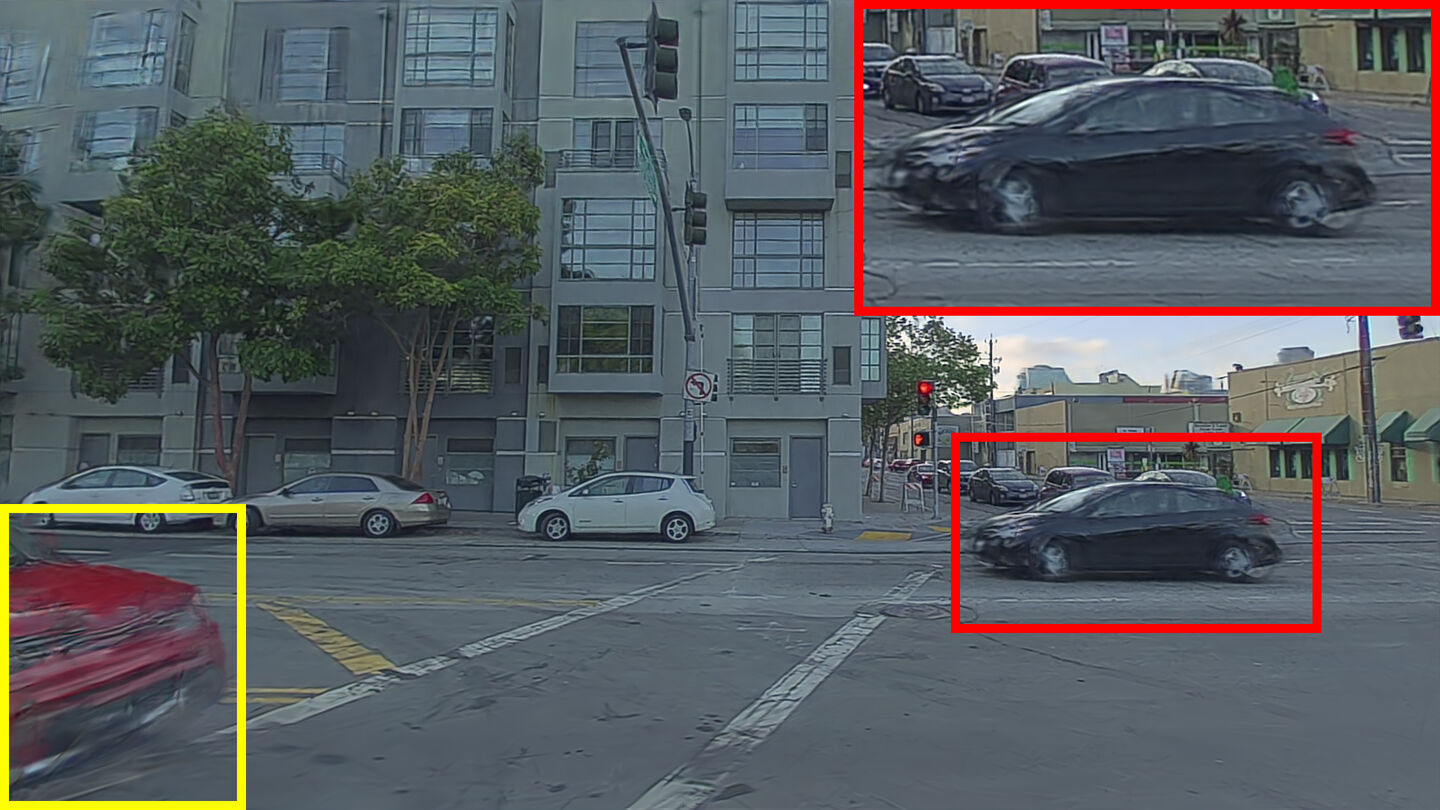} & 
            \includegraphics[width=0.245\linewidth]{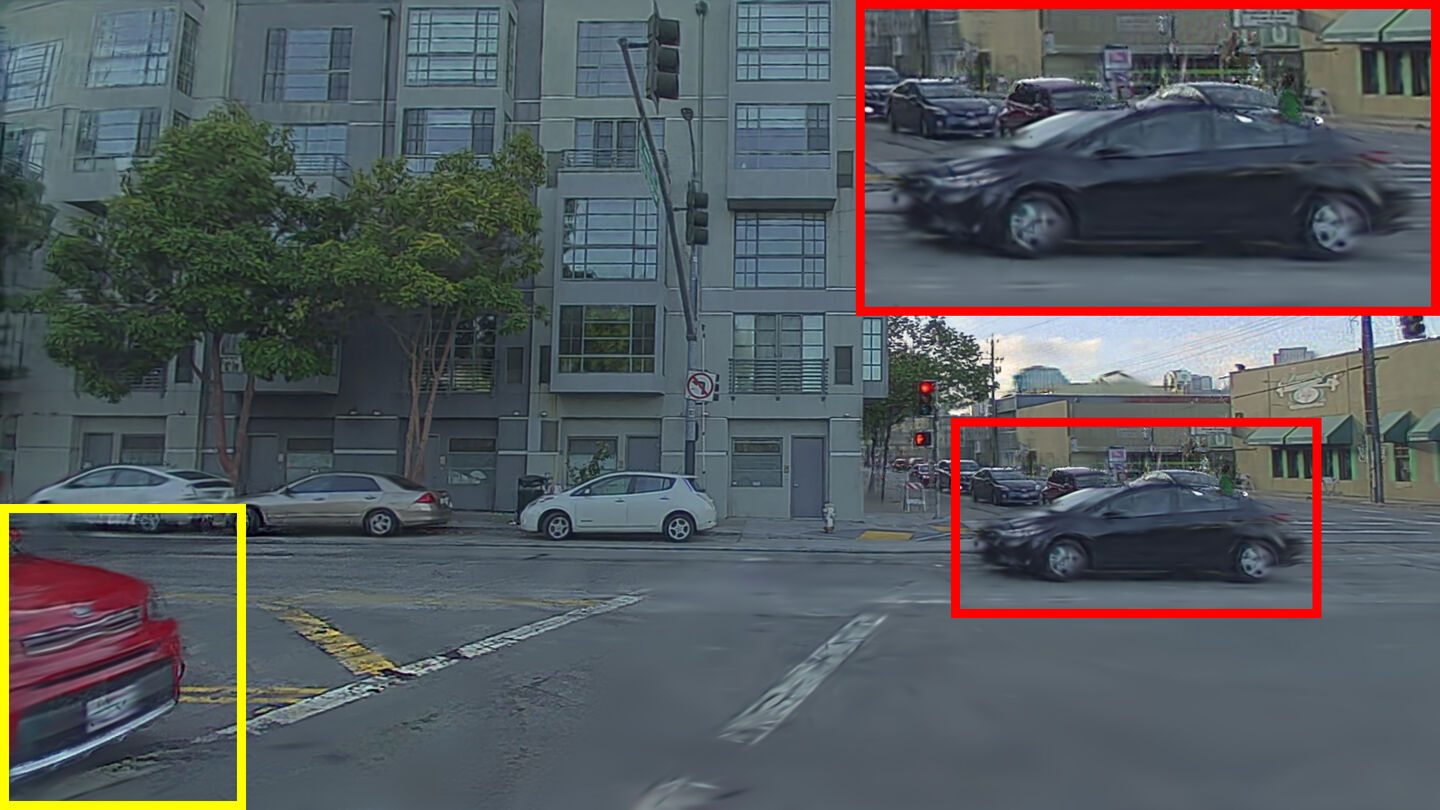} \\

        \includegraphics[width=0.245\linewidth]{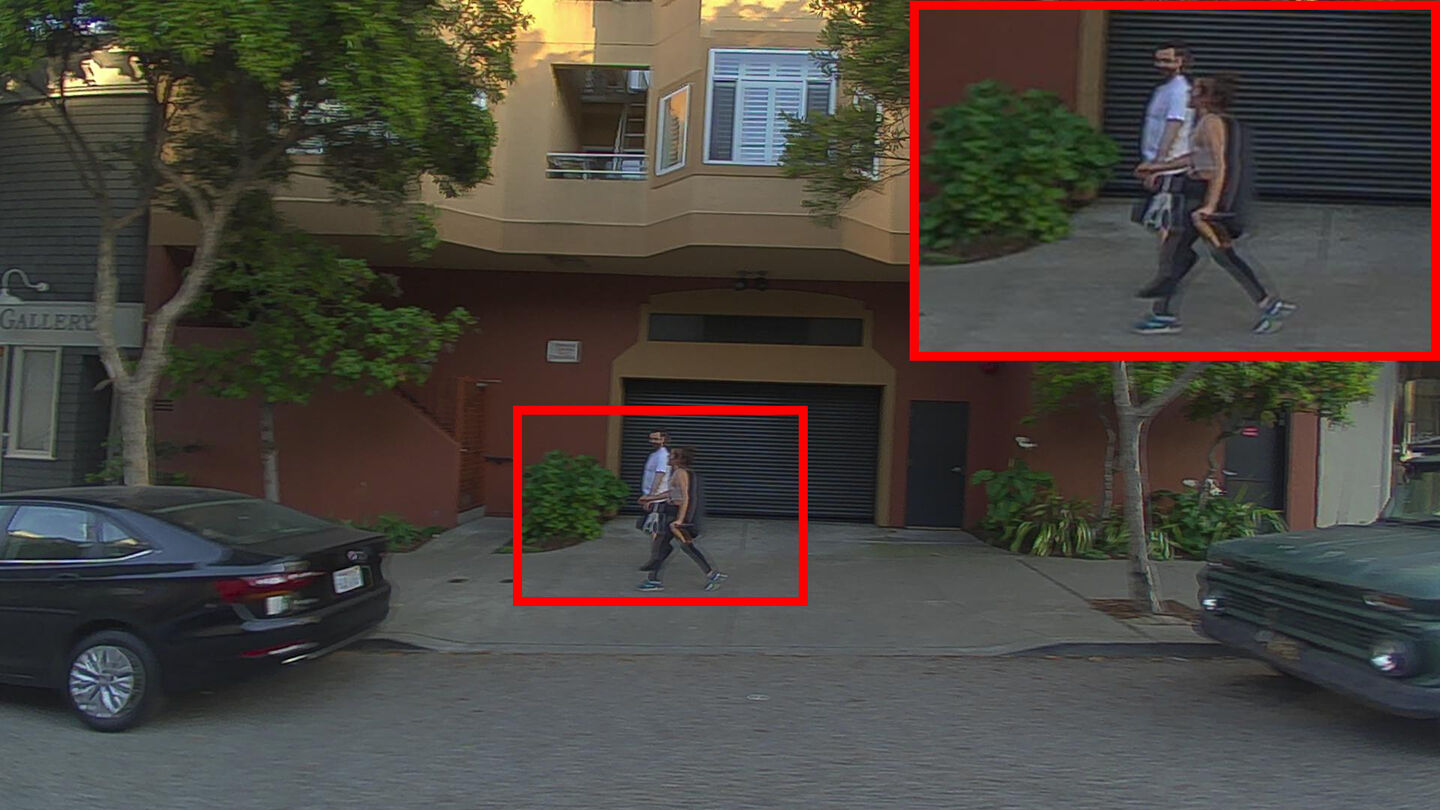} & 
            \includegraphics[width=0.245\linewidth]{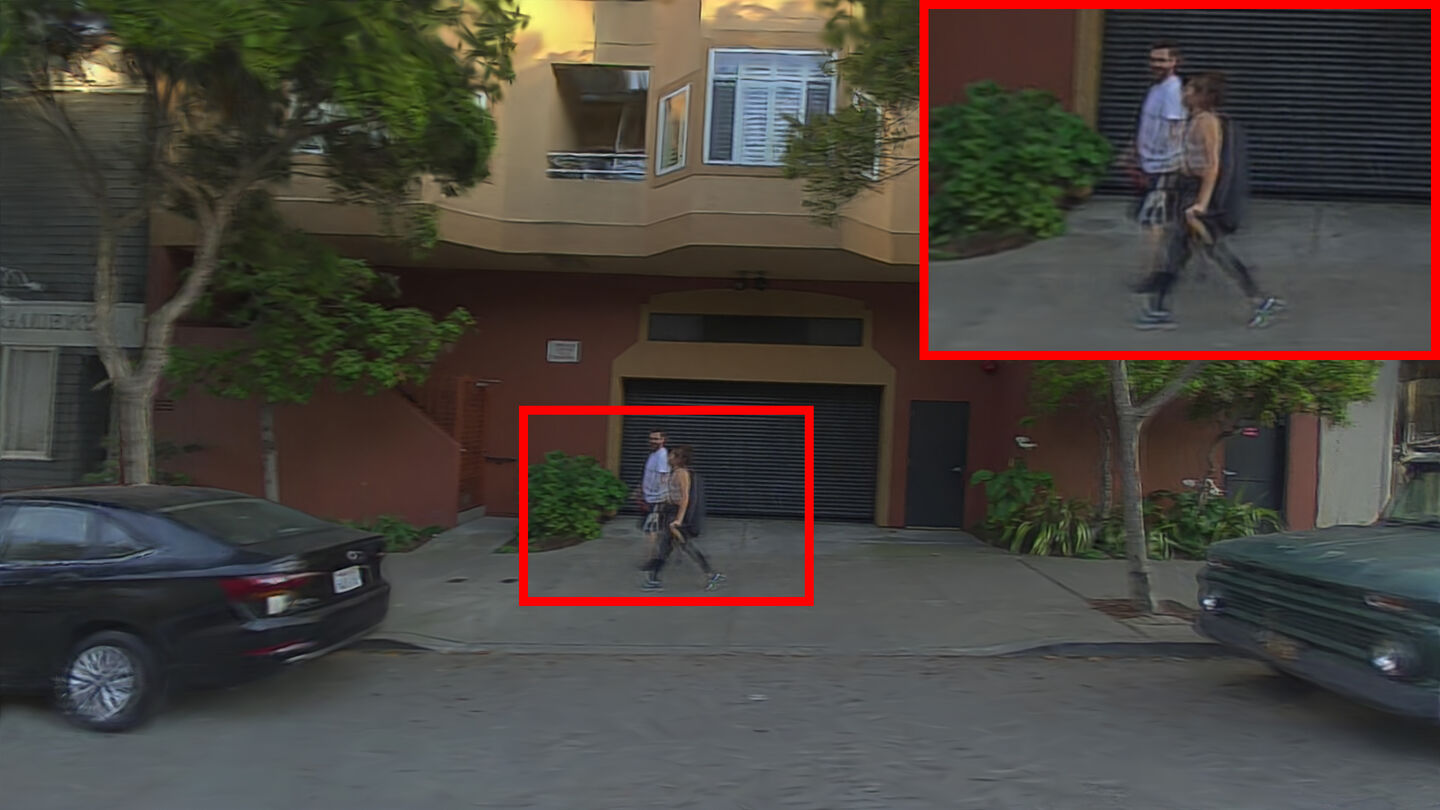} & 
            \includegraphics[width=0.245\linewidth]{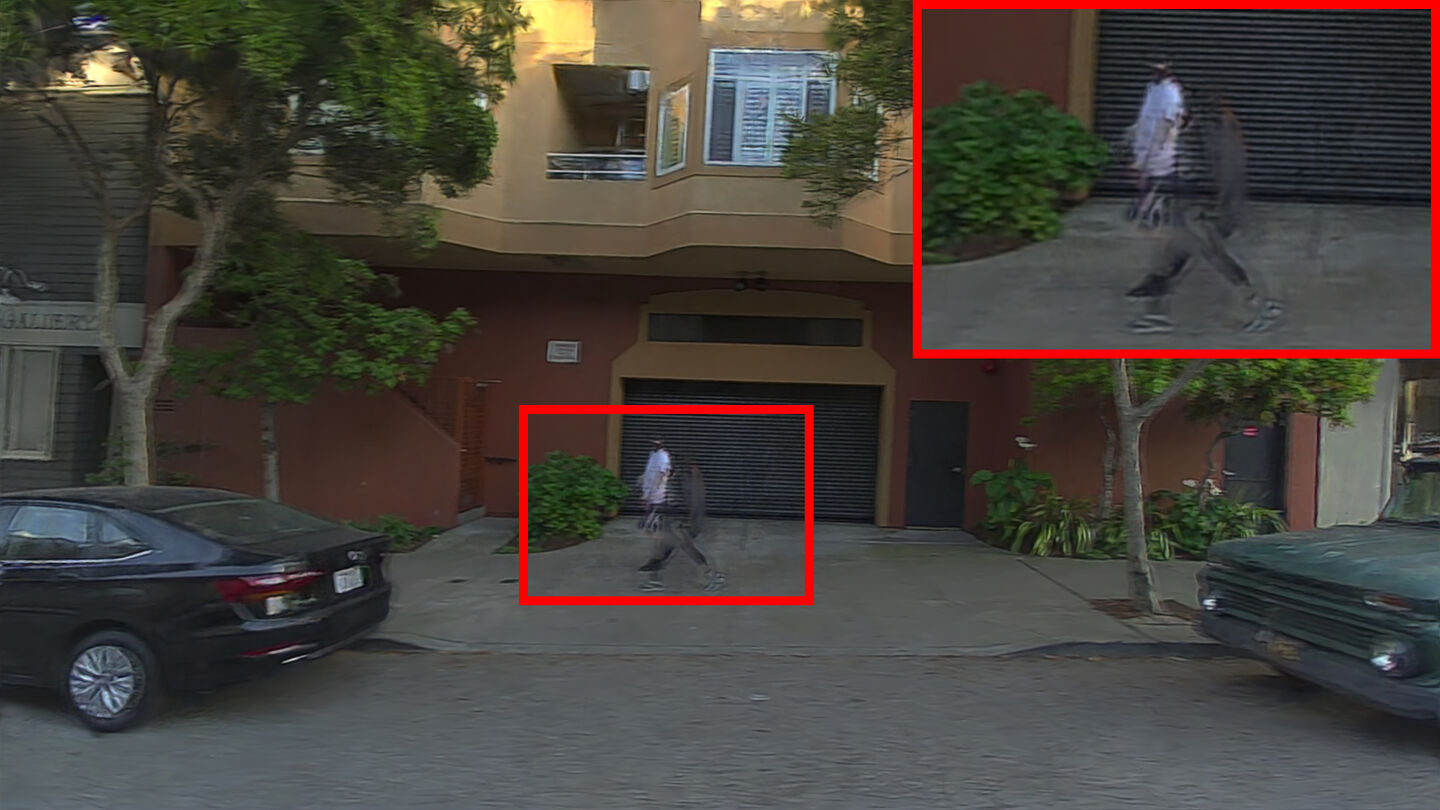} & 
            \includegraphics[width=0.245\linewidth]{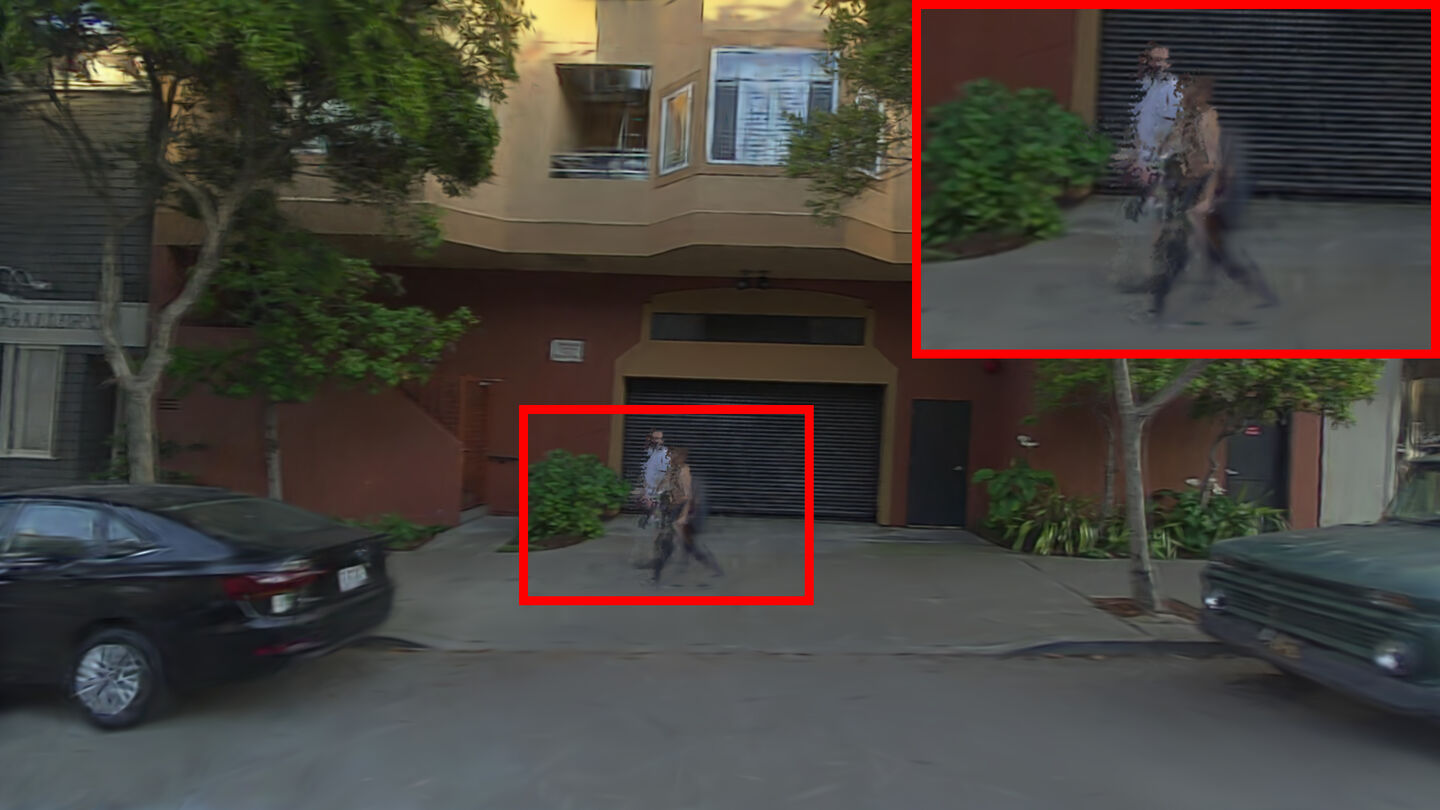} \\

        \includegraphics[width=0.245\linewidth]{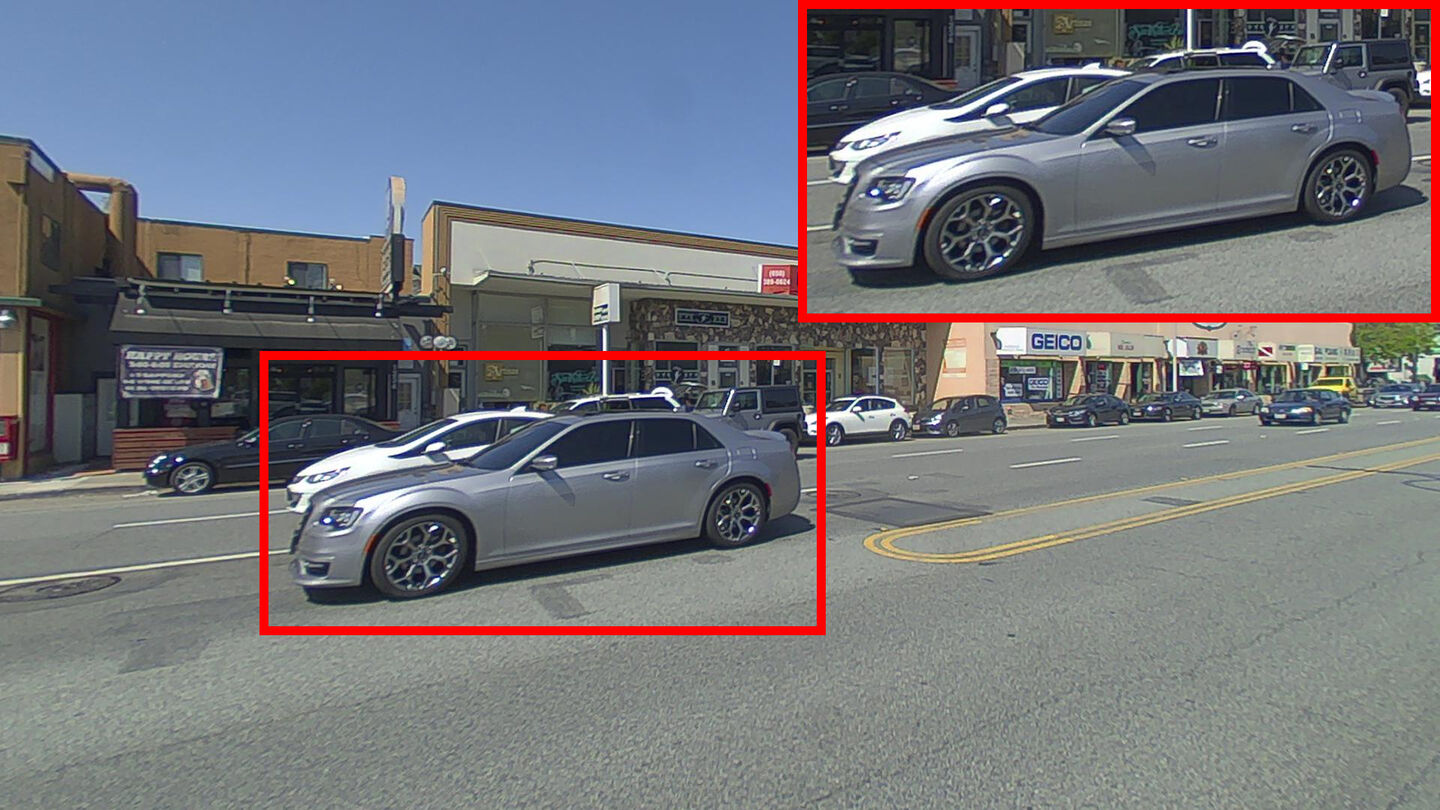} & 
            \includegraphics[width=0.245\linewidth]{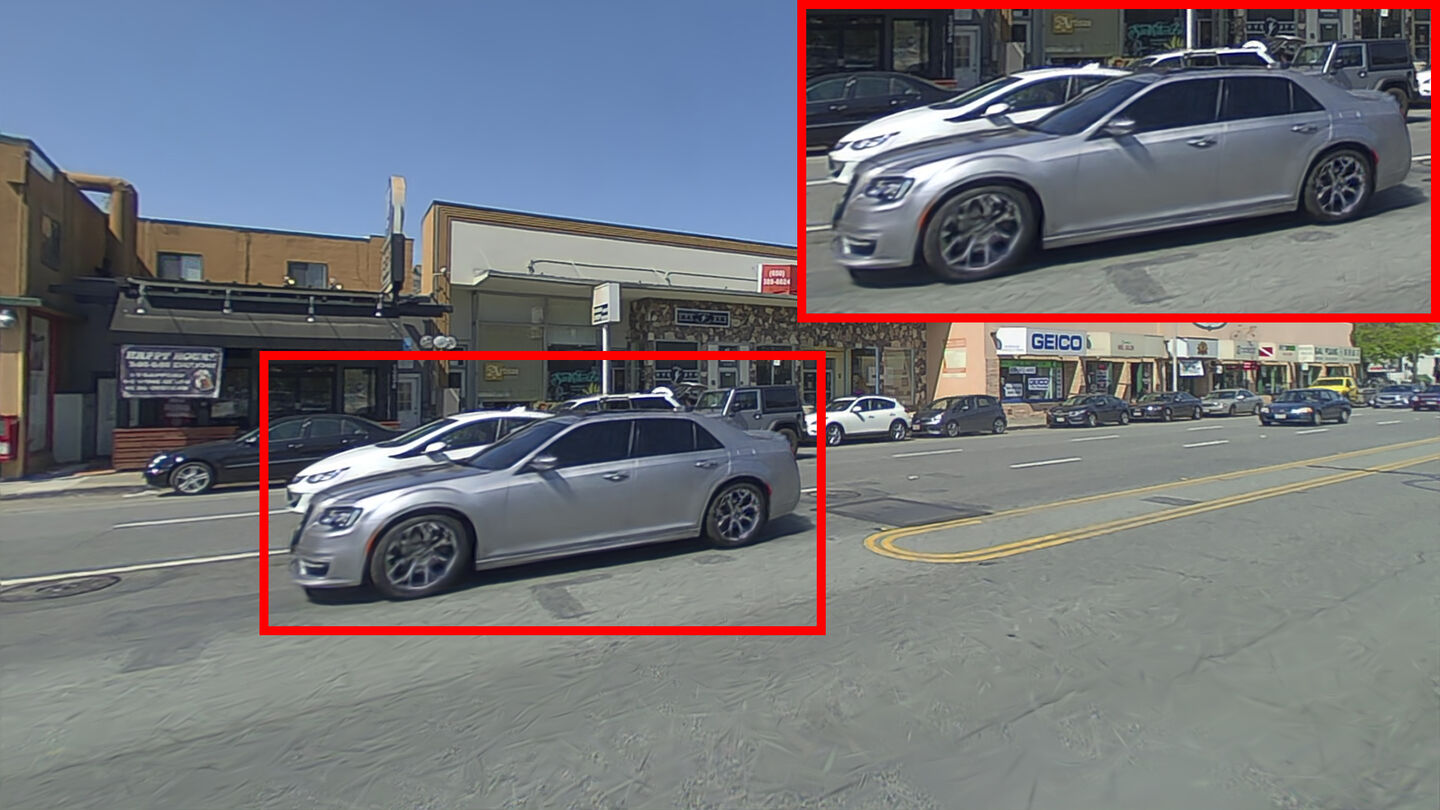} & 
            \includegraphics[width=0.245\linewidth]{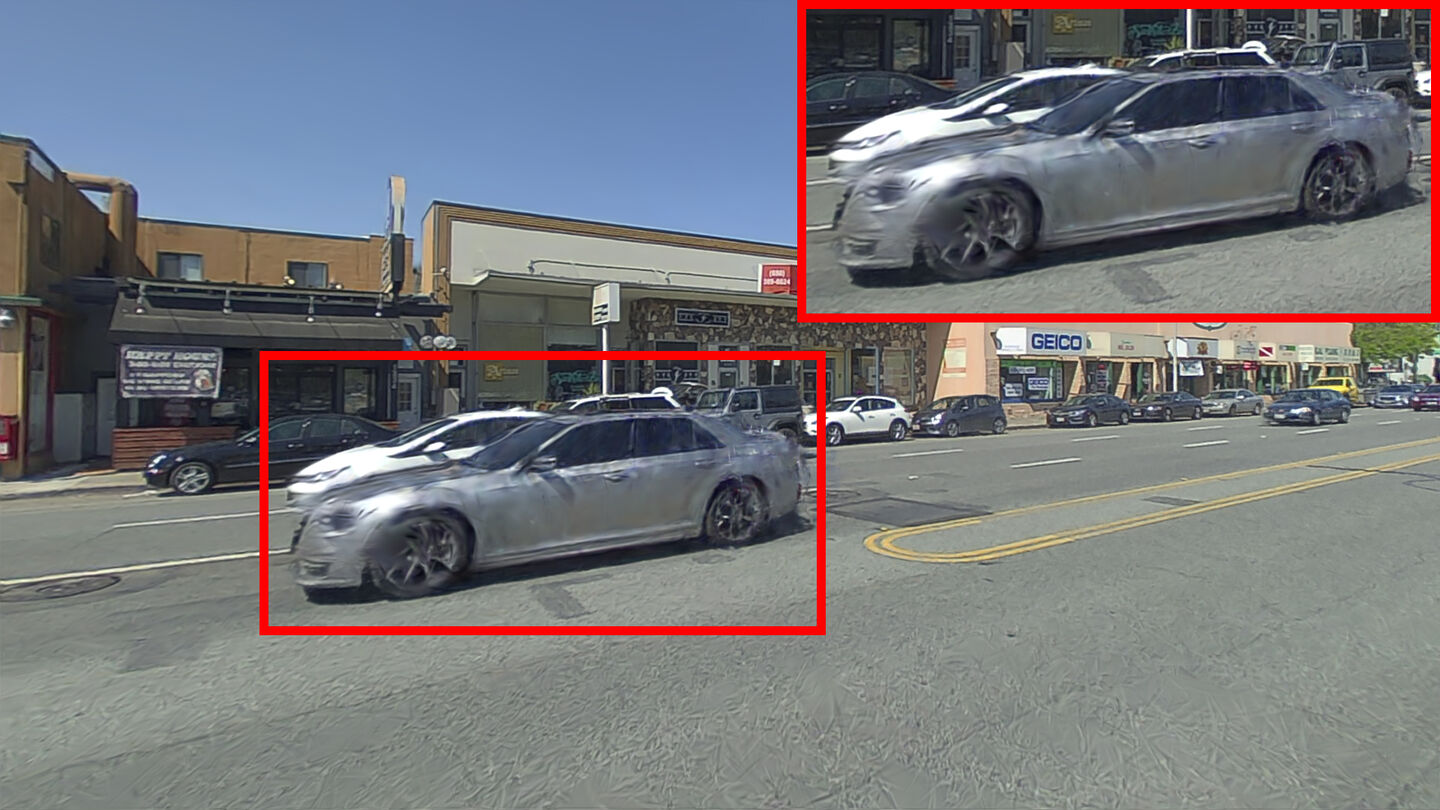} & 
            \includegraphics[width=0.245\linewidth]{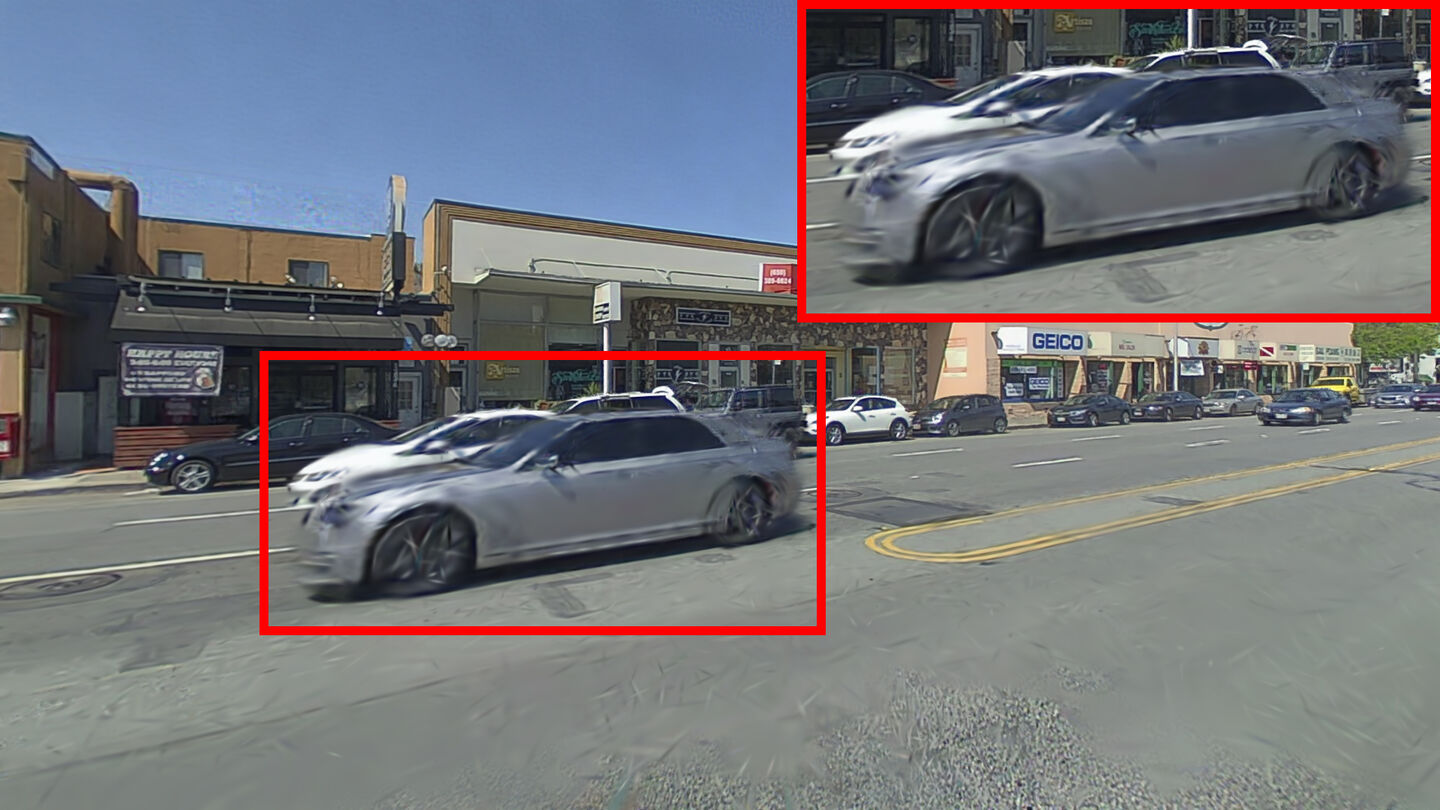} \\

        \includegraphics[width=0.245\linewidth]{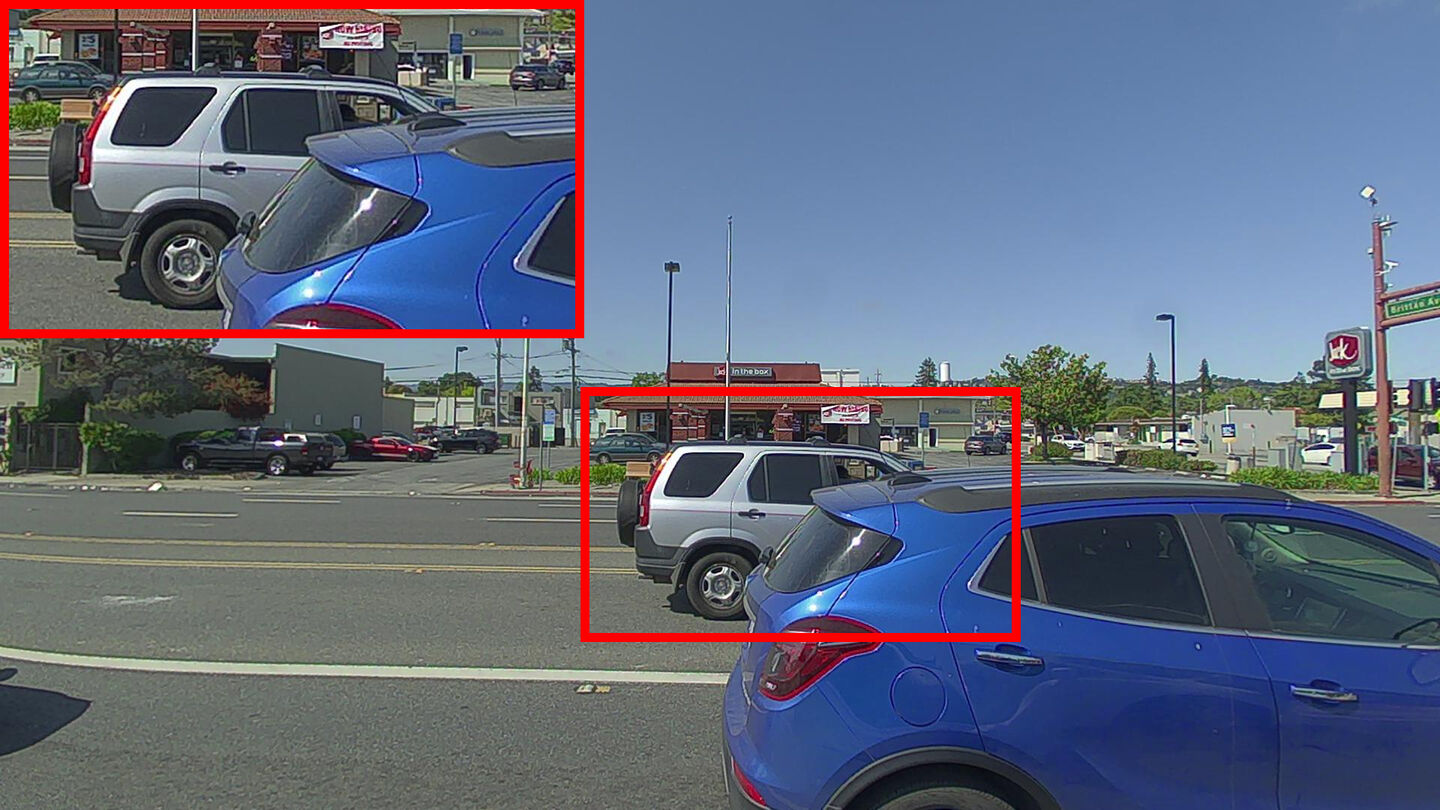} & 
            \includegraphics[width=0.245\linewidth]{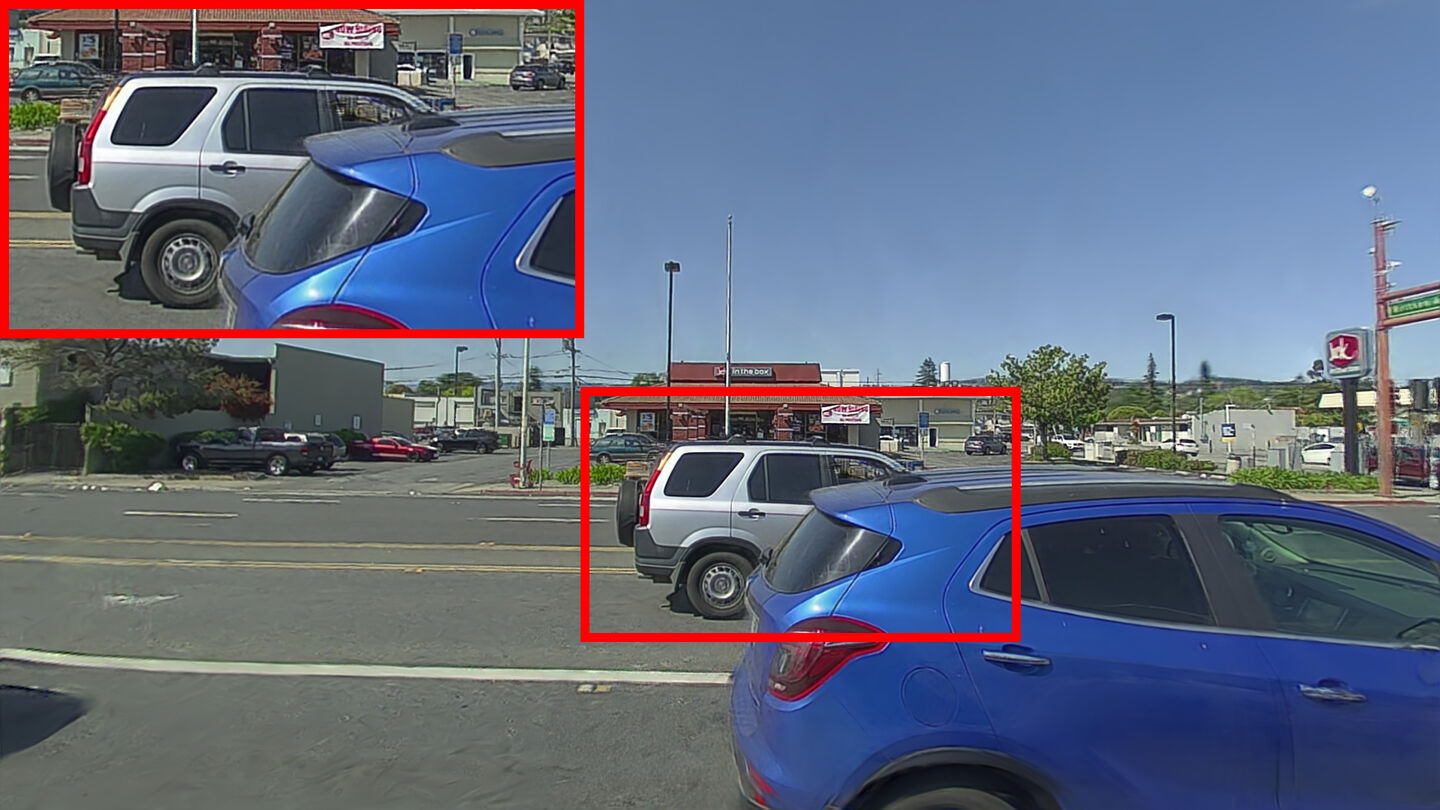} & 
            \includegraphics[width=0.245\linewidth]{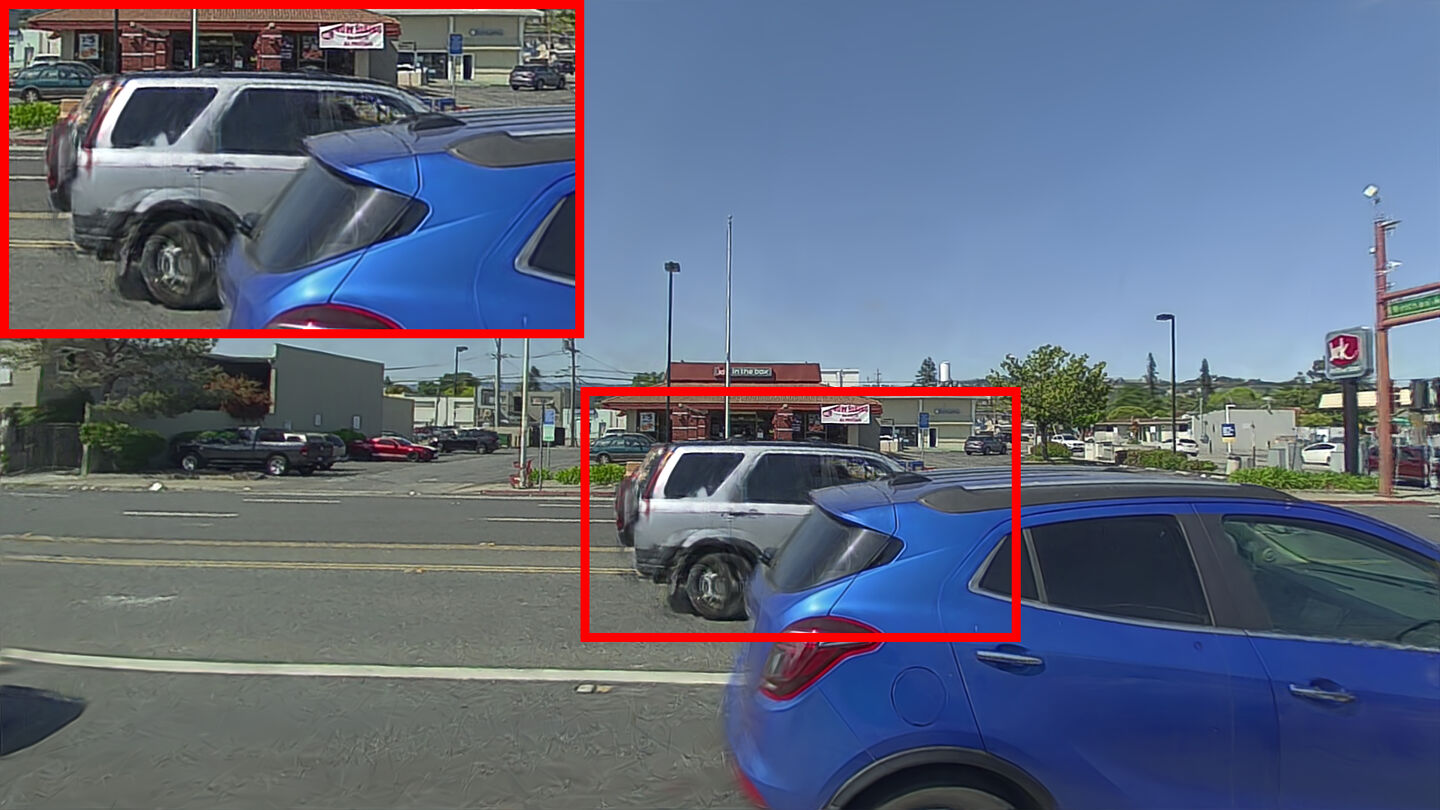} & 
            \includegraphics[width=0.245\linewidth]{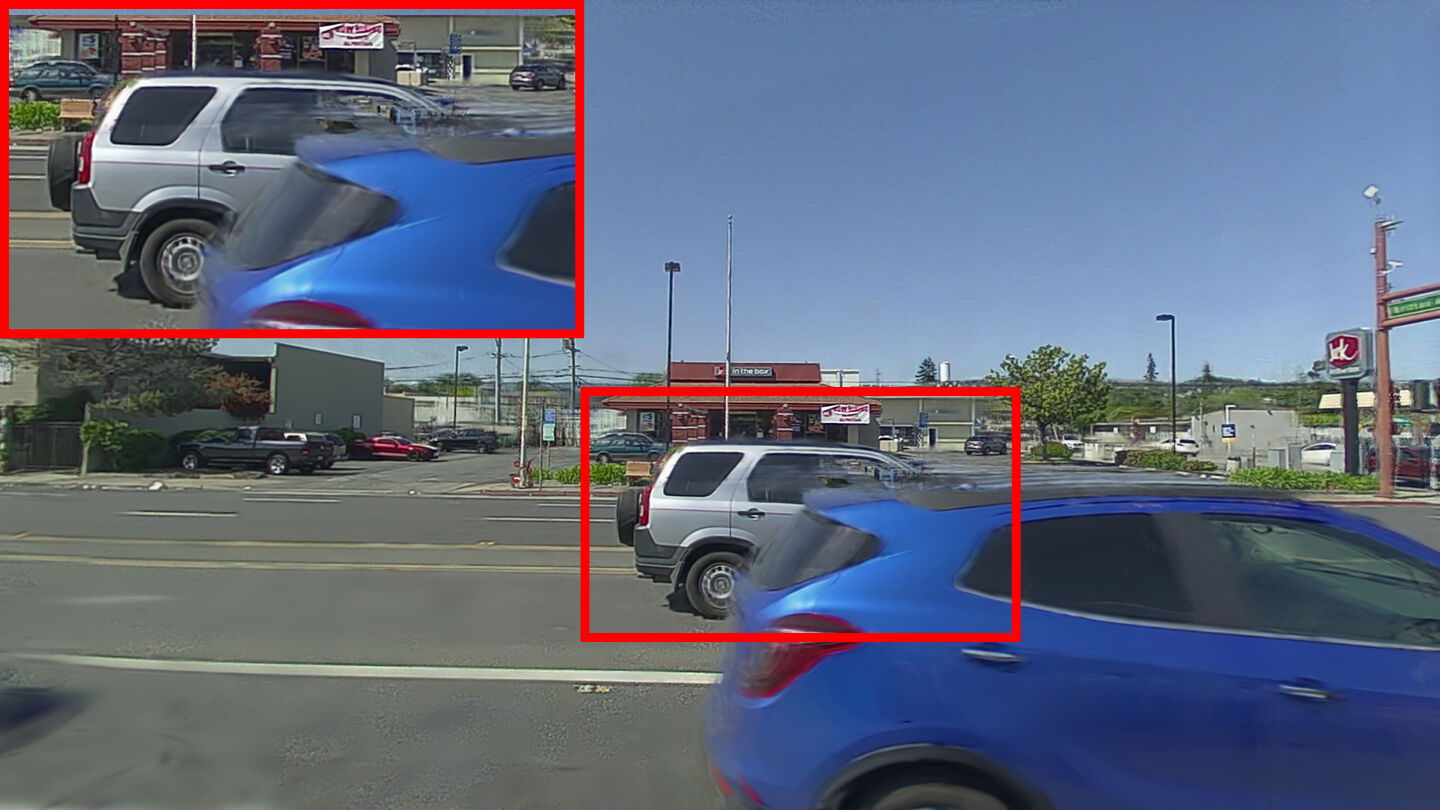} \\

        \includegraphics[width=0.245\linewidth]{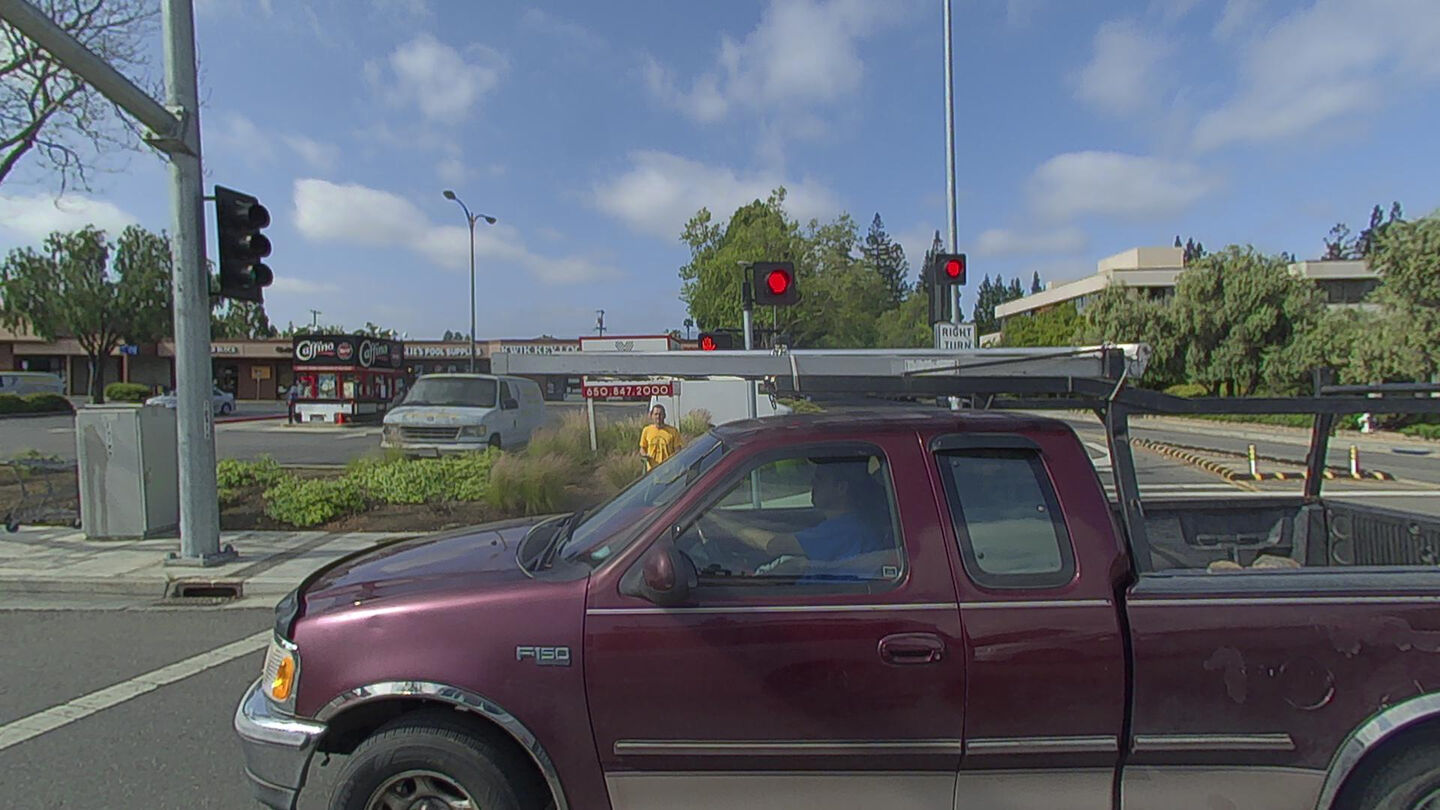} & 
            \includegraphics[width=0.245\linewidth]{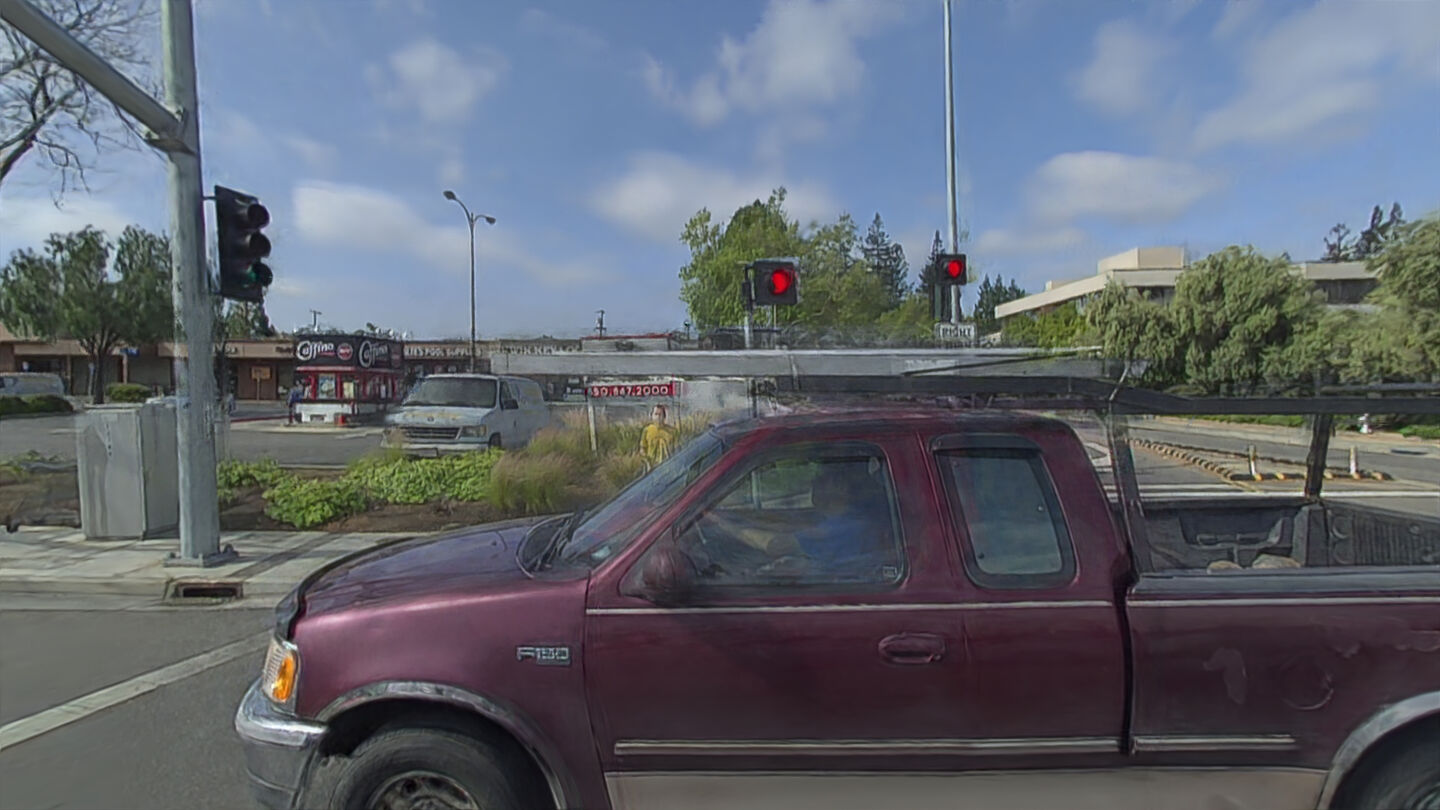} & 
            \includegraphics[width=0.245\linewidth]{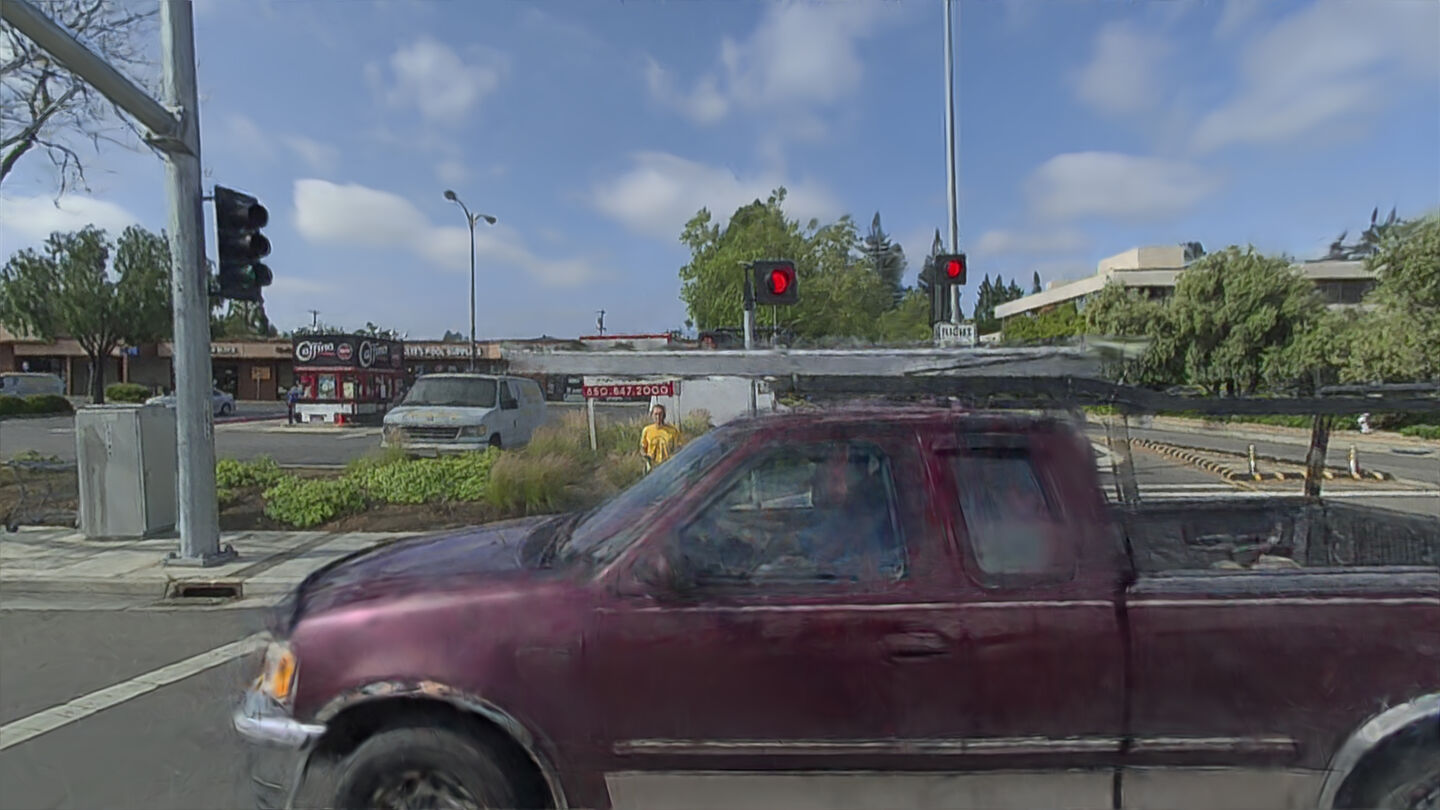} & 
            \includegraphics[width=0.245\linewidth]{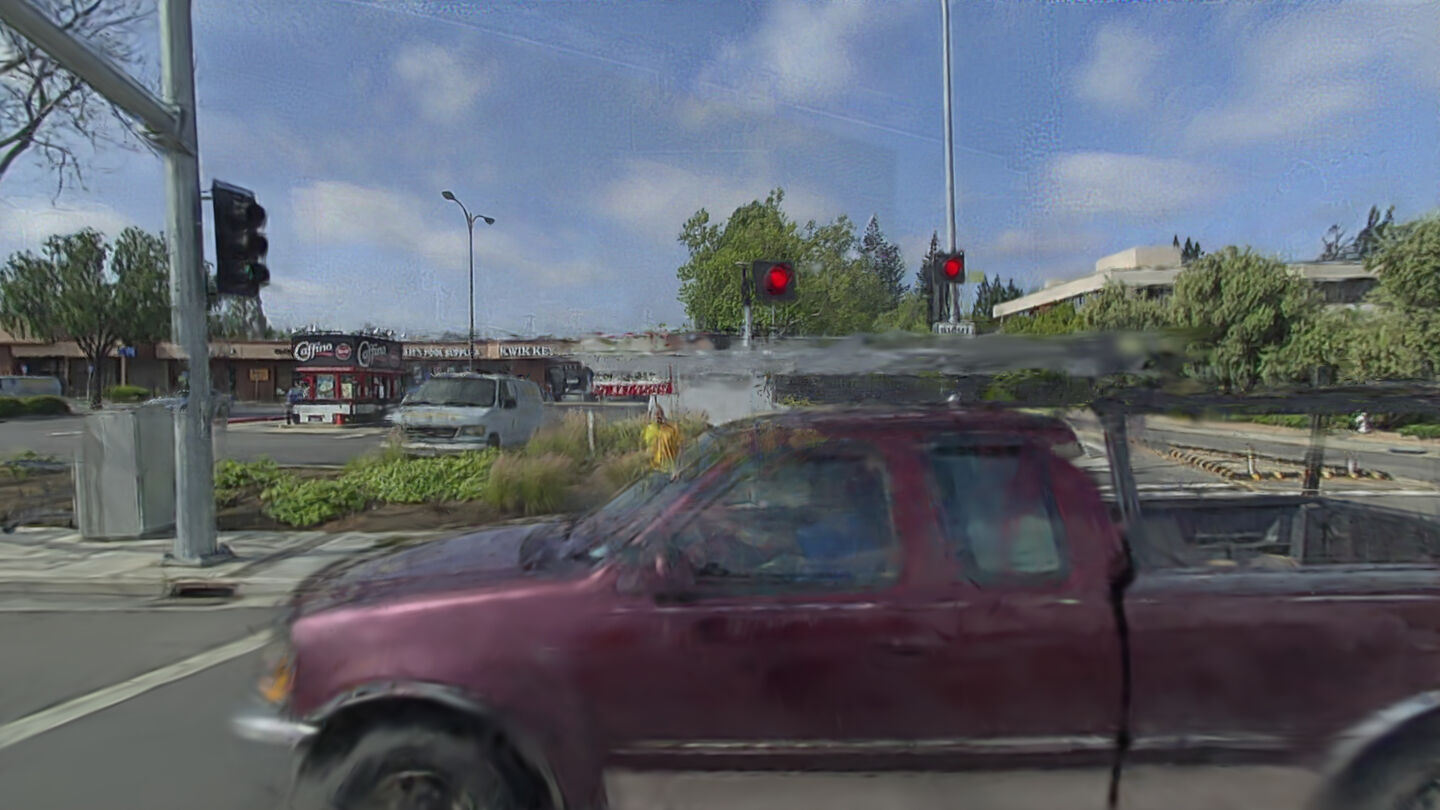} \\
        
    \end{tabular}
    \caption{Pandaset dataset. RGB camera rendering comparison. Our general rolling-shitter modeling enables accurate actor reconstructions with precise appearance and geometry under dynamic complex lighting conditions.}\label{fig:pandaset}
\end{figure*}

In this section we provide additional qualitative comparisons with previous methods.
While previous methods provide noisy geometry reconstructions illustrated in \Cref{fig:lidars}, XSIM precisely reconstructs geometry without artifacts~\Cref{fig:depth2}. Consistency between reconstructed appearance and geometry representations allow XSIM to produce clean renders with low distortions even for laterally shifted trajectories, as illustrated in \Cref{fig:nvs}. 
Additionally, we demonstrate comparison of RGB camera renders on Pandaset dataset in \Cref{fig:pandaset}. 

\end{appendices}

\end{document}